\documentclass[acmtog,table]{acmart}
\AtBeginDocument{%
  \providecommand\BibTeX{{%
    \normalfont B\kern-0.5em{\scshape i\kern-0.25em b}\kern-0.8em\TeX}}}

\usepackage{nicefrac}
\usepackage{xspace}
\usepackage{amsmath}
\usepackage{bm}
\usepackage{pifont}
\usepackage{multirow}
\usepackage{wrapfig}
\usepackage{array}
\newcolumntype{P}[1]{>{\raggedright\arraybackslash}p{#1}}
\usepackage{subfig}
\usepackage{tabularx}
\usepackage{xcolor}
\usepackage{adjustbox}

\newcommand{\figref}[1]{Fig.~\ref{fig:#1}}

\newcommand{\secref}[1]{Sec.~\ref{sec:#1}}

\newcommand{\tableref}[1]{Table~\ref{tab:#1}}

\newcommand{\cdms}{\,cd/m$^2$\xspace}

\usepackage{etoolbox}
\ifdef{\LetLtxMacro}{}{\usepackage{letltxmacro}}
\LetLtxMacro{\originaleqref}{\eqref}
\renewcommand{\eqref}[1]{Eq.~\originaleqref{eq:#1}}

\newcommand{\ourmethod}{\href{https://github.com/gfxdisp/ColorVideoVDP}{ColorVideoVDP}}
\newcommand{\ourdataset}{\href{https://doi.org/10.17863/CAM.108210}{XR-DAVID}}

\newcommand{\ind}[1]{\text{#1}}

\newcommand{\sust}{\mathcal{S}}

\DeclareMathOperator*{\argmin}{argmin}

\newcommand{\pixcoord}{\bm{x}} 

\newcommand{\testF}{\textrm{test}} 
\newcommand{\refF}{\textrm{ref}} 
\newcommand{\maskF}{\textrm{mask}} 

\newcommand{\gausspyr}{\mathcal{G}} 
\newcommand{\laplpyr}{\mathcal{L}} 

\newcommand{\cellNo}{\cellcolor{red!10} No}
\newcommand{\cellYes}{\cellcolor{green!10} Yes}

\setcopyright{rightsretained}
\acmJournal{TOG}
\acmYear{2024} \acmVolume{43} \acmNumber{4} \acmArticle{129} \acmMonth{7}\acmDOI{10.1145/3658144}

\acmConference[SIGGRAPH '24 Technical Papers]{SIGGRAPH 2024 Technical Papers}{July 28--Aug 1, 2024}{Denver, USA}
\acmBooktitle{SIGGRAPH 2024 Technical Papers (SIGGRAPH '24 Technical Papers), Jul 28-- Aug 1, 2024, Denver, USA}

	

\acmSubmissionID{papers\_236}

\begin{document}

\title{ColorVideoVDP: A visual difference predictor for image, video and display distortions}

\author{Rafa{\l} K. Mantiuk}
\email{rafal.mantiuk@cl.cam.ac.uk}
\orcid{0000-0003-2353-0349}
\affiliation{%
  \institution{University of Cambridge}
  \streetaddress{William Gates Building, 15 JJ Thomson Avenue}
  \city{Cambridge}
  \country{UK}
  \postcode{CB3 0FD}
}

\author{Param Hanji}
\email{param.hanji@gmail.com}
\orcid{0000-0002-8142-5611}
\affiliation{%
  \institution{University of Cambridge}
  \streetaddress{William Gates Building, 15 JJ Thomson Avenue}
  \city{Cambridge}
  \country{UK}
  \postcode{CB3 0FD}
}

\author{Maliha Ashraf}
\email{ma905@cam.ac.uk}
\orcid{0000-0002-8142-5611}
\affiliation{%
  \institution{University of Cambridge}
  \streetaddress{William Gates Building, 15 JJ Thomson Avenue}
  \city{Cambridge}
  \country{UK}
  \postcode{CB3 0FD}
}

\author{Yuta Asano}
\email{yasano@meta.com}
\orcid{0009-0003-8612-9349}
\affiliation{%
  \institution{Reality Labs}
  \streetaddress{}
  \city{Redmond}
  \country{USA}
  \postcode{}
}

\author{Alexandre Chapiro}
\email{alex@chapiro.net}
\orcid{0000-0002-7367-0131}
\affiliation{%
  \institution{Reality Labs}
  \streetaddress{}
  \city{Sunnyvale}
  \country{USA}
  \postcode{}
}


\begin{abstract}
ColorVideoVDP is a video and image quality metric that models spatial and temporal aspects of vision for both luminance and color. The metric is built on novel psychophysical models of chromatic spatiotemporal contrast sensitivity and cross-channel contrast masking. It accounts for the viewing conditions, geometric, and photometric characteristics of the display. It was trained to predict common video-streaming distortions (e.g., video compression, rescaling, and transmission errors) and also 8 new distortion types related to AR/VR displays (e.g., light source and waveguide non-uniformities). To address the latter application, we collected our novel XR-Display-Artifact-Video quality dataset (XR-DAVID), comprised of 336 distorted videos. Extensive testing on XR-DAVID, as well as several datasets from the literature, indicate a significant gain in prediction performance compared to existing metrics. ColorVideoVDP opens the doors to many novel applications that require the joint automated spatiotemporal assessment of luminance and color distortions, including video streaming, display specification, and design, visual comparison of results, and perceptually-guided quality optimization. The code for the metric can be found at \url{https://github.com/gfxdisp/ColorVideoVDP}. 
\end{abstract}



\begin{teaserfigure}
  \includegraphics[width=\textwidth]{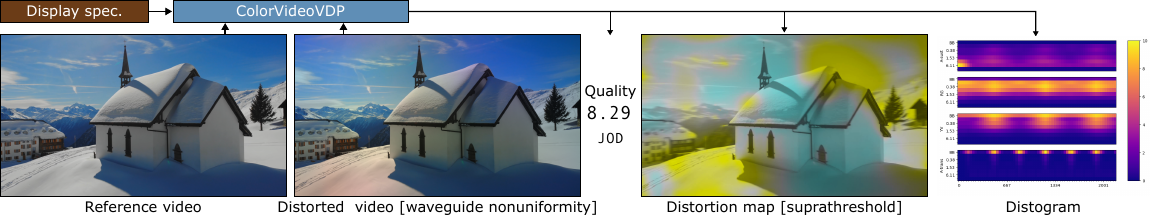}
  \caption{\ourmethod{} predicts the visibility of distortions for a pair of test and reference videos (or images) as seen on a display with a provided specification. The predictions are represented as a single quality value in Just-Objectionable-Difference (JOD) units, a distortion map video, and a \emph{distogram}, which visualizes the distortions over time, separately for each channel and spatial frequency band. See \figref{allartifacts} for the content attribution.
  }
  \Description{}
  \label{fig:teaser}
\end{teaserfigure}


\maketitle

\section{Introduction}
Evaluating the visual quality of displayed content is a perennial task in computer graphics and display engineering. The most direct route, involving visual appraisal by human observers, is often too costly and slow. Subjective studies may also be infeasible when a large trade-space of competing variables needs to be studied quickly to find optimal settings. In this case, automated metrics are of great importance as tool for evaluation and design, which are often employed as cost functions for optimization. 

This need led to the creation of many general-purpose image and video metrics, but these techniques often ignore important aspects of human vision, such as color or temporal vision. This happens due to the inherent complexity of the visual system, which does not allow for holistic modelling. Further, accurate models typically rely on psychophysical data to make predictions, but due to the multi-dimensional nature of color, data on interactions between color and spatiotemporal characteristics of stimuli can be difficult to model. 

Accurate reproduction of contrast and color is of central importance to the quality of displayed content. Achromatic artifacts stemming from graphics pipelines, such as visible blur or contrast loss, can be modelled by existing luminance-only metrics such as SSIM \cite{Wang2003d} or FovVideoVDP \cite{Mantiuk2021}, but color ones, such as chroma subsampling, cannot. On the other hand, color difference formulas, such as the popular CIEDE2000 \cite{Sharma2005,CIE15_2018}, do not model spatial or temporal aspects of vision, and as a consequence may ignore important aspects of an artifact, such as the spatial distribution or changes of color distortion over time. Notably, spatiotemporal color artifacts are especially problematic in modern display applications, in particular for emerging display technologies such as wide-color-gamut, virtual and augmented reality (XR) displays. The latter require novel architectural solutions, which lead to chromatic artifacts like color fringing caused by lens aberrations, or chromatic nonuniformity due to optical waveguides (see \figref{teaser}).

This work presents \ourmethod{}, a full-reference quality metric that models spatiotemporal achromatic and chromatic vision. 
The metric is built on novel psychophysical models of chromatic spatiotemporal contrast sensitivity and cross-channel contrast masking. Thanks to its psychophysical foundations, the metric accounts for the physical specification of a display (size, resolution, color characteristic) and viewing conditions (viewing distance and ambient light). This is the first video and image quality metric that explicitly models human spatiotemporal and chromatic vision simultaneously and is capable of modeling XR display artifacts. Our work builds on FovVideoVDP \cite{Mantiuk2021} borrowing its general processing pipeline. The main novel contribution is integrating the spatio-chromatic model of near- and supra-threshold contrast perception, which lets us quantify the visibility of both achromatic and chromatic distortions. As chromatic distortions tend to reside in lower spatial frequencies, \ourmethod{} also improves the modeling of low-frequency differences. Compared to FovVideoVDP, \ourmethod{} is fully differentiable\footnote{FovVideoVDP does not model the differences in low-frequency bands and, therefore, it cannot propagate low-frequency differences when used as a differentiable loss term. } and much improves the accuracy of quality predictions on non-foveated image and video datasets.


The key challenge of developing any new quality metric is its effective calibration and robust validation. To that end, and to ensure that the metric can provide reliable predictions for display applications, we collected a new XR-Display-Artifact-Video quality dataset (\ourdataset{}) with 8 common display artifacts (\secref{xr-david}). To ensure the diversity of distortion and content types, we combined our new \ourdataset{} dataset with a large HDR/SDR image dataset UPIQ \cite{Mikhailiuk2021}. As the combined datasets consist of terabytes of data, calibration required a non-trivial mixture of end-to-end, and feature-space training. This effort allowed us to match and exceed the state-of-the-art results on unseen datasets (in a cross-dataset validation) and on the testing portion of the training datasets (cross-content validation). In \secref{applications}, we demonstrate new applications of \ourmethod{}, including analysis of chroma subsampling, display color tolerance specification, and quantifying observer metamerism variations on a target dataset. 
\paragraph{Limitations} \ourmethod{} lacks higher-level models of saliency or annoyance, resulting in lower accuracy when the semantic content has a strong influence on the quality judgments. It was not trained to predict accurate spatial distortion maps \cite{Ye2019} (as no such data is available for video).
\ourmethod{} does not model the effect of glare (inter-ocular light scatter, found in HDR-VDP \cite{Mantiuk2023}), gaze-contingent vision (found in FovVideoVDP \cite{Mantiuk2021}), eye motion \cite{Laird2006,Denes2020}, or binocular vision \cite{Didyk2011}.

\section{Related Work}

\begin{table*}[!t]
\caption{Quality/difference metrics and their capabilities. Columns indicate whether the metric models spatial vision, temporal vision, color vision, and whether it accounts for display geometry and photometry. }
 \label{tab:metric-table}
 \small
\begin{tabular}{*4lp{20mm}p{55mm}}    \toprule
\emph{Metric} & Spatial & Temporal & Color & Display model & Approach\\ \midrule

PSNR    & \cellNo  & \cellNo & \cellNo & \cellNo & Signal quality   \\ 

\midrule

CIEDE2000~\cite{CIE15_2018} & \cellNo  & \cellNo & \cellYes & \cellYes & Color difference formula  \\ 

$\Delta\text{E}_\ind{ITP}$~\cite{ITU-RBT.2124} & \cellNo  & \cellNo & \cellYes & \cellYes & Color difference formula \\ 

\midrule

sCIELAB~\cite{Zhang1997a}  & \cellYes  & \cellNo & \cellYes & \cellYes & CSF + Color difference formula   \\ 

$\Delta\text{E}_\ind{ITP}^\ind{S}$~\cite{Choudhury2021}  & \cellYes  & \cellNo & \cellYes & \cellYes & CSF + Color difference formula   \\ 

\midrule

MS-SSIM~\cite{Wang2003d}  & \cellYes  & \cellNo  & \cellNo & \cellNo  & Multi-scale structural similarity   \\

FSIMc~\cite{LinZhang2011}  & \cellYes  & \cellNo  & \cellYes & \cellNo  & Similarity of phase congruency and gradients    \\

VSI~\cite{Zhang2014}  & \cellYes  & \cellNo  & \cellYes & \cellNo  & Saliency + SSIM   \\

LPIPS~\cite{zhang2018perceptual}  & \cellYes & \cellNo  & \cellYes  & \cellNo & Difference of CNN features  \\ 

FLIP~\cite{Andersson2020} & \cellYes & \cellNo  & \cellYes  & \cellNo & CSF + Color difference + edge detectors  \\ 

IQT~\cite{Cheon_Yoon_Kang_Lee_2021} & \cellYes & \cellNo  & \cellYes  & \cellNo & CNN features + transformer autoencoder  \\ 

\midrule

STRRED~\cite{soundararajan2012video}  & \cellYes & \cellYes & \cellNo  & \cellNo & Entropy differences in wavelet subbands  \\ 

VMAF~\cite{Li2016}  & \cellYes & \cellYes & \cellNo & \cellNo & Features + SVR  \\ 

FUNQUE~\cite{Venkataramanan_2022}  & \cellYes & \cellYes & \cellNo & \cellNo & Wavelet decomposition + features + SVR  \\ 

\midrule

HDR-VDP-3~\cite{Mantiuk2023}   & \cellYes & \cellNo  & \cellNo & \cellYes & Psychophysical model   \\

FovVideoVDP~\cite{Mantiuk2021}  & \cellYes & \cellYes & \cellNo & \cellYes & Psychophysical model  \\ 

\textbf{\ourmethod{} (ours)}  & \cellYes & \cellYes & \cellYes & \cellYes & Psychophysical model \\ 
\bottomrule
 \hline
\end{tabular}
\vspace{-3mm}
\end{table*}

This section reviews the existing work that addresses the problem of predicting visible differences or quality in color images and video --- the main focus of our metric. The representative examples of these methods are listed in \tableref{metric-table}. The table also specifies whether the metric attempts to model spatial vision, temporal vision, offers distinctive processing of color and whether it accounts for the colorimetric characteristic of the display, its resolution, and viewing distance. As shown in the table, no existing metric is capable of addressing these four important areas of image quality. Next, we review the metrics by the groups indicated in the table.


\paragraph{Color difference formulas} Perceived color differences for uniform patches can be predicted using one of the standard display formulas, such as CIE $\Delta\text{E}^{*}_{ab}$ or CIEDE2000 \cite{Sharma2005,CIE15_2018}. The standard CIE formulas, however, were not meant to predict differences for luminance below 1\cdms{} or above the illuminance of 1\,000\,lux \cite{CIE_1993} (this translates to approximately 318\cdms for a Lambertian white surface). $\Delta\text{E}_\ind{ITP}$~\cite{ITU-RBT.2124} was proposed as a color difference formula for luminance levels found in wide-color-gamut high-dynamic-range television, and potentially intended for video content. It is unclear how to use the color difference formula with complex images and the color difference is typically computed per pixel and then averaged. Such treatment obviously ignores all spatial and temporal aspects of vision, which are partially addressed by the next group of metrics. 

\paragraph{Spatial color difference formulas} The spatial component of vision was included in a spatial extension of the CIELAB difference formula --- sCIELAB \cite{Zhang1997a}. The authors proposed to compute the CIE $\Delta\text{E}^{*}_{ab}$ color differences on images prefiltered by a contrast sensitivity function (CSF). Flip and HDR-Flip metrics \cite{Andersson2020,Andersson2021a} improve on sCIELAB by employing a color difference formula that better quantifies large color differences. Both metrics also emphasize differences at edges, which tend to be more salient. \citeauthor{Choudhury2021} proposed to compute color differences in the ITP color space \shortcite{Choudhury2021}, which is more suitable for HDR color values. The strength of such spatial extensions is their simplicity. The main weakness is that such an application of the CSF is overly simplistic --- it does not account for the changes in contrast sensitivity with luminance and does not account for supra-threshold vision (e.g. contrast masking and contrast constancy). 

\paragraph{Image quality} The most popular image quality metrics, such as SSIM or MS-SSIM \cite{Wang2003d}, do not attempt to explicitly model human vision, but, instead, they combine hand-crafted statistical measures that are likely to correlate with quality judgments. Although the early metrics, such as SSIM and MS-SSIM, operate only on the luma channel of the image, some later metrics, including FSIMc \cite{LinZhang2011} and VSI \cite{Zhang2014}, separate images into luma and two chroma channels, akin to the color space transforms used in video compression. Those metrics, however, do not account for the geometry of a display (e.g. resolution, size, viewing distance), nor for its photometry (e.g. peak brightness, black level). The latter shortcoming can be addressed by employing the Perceptually Uniform (PU) transform \cite{Mantiuk2021pu}, which also extends these metrics to operate on high-dynamic-range images. 


\citeauthor{zhang2018perceptual} \shortcite{zhang2018perceptual} observed that the activation layers of many convolution neural networks (CNNs) provide features that are well correlated with human judgments of image similarity. Their proposed metric, LPIPS, became very popular in computer vision, despite its underwhelming performance on image quality datasets \cite{Ding2021}. \citeN{Prashnani2018} proposed training a CNN on triplets of patches, two distorted and one reference, with supervision based on the Bradley-Terry pairwise comparison model. Akin to LPIPS training, their PieAPP metric was trained on a large dataset of patches with pairwise comparison labels. \citeN{Cheon_Yoon_Kang_Lee_2021} combined CNN features with transformer-based embeddings to regress quality scores. Their IQT metric was ranked first among 13 participants in the NTIRE 2021 perceptual image quality assessment challenge.

Image metrics are not meant to predict video quality, however, they can perform surprisingly well in this task. The typical route to employ image metrics to video is to average predictions across individual frames. 

\paragraph{Video quality} Video quality metrics combine both spatial and temporal features. STRRED \cite{soundararajan2012video} compares the per-frame entropy of fitted local distributions of wavelet coefficients. This entropy is then weighted by local spatial and temporal variance. VMAF \cite{Li2016} combines two spatial features (VIF and DLM) with a mean absolute difference of consecutive frames, and then maps those into quality scores using a support vector regression (SVR). The success of VMAF motivated a series of fusion-based metrics, such as FUNQUE \cite{Venkataramanan_2022} and 3C-FUNQUE+ \cite{Venkataramanan_2023}, which use wavelet transform modulated by the contrast sensitivity function as input to several individual predictors, which similarly as VMAF, are combined using SVR. However, all those metrics use simplified models of contrast sensitivity, which ignore the effect of luminance and do not model spatiotemporal vision. For example, the temporal processing of all of these state-of-the-art metrics considers just two consecutive frames, and it does not model the temporal characteristics of human vision. Our metric explicitly models achromatic and chromatic visual channels to address this shortcoming. 

\paragraph{Visual Difference Predictors} The visual difference predictors (VDPs), such as DCTune \cite{Watson1993}, VDP \cite{Daly1993}, HDR-VDP-2 \cite{Mantiuk2011a} and HDR-VDP-3 \cite{Mantiuk2023}, explicitly model aspects of low-level human vision, such as contrast sensitivity and masking. The advantage of this approach is that these metrics are built on sound psychophysical models and generalize more easily to unseen conditions, such as displays of varying size, resolution, or peak luminance. While it is tempting to assume that this type of general modeling would perform worse or be more computationally expensive than metrics relying on hand-crafted features, as we will demonstrate in \secref{evaluation}, this is not the case. Modern VDPs perform on par or better than metrics with hand-crafted features. When they are optimized to run on a GPU, they are as fast as feature-based metrics. 


Our \ourmethod{} metric takes inspiration from and is based on similar building blocks as FovVideoVDP \cite{Mantiuk2021} but with several important improvements. First, \ourmethod{} models the visibility of chromatic (color) differences by employing a novel spatiotemporal-chromatic contrast sensitivity function (castleCSF) \cite{Ashraf2024}. This involves modeling both spatial and temporal chromatic vision. Second, we account for supra-threshold color differences so that the perceived magnitude of achromatic and chromatic contrast is properly mapped to the metric response. Third, while FovVideoVDP is insensitive to low-frequency distortions (below 1\,cpd\footnote{cpd --- cycles per visual degree --- the measure of spatial frequency.}), \ourmethod{} correctly accounts for those. This lets us account for low-frequency display artifacts (e.g., waveguide non-uniformity) and use \ourmethod{} as a differentiable loss function, which can propagate both low and high-frequency differences. Fourth, \ourmethod{} models within-channel and cross-channel masking, where each of the four modeled channels (two achromatic and two chromatic) can be masked by the combination of contrast in the other channels. Finally, \ourmethod{} is trained and tested on multiple image and video datasets, both SDR and HDR, including a novel dataset of display artifacts described in \secref{xr-david}. Unlike FovVideoVDP, we do not model foveated vision due to the increased overhead of modeling interactions between eccentricity and color. While modeling foveated vision is necessary in some specific usage scenarios, such as foveated rendering, it is less useful in most general-use cases, such as video streaming, display engineering, and non-foveated algorithm design.


\section{Color video visual difference predictor}

\begin{figure*}
    \centering
    \includegraphics[width=\textwidth]{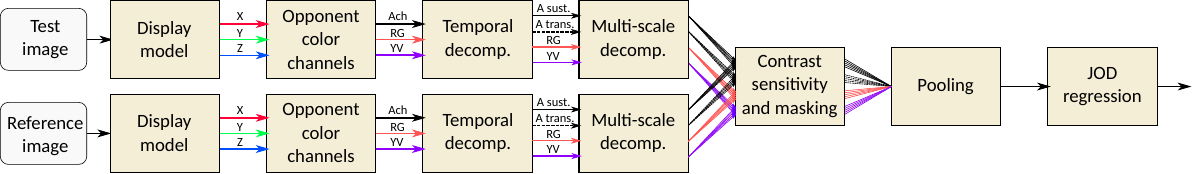}
    \caption{Processing stages of ColorVideoVDP. Test and reference images are first processed by the same pipeline: the display model maps pixel values to linear color (CIE 1931 XYZ color space), linear color is transformed to the opponent color space (DKL), the achromatic channel is decomposed into sustained (continuous lines) and transient (dashed lines) temporal channels, then each of those is decomposed into multiple spatial bands (Laplacian pyramid). The decomposed video/image goes into the contrast sensitivity and masking models, explained in more detail in Figure~\ref{fig:masking-model}. The result of the masking model is pooled across all spatial bands, temporal and color channels, and finally regressed to a JOD score.}
    \label{fig:cvvdp-flowchart}    
\end{figure*}

Our goal is to create an image and video quality metric that accounts for color and spatiotemporal perception and has sound psychophysical foundations. The metric should rely on psychophysical models as they can help to extrapolate predictions to unseen conditions, such as different frame rates, spatial resolution, or absolute luminance levels. The metric should be able to predict a single-valued quality correlate that can help to both evaluate and optimize visual results. It should also produce spatial and temporal difference maps, which provide the visual explanation for the predicted quality correlate. Finally, the metric should be efficient to compute and fully differentiable so that it could be used as an optimization criterion. 

The processing diagram of the metric is shown in \figref{cvvdp-flowchart}. First, the test and reference content, typically stored as video or images, is transformed into colorimetric quantities of light emitted from a given display. Then, the frames are decomposed into color opponent channels, temporal channels and spatial frequency bands, mimicking the mechanisms of the visual system. The core component of the metric is the model of contrast sensitivity and masking, which computes a per-band visual difference between test and reference content. It relies on the near-threshold and supra-threshold models of contrast detection and discrimination. In the last two steps, the visual differences are pooled across the bands and channels and then the resulting visual difference value is regressed into an interpretable Just-Objectionable-Difference (JOD) scale. The following sections explain each step in detail.

\subsection{Display model}
\label{sec:display-model}

The display model has two purposes: to convert spatial pixel coordinates to perceptually meaningful units of visual degrees, and to model the display's photometric response in a given environment. We assume a flat panel display spanning a limited field of view, so we can approximate the conversion from pixel coordinates to degrees in the visual field with a single constant:
\begin{equation}
n_\ind{ppd}=\frac{\pi}{360\,\arctan\left(\frac{0.5\,d_{\ind{width}}}{r_\ind{w}\,d_\ind{v}}\right)},
\end{equation}
where $n_\ind{ppd}$ is the effective display resolutions in pixels per degree, $d_{\ind{width}}$ is the width of the screen, $r_\ind{w}$ is the screen's horizontal resolution in pixels and $d_\ind{v}$ is the viewing distance. The display width and viewing distance must be provided in the same units (e.g. meters). We rely on $n_\ind{ppd}$ later when expressing the spatial frequencies in cycles per visual degree. It should be noted that the above approximation is inaccurate for near-eye displays spanning a large field of view, and accurate geometric mapping should be used for these types of displays (see Equation~2 in \cite{Mantiuk2021}).

The second responsibility of the display model is to convert pixel values represented in one of the standard color spaces into colorimetric quantities of light emitted from a given display. It accounts for the display's peak luminance, color gamut, its black level, and ambient light reflected from the display. The display-encoded pixel values, $I_{\ind{de},c}$ for color channel $c$ ($c\in\{\text{R},\text{G},\text{B}\}$), are transformed into absolute linear colorimetric values:
\begin{equation}
    I_{\ind{lin},c}(\pixcoord) = \min \left\{ (L_\ind{peak}-L_\ind{black})\,E( I_{\ind{de},c}(\pixcoord) ) + L_\ind{black}, L_\ind{peak}\right\} + L_\ind{refl},
\end{equation}
where $L_\text{peak}$ is the peak luminance of the display and $L_\ind{black}$ is its black level. $\pixcoord$ is used to denote spatial pixel coordinates throughout the paper. $E(\cdot)$ is the electro-optical transfer function (EOTF) of particular pixel coding, for example, the sRGB non-linearity (IEC 61966-2-1:1999) for standard dynamic range content and PQ (Perceptual Quantizer, SMPTE ST 2084) for high dynamic range content. Because the PQ EOTF transforms display-encoded values into absolute linear values (between 0.005 and 10\,000), we do not multiply the EOTF by $(L_\ind{peak}-L_\ind{black})$ if PQ is used. We currently do not model tone-mapping, which is present in most displays with HDR capabilities, because it varies from one display to another. Instead, we clip the values at $L_\ind{peak}$.

The amount of light reflected from the display, $L_\ind{refl}$, is computed as:
\begin{equation}
    L_\ind{refl} = k_\ind{refl}\frac{E_\ind{amb}}{\pi}\,,
\end{equation}
where $k_\ind{refl}$ is the reflectivity of the screen (typically 0.01--0.05 for glossy screens, 0.005--0.015 for matt screens) and $E_\ind{amb}$ is the ambient illumination in lux units. As the last step, the linear color values are converted into device-independent CIE XYZ color space. This conversion is standardized for popular color spaces, such as BT.709 or BT.2020, or alternatively can be computed for the primaries of a display.

\subsection{Opponent color channels}
\label{sec:opponent-cs}

The sensitivity to chromatic changes is typically explained for color modulations represented in a space that separates three cardinal mechanisms of human color vision: achromatic channel and two chromatic channels, the latter commonly known as red-green and violet-yellow \cite{Stockman2010}. Here, we use the same color space as our contrast sensitivity function --- the Derrington-Krauskopf-Lennie (DKL) colorspace \cite{Derrington1984}. DKL was selected for castleCSF as this color space is linearly related to cone responses, is well established in vision science, and much of the chromatic contrast detection data was collected in that space. The DKL space coordinates can be computed from the device-independent XYZ (provided by the display model) as:
\begin{equation}
    \begin{bmatrix}
    I_\ind{ach}(\pixcoord) \\
    I_\ind{rg}(\pixcoord) \\
    I_\ind{vy}(\pixcoord) 
    \end{bmatrix}=\begin{bmatrix}
	1 & 1 & 0 \\
	1 & -\frac{L_0}{M_0} & 0 \\
	-1 & -1 & \frac{L_0 +  M_0}{S_0}
\end{bmatrix}
    M_{\ind{XYZ}{\rightarrow}\ind{LMS}}
    \begin{bmatrix}
    I_\ind{X}(\pixcoord) \\
    I_\ind{Y}(\pixcoord) \\
    I_\ind{Z}(\pixcoord) 
    \end{bmatrix}\,,
\end{equation}
where $M_{\ind{XYZ}{\rightarrow}\ind{LMS}}$ is a matrix converting CIE 1931 XYZ coordinates into the LMS cone responses\footnote{Our CSF is defined using CIE 2006 color matching functions while most of the content still relies on the CIE 1931 color matching functions. The matrix was derived to convert between the two using the spectral emission data for an LCD with an LED backlight.}:
\begin{equation}
M_{\ind{XYZ}{\rightarrow}\ind{LMS}} = \begin{bmatrix} 
   0.187596 &  0.585169 & -0.026384 \\
  -0.133397 &  0.405506 &  0.034502 \\
   0.000244 & -0.000543 &  0.019407    
   \end{bmatrix}\,.
\end{equation}
$L_0$, $M_0$ and $S_0$ specify chromaticity of the adapting color. Here, we assume adaptation to a D65 background: CIE 1931 $(x,y) = (0.3127, 0.3290)$, $(L_0, M_0, S_0) = (0.7399, 0.3201, 0.0208)$.

\subsection{Temporal channels}
\label{sec:temp-channels}

\begin{figure}
    \centering
    \includegraphics[width=\columnwidth]{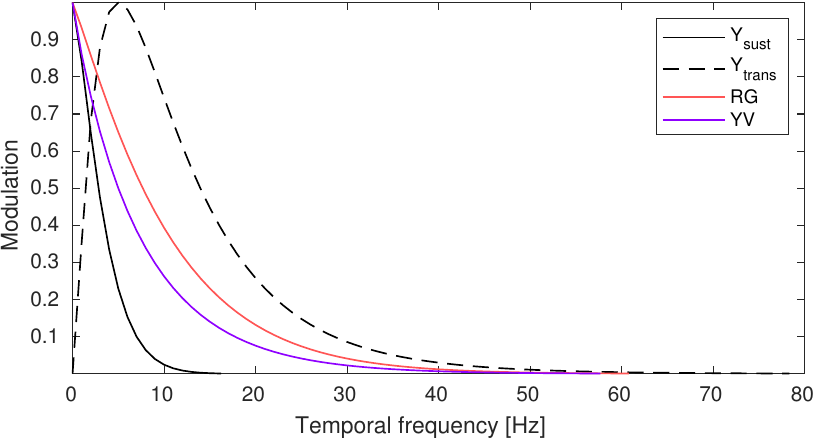}
    \caption{The frequency characteristic of the four temporal channels used in \ourmethod{}.}
    \label{fig:temp-channels}
\end{figure}

Psychophysical masking experiments showed evidence that the information is processed in the visual system by separate temporal channels. Two or three channels have been identified for the achromatic mechanism \cite{Anderson1985,Hess1992}, and one or two channels for the chromatic mechanisms \cite{McKeefry2001,Cass2009}. Here, we assume two temporal achromatic channels, as the third channel was observed only for low frequencies \cite{Hess1992}. We assume just one temporal channel for the red-green and violet-yellow cardinal directions as the evidence for the second channel shows that it has a less prominent role \cite{Cass2009}. Modeling fewer temporal channels also has the benefit of lower memory and computational cost. 

The temporal channels of \ourmethod{} are directly defined by castleCSF \cite{Ashraf2024} (see next section). The frequency tuning of the channels can be seen in \figref{temp-channels}. To find digital filters, we perform a real-valued (symmetric) fast Fourier transform on the frequency space filters. We found that a filter support of 250\,ms is sufficient to capture filter characteristics. Finally, the two achromatic and two chromatic channels are convolved with the digital filters along the time dimension. This splits the achromatic signal into sustained (low-pass) and transient (band-pass) channels. The chromatic channels are low-pass filtered and become insensitive to high-frequency chromatic flicker.


\subsection{Contrast sensitivity}
\label{sec:CSF}

\begin{figure*}
    \centering
    \includegraphics[width=\textwidth]{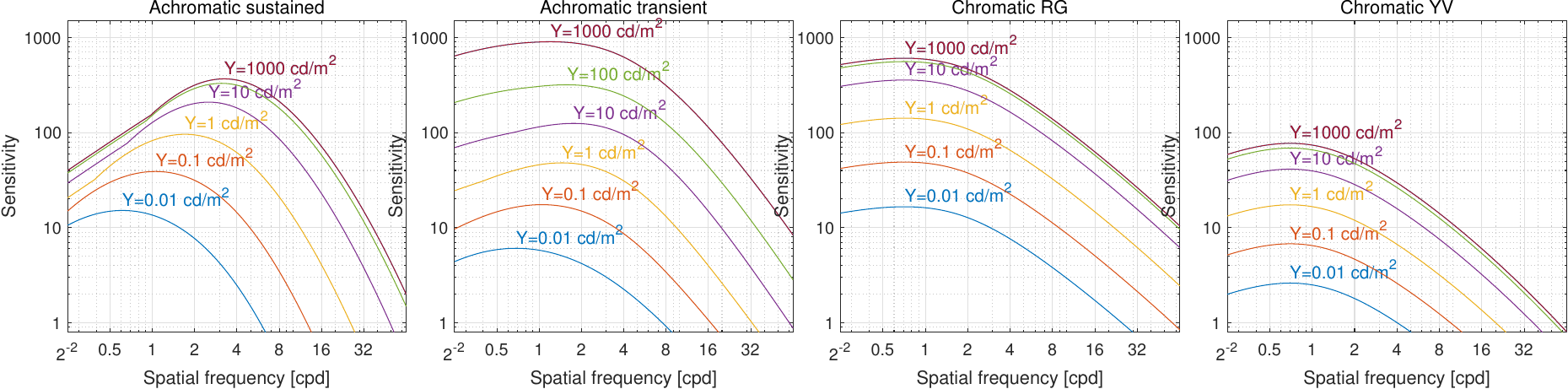}
    \caption{castleCSF contrast sensitivity function for the four channels of \ourmethod{}. The sensitivity is expressed in the cone-contrast units \cite{Wuerger2020}. Note that the achromatic transient and chromatic RG channels appear to have higher sensitivity than the achromatic sustained channel. This is due to the scaling used in the DKL color space \cite{Derrington1984} to represent chromatic contrast units. In practice, the transient and two chromatic channels are much less sensitive to patterns found in complex images. }
    \label{fig:castleCSF}
\end{figure*}

A contrast sensitivity function models our ability to detect patterns of different spatial and temporal frequency, size, and shown at different luminance levels. It is a cornerstone of \ourmethod{} --- it defines the temporal and chromatic channels and enables modeling of contrast masking, which is the key component of our metric. Since there is no contrast sensitivity model that could explain sensitivity to both chromatic modulations and different temporal frequencies, we have created our own model, named castleCSF. This CSF models color, area, spatial and temporal frequencies, luminance, and eccentricity. Because of its complexity, castleCSF is explained in detail in a separate paper \cite{Ashraf2024}. Here, we summarize its key components. 

castleCSF decomposes contrast into three cardinal directions of the DKL color space, the same as those we used to separate achromatic and chromatic channels in \secref{opponent-cs}. The achromatic direction is split into sustained and transient channels, again the same as used by our metric. That lets us directly map the mechanism modeled in castleCSF to the channels in \ourmethod{}. 

To find the detection threshold, castleCSF modulates (multiplies) the contrast associated with each mechanism by the sensitivity of that mechanism and then pools those to form the contrast energy. The detection threshold is assumed to be the contrast at which the energy is equal to 1. The sensitivity of each mechanism is modeled as a function of spatial and temporal frequency, area, background luminance and eccentricity, in a similar manner as for stelaCSF \cite{mantiuk2022stelacsf}. castleCSF was optimized to predict the data from 19 contrast sensitivity datasets (10 achromatic, 6 chromatic and 3 mixed). 

castleCSF predicts sensitivity as the inverse of cone contrast \cite{Wuerger2020} while \ourmethod{} operates on a contrast in the DKL color space. To obtain the sensitivity units consistent with our contrast definition, we use the contrast transformation procedure from \cite{Kim2021} and explained in more detail in the \href{https://www.cl.cam.ac.uk/research/rainbow/projects/colorvideovdp/}{supplementary}. Because we do not account for foveated viewing in \ourmethod{}, we assume that the eccentricity is 0. We found in ablations that the metric performs the best when the area is set to $a=\pi\,1.5^2$\,deg$^2$. The resulting sensitivity functions are plotted in \figref{castleCSF}.


\subsection{Multi-scale decomposition and color contrast}

In addition to the temporally-tuned channels, psychophysical \cite{Foley1994a,stromeyer1972spatial} and neuropsychological \cite{de1982spatial} evidence proves the existence of channels that are tuned to the bands of spatial frequencies and orientations. We must account for such channels to model visual masking, as explained in the next section. Following FovVideoVDP \cite{Mantiuk2021}, we use the Laplacian pyramid \cite{Burt1983} to decompose each of the four temporal channels (Y-sustained, Y-transient, RG, YV) into spatial-frequency selective bands. However, unlike FovVideoVDP, we consider low frequencies, including the base band (the coarsest low-pass band in the Laplacian pyramid). This change is due to finding that many display distortions can only be detected in the low-spatial frequency bands (see \figref{teaser}). We select the height of the pyramid so that the lowest peak frequency of the band-pass filter is greater than 0.2\,cpd. Similarly as in FovVideoVDP, we do not consider orientation-selective channels because of the substantial computational overhead of those.

The contrast at the spatial frequency band $b$, channel $c$ and frame $f$ is computed as:
\begin{equation}
    C_{b,c,f}(\pixcoord)=\frac{\laplpyr_{b,c,f}(\pixcoord)}{\Uparrow\gausspyr_{b+1,\sust,f}(\pixcoord)}=\frac{\laplpyr_{b,c,f}(\pixcoord)}{L_{\ind{bkg},b,f}(\pixcoord)}\,,
    \label{eq:local-contrast}
\end{equation}
where $\laplpyr_{b,c,f}$ represents the $b$-th band of the Laplacian pyramid, $\gausspyr_{b+1,\sust,f}$ is the $(b+1)$ band of the Gaussian pyramid and $\Uparrow$ is the upsampling (expand) operator. The Laplacian pyramid is created for each frame of each channel, obtained by temporal filtering (\secref{temp-channels}). 
The Laplacian pyramid coefficients are divided by the values from the Gaussian pyramid at the band $b+1$ (one lever coarser) of the sustained luminance channel ($\sust$). Note that the upsampled version of the Gaussian pyramid band $b+1$ is a by-product of computing a Laplacian pyramid so can be obtained at no computational cost. This formulation is similar to that of local band-limited contrast \cite{Peli1990}, but here we extend it to both achromatic and chromatic channels. These contrast values are computed separately for the test and reference frames. The values in all bands except the low-pass (base-band) and the high-pass bands are multiplied by 2 to account for the reduction in amplitude \cite[Fig.~5]{Mantiuk2021}.

\subsection{Cross-channel contrast masking}
\label{sec:xcm}

\begin{figure}
\centering
    {\small Reference image}\\
    \includegraphics[width=\columnwidth]{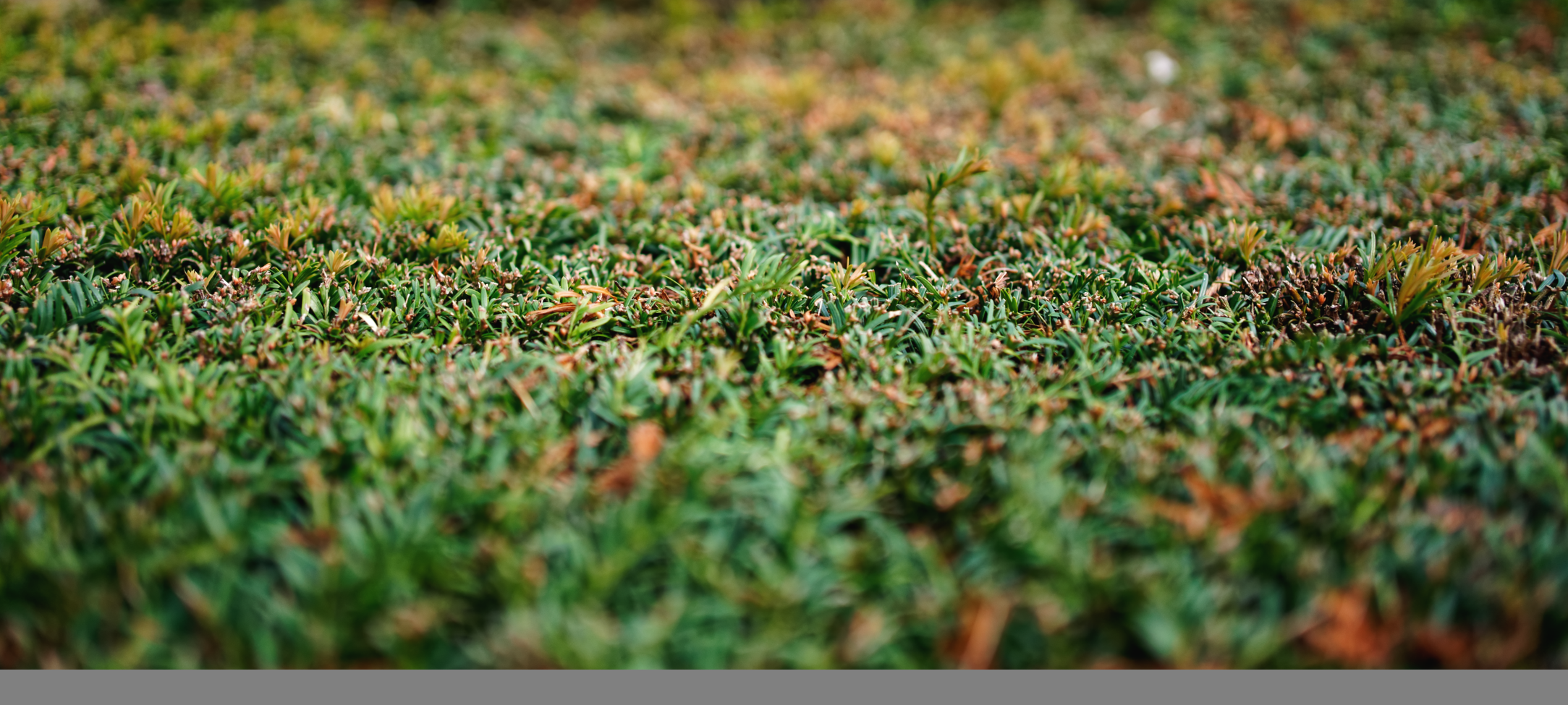}\\
    {\small Test image}\\
    \includegraphics[width=\columnwidth]{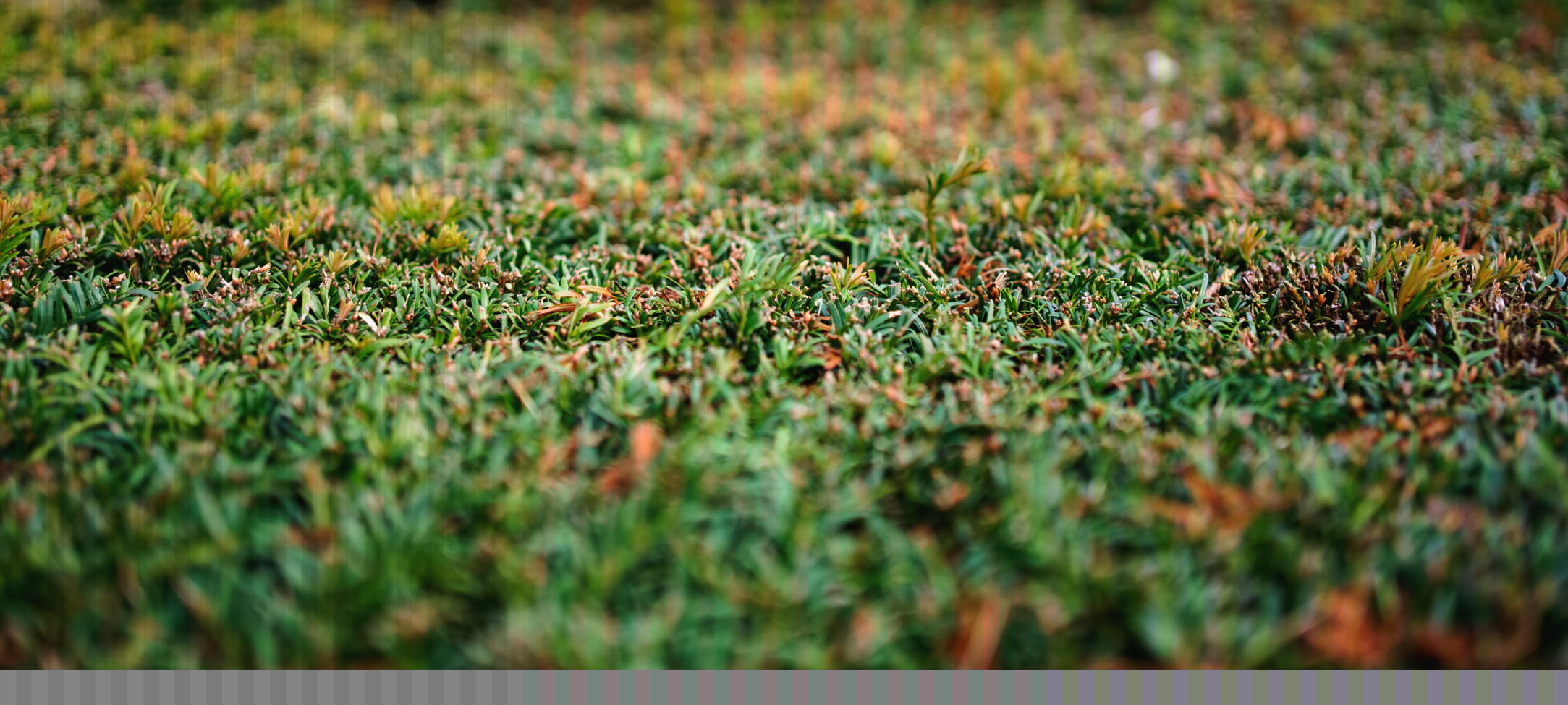}\\    
    {\small Visual difference map --- no masking}\\
    \includegraphics[width=\columnwidth]{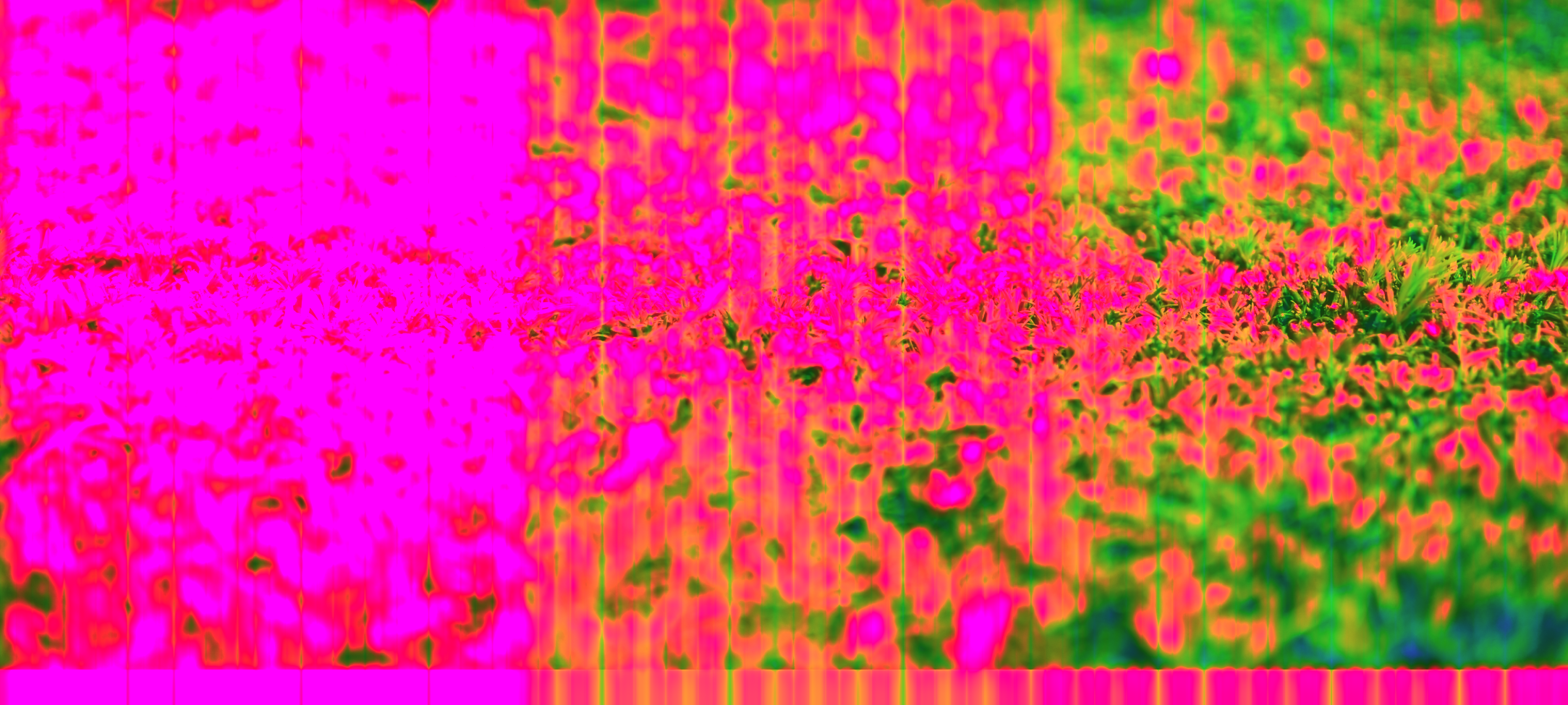}\\        
    {\small Visual difference map --- with masking}\\
    \includegraphics[width=\columnwidth]{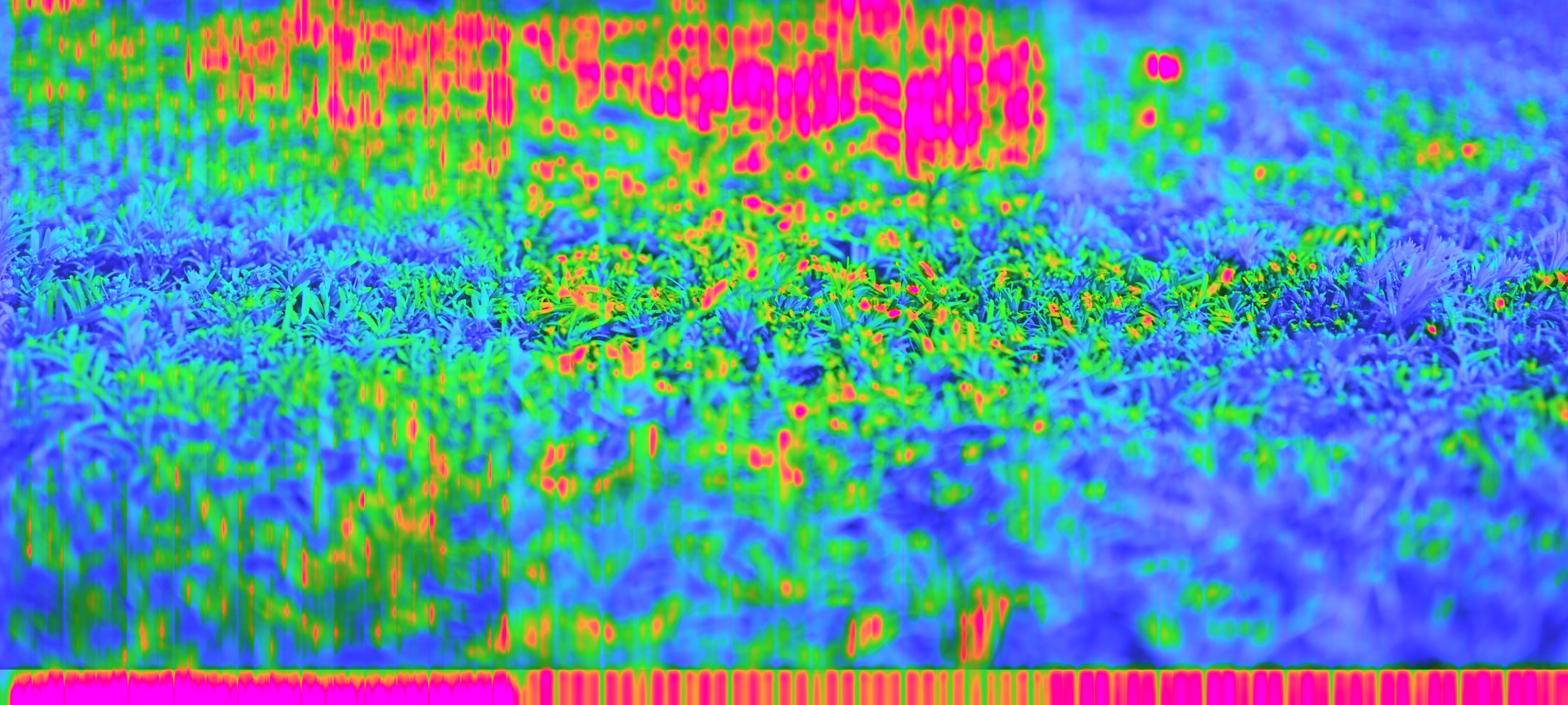}\\    
\caption{Example of contrast masking. A sinusoidal grating of 4\,cpd (when seen printed from 40\,cm) was added to a reference image (top row) to obtain a distorted image (second row). The image was split into three parts and the grating was modulated along achromatic, red-green, and yellow-violet directions in each respective part. 
The third row shows the visual difference map generated by \ourmethod{} without masking but still using the CSF. The map over-predicts the visibility in textured areas. The bottom row shows the prediction with the masking model. It is worth noting: although all three color directions are equally visible in the gray bar at the bottom of the image (where there is little masking), the red-green pattern is more visible in the textured area because it is weakly masked by the achromatic channel. \ourmethod{} correctly predicts this phenomenon.}
      \label{fig:masking-example}
\end{figure}


\begin{figure}
    \centering
    \includegraphics[width=\columnwidth]{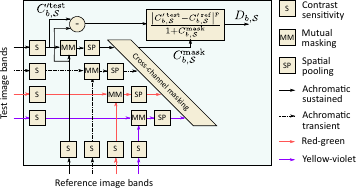}
    \caption{Our masking model. Here, the resulting visual difference for the sustained channel, $D_{b,\sust}$, is visualized for a single spatial frequency band (the frame index $f$ is omitted for clarity). Each band response is multiplied by the contrast sensitivity function ("S" boxes). The CSF-normalized band-responses are used to calculate the difference between test and reference frames. The masking signal is computed by first finding mutual masking between both channels ("MM" blocks), applying a spatial pooling in the local neighborhood of each pixel ("SP" blocks) and then combining the masking signal from multiple channels (cross-channel masking). The visual per-channel and per-band difference between the test and reference is calculated as the ratio of excitatory difference between the test and reference images, and the inhibitory masking signal, as shown in the equation in the box and in \eqref{masking-model}. } 
    \label{fig:masking-model}    
\end{figure}

The model of cross-channel contrast masking transforms local physical contrast differences between the test and reference frames into perceived differences --- differences that are scaled by the local contrast visibility. It accounts for both contrast sensitivity, such as lower sensitivity to high spatial and temporal frequencies, and contrast masking. Masking accounts for differences being less likely to be noticed in heavily textured areas. The cross-channel component models that a strong contrast in one channel can reduce the visibility of contrast in another channel \cite{Switkes_1988}, for example, contrast in the chromatic channels can mask contrast in achromatic channels. Finally, the masking model also needs to account for suprathreshold contrast perception --- must match the perceived magnitude of contrast (changes) across luminance, frequencies, and the directions of color modulation.

Contrast masking is the critical component of the metric that determines its performance, as we show later in the ablation studies (\secref{ablations}). To make the optimal choice, we analyzed and compared six models: two contrast encodings (additive and multiplicative) combined with three masking models. We tested the mutual masking model from the original VDP \cite{Daly1993}, the contrast transducer proposed by \citeN{Watson1997}, and the contrast similarity formula used in SSIM and many other metrics. To keep this text concise, we describe below only the mutual masking model, which performed the best in our tests. We encourage the reader to refer to the \href{https://www.cl.cam.ac.uk/research/rainbow/projects/colorvideovdp/}{supplementary materials} with the detailed description and analysis of all the models.

Contrast masking shows different characteristics for different spatial frequency bands and color channels. However, it was shown that for both luminance \cite{Daly1993} and chromatic channels \cite{Cass2009,Switkes_1988} masking characteristics can be unified if both the test and masker contrast are multiplied by the contrast sensitivity function $S_{b,c}(\pixcoord)$:
\begin{equation}
    C^{\prime}_{b,c,f}(\pixcoord) = C_{b,c,f}(\pixcoord)\,S_{b,c,f}(\pixcoord)\,.
    \label{eq:csf-normalization}
\end{equation}
In our case, contrast sensitivity is provided by the castleCSF model:
\begin{equation}
S_{b,c,f}(\pixcoord) = s_\ind{corr}\,s_{\ind{ch},c}\,S_c(\rho_b,\omega_c,L_{\ind{bkg},b,f}^\ind{ref}(\pixcoord))\,,
\end{equation}
where $\rho_b$ is the spatial frequency of band $b$ in cycles per degree, $\omega_c$ is the peak temporal frequency of channel $c$, and $L_\ind{bkg}^\ind{ref}(\pixcoord)$ is the local background luminance for the reference image (see \eqref{local-contrast}). $s_\ind{corr}$ is a trainable parameter adjusting the absolute sensitivity of the model (all trained parameters can be found in \tableref{cvvdp_params}). The additional benefit of the multiplicative contrast encoding from \eqref{csf-normalization} above is that it helps in matching the perceived magnitude of suprathreshold contrast across color modulation directions \cite{Switkes1999}, luminance \cite{Peli1995} and to a lesser extent across spatial frequencies (see the full analysis in the \href{https://www.cl.cam.ac.uk/research/rainbow/projects/colorvideovdp/}{supplementary}). However, we found that CSF alone is too inaccurate to match suprathreshold contrast across color modulation directions, and we had to introduce the color matching correction factor $s_{\ind{ch},c}$ based on the contrast matching data of \citeN{Switkes1999}: $s_{\ind{ch},c}=\begin{bmatrix}1 & 1 & 1.7 & 0.237\end{bmatrix}$ corresponding to achromatic sustained, transient, red-green and yellow-violet channels (see the full explanation in the \href{https://www.cl.cam.ac.uk/research/rainbow/projects/colorvideovdp/}{supplementary}). 

\begin{figure}
    \centering
    \includegraphics[width=\columnwidth]{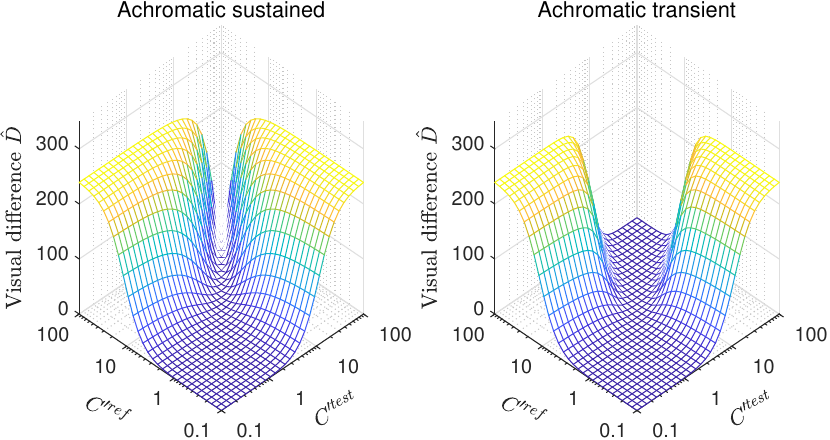}
    \caption{The visual difference for a pair of test and reference contrast values (normalized by the CSF), as predicted by the masking model (see \eqref{masking-model} and \eqref{mask-clamped}). The visual difference $D$ is zero when the test and reference contrasts are identical and increases as they start to differ. However, there is no increase for small contrast values due to contrast sensitivity.}
    \label{fig:masking-3d}
\end{figure}

\begin{figure}
    \centering
    \includegraphics[width=\columnwidth]{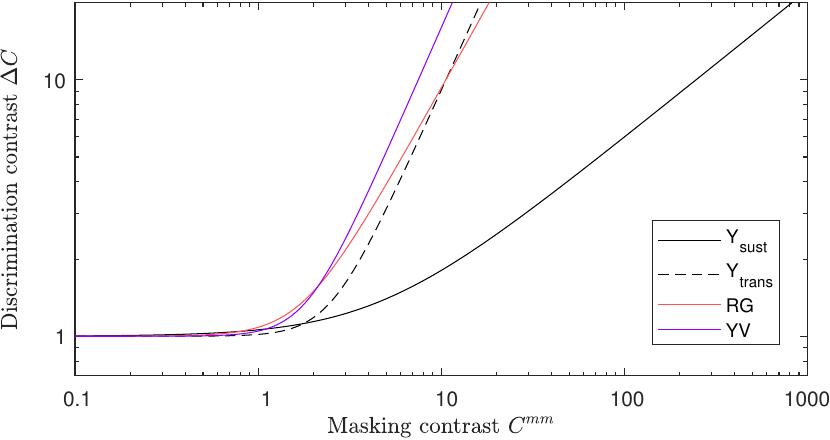}
    \caption{Contrast discrimination functions for our masking model for the sustained and transient channels. The value on the y-axis denotes the difference in the normalized contrast (${\Delta}C=\left|C^{\prime\,\testF}-C^{\prime\,\refF}\right|$) required to obtain the response $D=1$ (see \eqref{masking-model}). }
    \label{fig:masking-thresholds}
\end{figure}

\begin{figure}
    \centering
    \includegraphics[width=0.8\columnwidth]{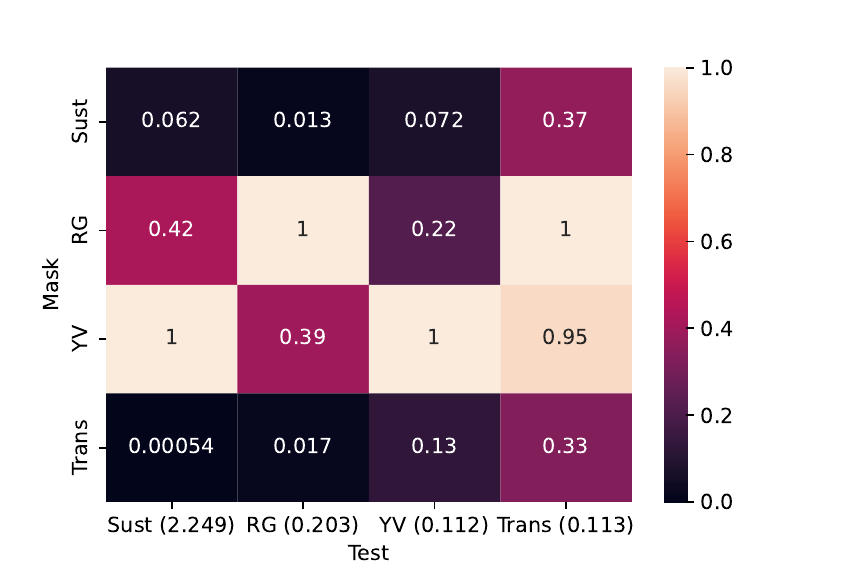}
    \caption{Weights of the cross-channel masking model --- $k_{i,c}$. The values are normalized to one in each column to better show the contribution of each channel. The absolute values can be obtained by multiplying the cell value by a constant shown in the parenthesis at the bottom of each column. The matrix shows that both chromatic channels strongly mask the sustained achromatic channel (1st column), but the achromatic channels do not mask chromatic channels.}
    \label{fig:xcm-weights}
\end{figure}

Once the encoded contrast $C^\prime$ is calculated, the visual difference between the bands is calculated as:
\begin{equation}
    D_{b,c,f}(\pixcoord)=\frac{\left|C^{\prime\,\testF}_{b,c,f}(\pixcoord)-C^{\prime\,\refF}_{b,c,f}(\pixcoord)\right|^p}{1 + C^\maskF_{b,c,f}(\pixcoord)}\,,
    \label{eq:masking-model}
\end{equation}
where $p$ is a parameter of the model. The masking signal $C^\maskF_{b,c}$ combines local contrast across test and reference images, local spatial neighborhood, and channels. First, similarly as in \cite[p.192]{Daly1993}, we calculate the mutual masking of test and reference bands (see also \figref{masking-model}):
\begin{equation}
    C^\text{mm}_{b,c,f}(\pixcoord) = \min\left\{\left|C^{\prime\,\testF}_{b,c,f}(\pixcoord)\right|,\left|C^{\prime,\refF}_{b,c,f}(\pixcoord)\right|\right\}\,.
    \label{eq:mutual-masking}
\end{equation}
Then, the mutual masking signal is pooled in a small local neighborhood by convolving with a Gaussian kernel $g_{\sigma_\ind{sp}}$, and is combined across channels, accounting for the cross-channel masking: 
\begin{equation}
    C^\maskF_{b,c,f}(\pixcoord) = \sum_i k_{i,c} ((C^\ind{mm}_{b,i,f})^{q_c} * g_{\sigma_\ind{sp}})(\pixcoord)  \,.
    \label{eq:mask-min}
\end{equation}
where $k_{i,c}$ is the cross-channel masking coefficient, describing the contribution of channel $i$ to the masking signal of channel $c$. The weights for the trained model are shown in \figref{xcm-weights}. The exponents $q_c$ are model parameters set separately for each channel (two achromatic and two chromatic channels). One interpretation of \eqref{masking-model} is that the division by the absolute amplitude of the neighbors reduces redundancy in natural images since neighboring pixels are highly correlated \cite{Laparra_Simoncelli_2016}. Such normalization was shown to provide encoding that better correlates with our perception of differences.

The (low-pass) baseband of the Laplacian pyramid cannot be used to calculate contrast, as done in \eqref{local-contrast}. Instead, the differences in the baseband are computed directly on the Gaussian pyramid coefficients, which are multiplied by the sensitivity: 
\begin{equation}
    D_{B,c,f}(\pixcoord) = k_{\ind{B},c}\,\left| \gausspyr_{B,c,f}^\ind{test}(\pixcoord) - \gausspyr_{B,c,f}^\ind{ref}(\pixcoord)\right|\,S_{B,c,f}(\pixcoord)\,,
\end{equation}
where $B$ is the index of the baseband. Because baseband differences are in different units than those in other bands, we need to introduce a trainable scaling factor $k_{\ind{B},c}$, which varies across the channels.

One limitation of the original mutual masking model is that it results in excessively large contrast values when the sensitivity is high, and there is no masking signal (refer to the \href{https://www.cl.cam.ac.uk/research/rainbow/projects/colorvideovdp/}{supplementary}). We found it essential to restrict the maximum contrast in each band so that a few very large differences do not have an overwhelming impact on the predicted quality. We achieve that with a soft-clamping function
\begin{equation}
    \hat{D}_{B,c,f}(\pixcoord) = \frac{k_\ind{C}\,D_{B,c,f}(\pixcoord)}{k_\ind{C} + D_{B,c,f}(\pixcoord)}\,,
    \label{eq:mask-clamped}
\end{equation}
where $k_\ind{C}$ is a trainable parameter. The equation accounts for the limited dynamic range of the retinal cells, which cannot encode large contrast values.

The shape of the masking function for the sustained and transient achromatic channels is plotted in \figref{masking-3d}. Another visualization of the masking model is shown \figref{masking-thresholds} as a discrimination vs. masking plot. The plot shows that, as the mutual masking contrast $C^\ind{mm}$ increases, a higher difference between the test and reference bands is needed to trigger the same response ($D_{b,c,f}(\pixcoord)=1$ in this example). Such an increase is more gradual for the sustained achromatic channel and chromatic channels and more abrupt for the transient channel. The findings of Switkes et al. \shortcite{Switkes_1988} indicate that color can robustly mask luminance patterns, while luminance does not mask color but instead facilitates the detection of color patterns. Remarkably, the cross-channel masking weights obtained in metric fitting (\secref{evaluation}) and shown in \figref{xcm-weights} confirm these findings --- weights are very small in the 2nd and 3rd column of the top row, showing little influence of luminance (achromatic sustained channel) on masking of the color channels. Note that our mutual masking model cannot account for facilitation. An example of this effect is shown in \figref{masking-example}, in which luminance contrast does not mask red-green chromatic contrast.

\subsection{Pooling}
Once the differences between the bands, $D_{b,c,f}$, are computed, they need to be pooled into a single quality correlate. We follow a similar strategy as done in FovVideoVDP and pool the differences across all the spatial dimensions ($\pixcoord$) in each band, across spatial frequency bands ($b$), across the channels ($c$), and finally across all the frames ($f$):
\begin{equation}
    D_\textrm{pooled} = \frac{1}{F^{\nicefrac{1}{\beta_f}}}\left\Vert \left\Vert w_c \left\Vert \frac{1}{N_{b}^{\nicefrac{1}{\beta_x}}} \left\Vert \hat{D}_{b,c,f}(\pixcoord) \right\Vert_{\beta_x,\pixcoord} \right\Vert_{\beta_b,b} \right\Vert_{\beta_c,c} \right\Vert_{\beta_f,f}\,,
    \label{eq:pooling}
\end{equation}
where $\left\Vert\cdot\right\Vert_{p,v}$ is a $p$-norm over the variable $v$:
\begin{equation}
\left\Vert f(v) \right\Vert_{p,v} = \left( \sum_{v} \vert f(v) \vert^p \right)^{\nicefrac{1}{p}}\,.
\end{equation}
$w_c$ is the weight associated with each channel.
The main advantage of pooling differences first across pixels is that we avoid the expensive step of the synthesis of the distortion map from the multiple levels of the Laplacian pyramid. It should be noted that the differences across pixels are normalized by the number of pixels in each band, $N_b$ --- we do not want the bands represented with more pixels in the pyramid to contribute more to image quality. 
The pooling exponents $\beta_x$ and $\beta_b$ were optimized in FovVideoVDP, but we found that such an optimization is unstable (because of exploding gradients) and unnecessary. Instead, we set $\beta_x=2$ and $\beta_f=2$ to represent the energy summation across the spatial dimensions and frames \cite{Watson1983}. $\beta_b$ and $\beta_c$ are set to 4, which is representative of the summation across channels \cite{Quick1978}. Using exponents greater than 1 roughly corresponds to a "winner-take-all" strategy, in which the strongest visual differences have the highest impact on the pooled value.

\subsection{JOD regression}

\begin{figure}
    \centering
    \includegraphics[width=\columnwidth]{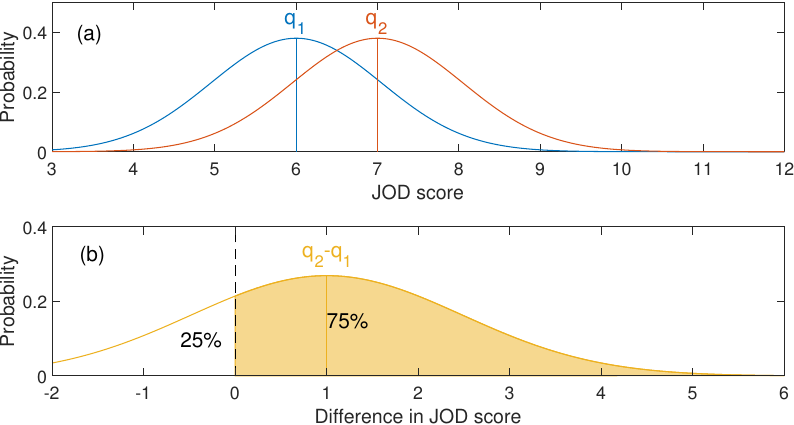}
    \caption{Interpretation of the JOD scores. (a) JOD quality scores are normally distributed random variables --- the reported JOD value is the mean of the distribution, which models that a certain portion of the population will rate the quality as higher or lower than the reported mean value. (b) The standard deviation of the distribution is selected in such a way that when the difference between the  means is 1\,JOD ($q_2-q_1=1$), 75\% of the population will select the condition with the higher JOD value as better. This allows us to interpret quality differences expressed in JOD units.}
    \label{fig:jod-illustration}
\end{figure}

The visual difference correlate is regressed into interpretable units of just-objectionable-differences (JODs), using the same formula as FovVideoVDP: 
\begin{equation}
Q_\textrm{JOD} = 10 -\alpha_\textrm{JOD}\,(D_\textrm{pooled})^{\beta_\textrm{JOD}}\,,
\label{eq:jod-regression}
\end{equation}
where $\alpha_\textrm{JOD}$ and $\beta_\textrm{JOD}$ are tuned parameters. Here, 10\,JOD represents the highest quality --- when test and reference images are identical. The JOD units are scaled in terms of inter-observer variance --- the drop in quality of 1\,JOD means that 75\% of observers will notice such a loss of quality in a pairwise comparison experiment. This concept is illustrated in \figref{jod-illustration}.


\subsection{Image quality}

When predicting the quality of images, we use the same processing stages as for video but with two changes. First, we skip temporal decomposition and do not create the achromatic transient channel --- all computations are performed on three channels. Second, we replace the temporal pooling from \eqref{pooling} with a single, trainable constant $k_\ind{I}$:
\begin{equation}
D_\textrm{pooled}^\textrm{image} = k_\ind{I} \left\Vert w_c \left\Vert \frac{1}{N_{b}^{\nicefrac{1}{\beta_x}}} \left\Vert \hat{D}_{b,c,f}(\pixcoord) \right\Vert_{\beta_x,\pixcoord} \right\Vert_{\beta_b,b} \right\Vert_{\beta_c,c}\,.
    \label{eq:image-pooling}
\end{equation}
$k_\ind{I}$ could be interpreted as a fixation time. It lets us unify the quality between images and video.  

\subsection{Visualization --- heatmaps and distograms}

\begin{figure}
    \centering
    \includegraphics[width=\columnwidth]{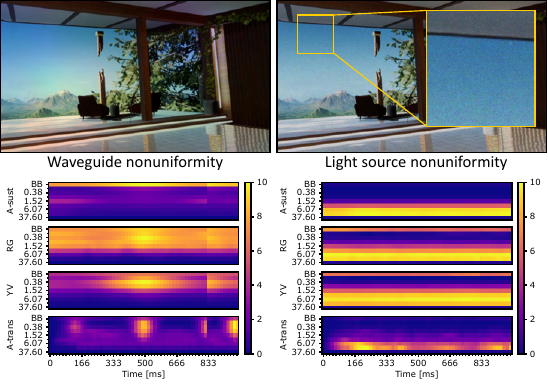}
    \caption{Examples of \emph{distogram} visualization of artifacts. The values on the y-axis denote spatial frequency in cpd. The higher (more yellowish) values denote stronger distortion. The low-frequency waveguide nonuniformity (left) is shown to be visible mostly in the base band (BB) and low frequency bands of both achromatic (A-sust) and chromatic (RG, YV) channels. The artifacts shows also as low-frequency flicker in the achromatic transient channel (A-trans). The (per-pixel) light source nonuniformity causes high-frequency shimmer, which is visible as high frequency distortions in the distogram on the right.}
    \label{fig:distogram-examples}
\end{figure}

\ourmethod{} offers two types of visualization that help to interpret the quality score. We can overlay a heatmap with per-pixel distortion intensity over the grayscale version of distorted content, as shown in \figref{teaser}. To obtain such a per-pixel distortion map, we pool the distortions across the channels
\begin{equation}
    H_{b}(\pixcoord) = \left\Vert w_c \hat{D}_{b,c,f}(\pixcoord) \right\Vert_{\beta_c,c} 
    \label{eq:heatmap-pooling}    
\end{equation}
and perform the Laplacian pyramid synthesis step to obtain the map with visual difference correlates per pixel and per frame. Those can then be transformed into the JOD units using \eqref{jod-regression}. It must be noted that the metric has not been trained to produce accurate heatmaps as there are no datasets that could be used for that purpose (such difference maps have been collected for images \cite{Ye2019} but not for video). The heatmaps are meant to help interpret the single-valued JOD quality predictions.

Another visualization, which we name a \emph{distogram}, lets us present video distortions within each channel and spatial frequency band, all in a single diagram. Two examples for two distortions from the \ourdataset{} dataset are shown in  \figref{distogram-examples}. The examples show that depending on the characteristics of the distortion, the artifacts will show in either low- or high-frequency bands, in achromatic or chromatic channels. This visualization decomposes the distortions into their frequency components, visual channels, and their time series, which helps interpret and explain how those distortions contributed to the final JOD score. 

\subsection{Implementation details}

\ourmethod{} is implemented in PyTorch and optimized for fast execution on a GPU. To take advantage of the massive parallelism offered by the GPU, we load as many video frames as we can process into GPU memory. Then, the operations are executed in parallel on a set of test and reference frames, all four channels, and all pixels (all stored in a single tensor); only the pyramid bands are processed sequentially as each one has a different resolution. To avoid the computational overhead of castleCSF, the function is precomputed and stored as a set of 2D look-up-tables (LUTs) of luminance and spatial frequency, with a separate LUT for each channel (see \figref{castleCSF}). Thanks to these techniques, \ourmethod{} has processing times comparable to the state-of-the-art metrics, while relying on a much more complex visual model. The timings can be found in the \href{https://www.cl.cam.ac.uk/research/rainbow/projects/colorvideovdp/}{supplementary document}. \ourmethod{} is also fully differentiable, which lets us use it as an objective function in optimization, or calibrate its parameters, as explained in \secref{evaluation}.

\section{XR Display Artifact Video Dataset (XR-DAVID)}
\label{sec:xr-david}

Calibrating a metric to real subjective data is an extremely important step. Notably, the datasets used for calibration must be representative of the types of artifacts that users will be applying the metric to. This can be challenging in cases where there is little prior art, for instance, while traditional artifacts like image and video compression are relatively well-studied, artifacts stemming from novel XR display architectures like waveguide nonuniformity are not. 

To address this gap in quantitative data for XR distortions, we enumerated the most relevant artifacts for our study. We placed special emphasis on distortions that have a color or temporal component, as these are not well served by existing metrics and are the focus of \ourmethod{}. We used this dataset of distortions to conduct a large-scale subjective study, collecting JOD quality scores in a new \ourdataset{} video quality dataset\footnote{The XR-DAVID dataset can be downloaded from \url{https://doi.org/10.17863/CAM.108210}}, which is suitable for calibration of \ourmethod{}. 




\begin{figure}
    \centering
    \includegraphics[width=0.6\linewidth]{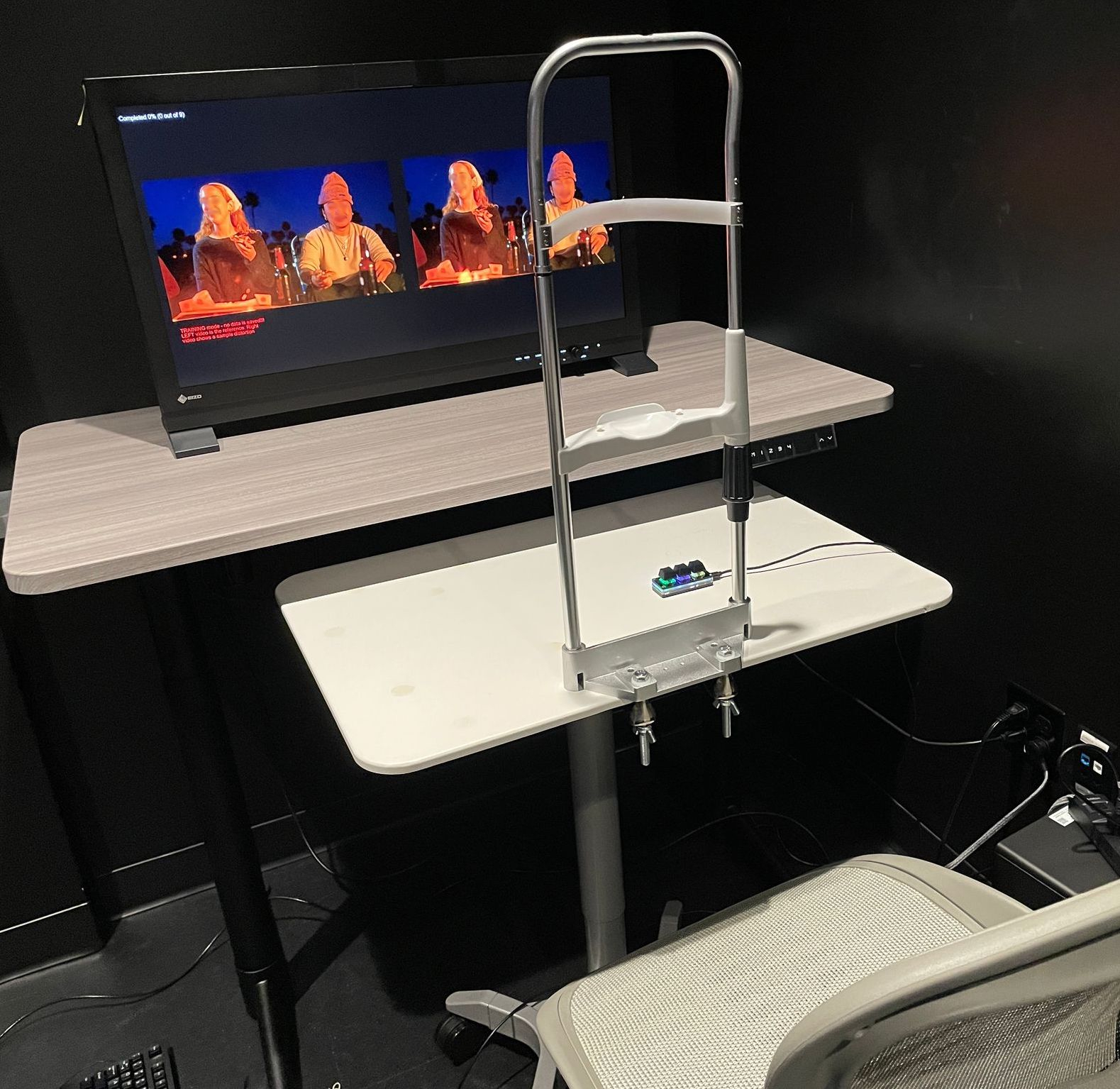}
    \caption{Physical hardware setup for the \ourdataset{} data collection. Participants used a chinrest to fix their distance from the display at 0.73\,m (two display heights) and used a 3-key keyboard to record their choices.}
    \label{fig:color-quality-study-info}
\end{figure}

\subsection{Experimental setup}
\paragraph{Physical Setting} We selected an Eizo CG3146 professional reference monitor\footnote{Eizo CG3146, see \url{https://www.eizo.com/products/coloredge/cg3146/} for detail} as our test vehicle. This 31.1" diagonal display has a resolution of $4096\times2160$, and a contrast of 1,000,000:1 claimed by the manufacturer.
The monitor was set to a maximum luminance of 300\cdms, with sRGB EOTF, P3 color primaries, and a 60\,Hz refresh rate. This display has a built-in colorimeter, which allowed for high-precision daily calibration to ensure accuracy as the study progressed. The distance of the observer was controlled using a chin rest, which was placed so that the effective resolution of the display was 77
pixels per visual degree (see \figref{color-quality-study-info}). 


\paragraph{Participants} We conducted the study using paid external participants. After two pilots (N=5 each), 77 naïve users took part in 1 hour long sessions of our main study, which included training, data gathering, and a short break half-way through.
All participants signed informed consent forms, and the study was approved by an external ethics committee. Participants were screened via an Ishihara color test~\cite{hardy1945tests} to ensure normal color vision.

\paragraph{Reference videos} We selected 14 high-quality videos following practical considerations. Selected thumbnails of scenes used in the study are shown in the \href{https://www.cl.cam.ac.uk/research/rainbow/projects/colorvideovdp/}{supplementary document}. Our references included videos spanning real, rendered, and productivity content, which are typical for AR/VR applications.

\paragraph{Experimental procedure} Participants performed a side-by-side pairwise comparison task answering which of the two versions of a video was less distorted. This method was preferred to alternatives (e.g., direct ratings on the mean-opinion-score scale) as pairwise comparisons have been shown to produce more accurate results~\cite{zerman2018relation} by simplifying the task in each trial. Active sampling using the \emph{ASAP} method \cite{mikhailiuk2021active} was used to optimize information gain from each comparison, making it possible to explore a large number of artifacts without prohibitively increasing the number of pairwise comparisons. We allowed comparisons across different distortion types and levels but not across different video contents. A training session preceded the main experiment. In that session, the participants compared each reference video to its distorted version. A different distortion type was used in each trial. The training session was meant to familiarize the participants with the reference content and distortions.
The results of the main experiment were scaled to a unified perceptual scale in just-objectionable-difference units (JODs) using the \emph{pwcmp} software suite \cite{perez2019pairwise}. 

\subsection{Distortions} 
Each reference video was distorted by one of 8 artifacts, each of which was applied at one of 3 different strengths. All 8 distortions at level 3 (most distorted) are illustrated in Figure~\ref{fig:allartifacts}. We discuss each artifact in detail below.

\paragraph{Spatiotemporal dithering}
Display systems could be constrained such that bit depths of color channels are reduced. In such scenarios, dithering could be performed both spatially and temporally to improve image quality. This artifact simulates blue noise mask dithering per color channel and per frame for a given bit depth. The bit depths were 6, 5, and 4 for artifact levels 1, 2, and 3, respectively. 

\paragraph{Light source nonuniformity (LSNU)}
Display light sources like $\mu$OLED and $\mu$LED tend to exhibit high-frequency spatial nonuniformity, as each pixel consists of a separate light source module, which may vary in its light output. This artifact was simulated by assuming different levels of variations in pixel intensities for each color channel, resulting in spatial and color artifacts. 
Each pixel intensity was randomly modulated per color channel by up to 12\%, 16\%, and 18\% (values corresponding to twice the standard deviation, and half the limit at which truncation occurred) using a Gaussian distribution for artifact levels 1, 2, and 3, respectively. The modulation ratios were kept constant for each pixel across all frames. Simulations were done in linear color space.

\begin{figure}
    \centering
    \includegraphics[width=\linewidth]{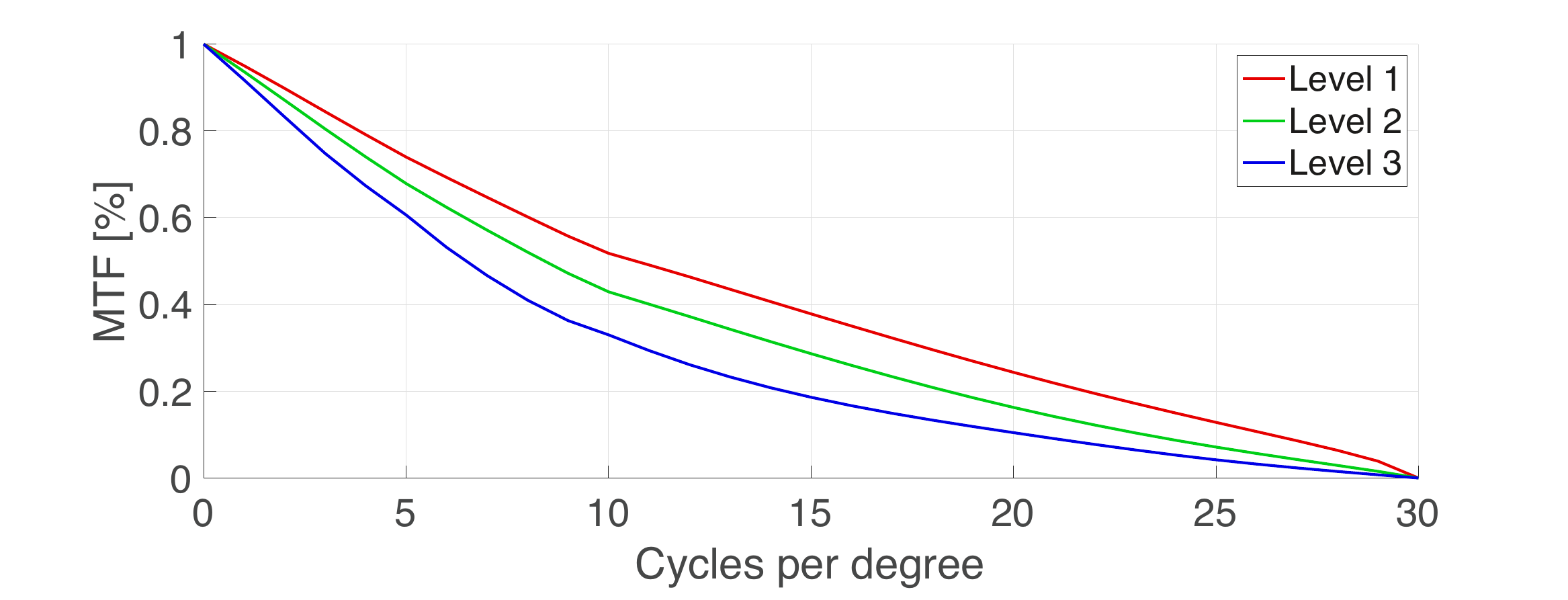}
    \caption{MTF Curves corresponding to the \emph{Blur} artifact at level 1, 2, and 3.}
    \label{fig:MTFCurves_wide}
\end{figure}

\paragraph{Blur (MTF degradation)}
Optical components such as lenses and waveguides in AR and VR displays will degrade the system MTF, resulting in blurry images. The artifact was simulated by applying a representative point spread function (PSF) with various severities as illustrated in \figref{MTFCurves_wide}. The same PSF was applied to all color channels. In addition to varying PSFs, the lateral chromatic aberration was simulated for artifact level 2 and 3 by shrinking the green channel frame by two pixels. 

\paragraph{Reduced contrast}
It is often challenging to achieve good contrast in optical see-through displays, particularly in bright environments. 
As a first-order approximation, the artifact was simulated in linear space such that contrast was reduced by increasing the minimum value of input videos while maintaining the maximum value as is. The contrast, defined as max/min, was reduced to 100:1, 50:1, and 30:1 for artifact levels 1, 2, and 3 respectively. 

\begin{figure}
    \centering
    \includegraphics[width=\linewidth]{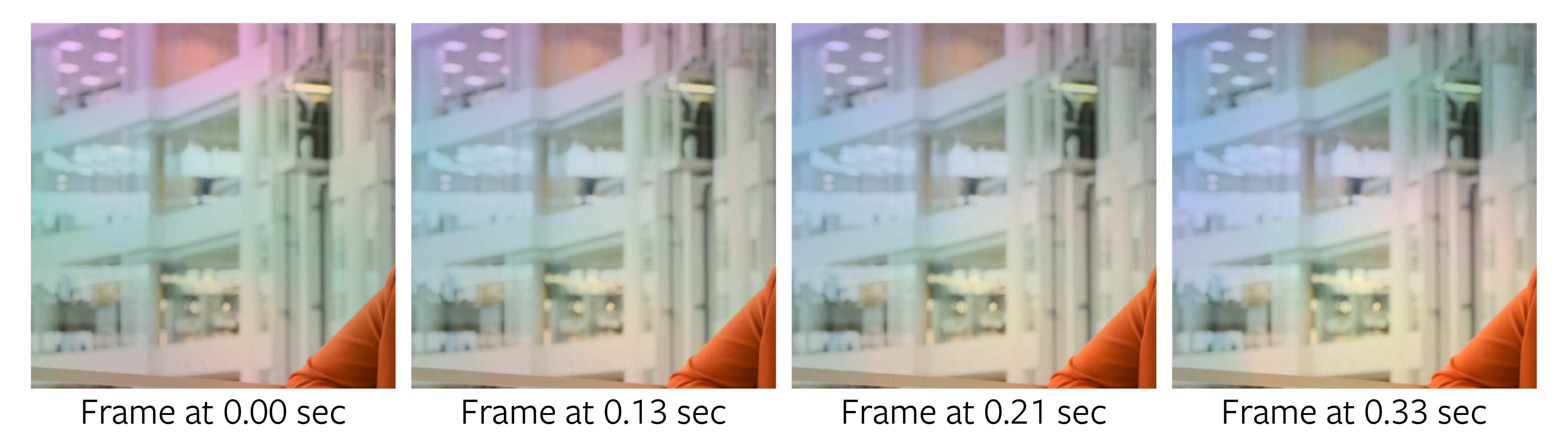}
    \caption{Examples of waveguide nonuniformity level 3 at different frames, using \emph{Business} scene (cropped for better visibility). nonuniformity changes from one pattern (left) to another (right).}
    \label{fig:Business_WGNU_overtime}
\end{figure}

 
\paragraph{Waveguide nonuniformity (WGNU)}
AR displays with diffractive waveguides such as Microsoft Hololens, Magic Leap, and Snap Spectacles \cite{ooi2022color} exhibit a characteristic spatially-varying color nonuniformity. This nonuniformity is typically low-frequency (less than ~1 cycle per degree). In addition, it is heavily dependent on pupil position within the eye box, which can vary depending on the user's fixation and eye movements. In particular, if a user keeps their gaze fixed on an object in the AR content while moving their head (engaged in what is typically termed a vestibulo-ocular reflex or VOR movement), they may notice its color shifting as the position of the eye changes in relation to the waveguide. In order to simulate the artifact, first, two nonuniformity patterns were obtained from empirical waveguide transmission data. Second, base variation maps were generated for the two patterns by cleaning noise in data and normalizing variations across channels. The variations were also scaled such that they appear to have similar magnitude across the two patterns. Finally, the amplitude of variations in the base map was varied to produce different artifact levels. The multipliers of the amplitude variations were 0.15, 0.3, and 0.4 for artifact levels 1, 2, and 3 respectively. To simulate changes with eye position, the nonuniformity pattern transitions from the first base pattern to the second and back over the course of the video. This transition is repeated 4 times for level 2, and 7 times for level 3. \figref{Business_WGNU_overtime} illustrates this temporally varying waveguide nonuniformity artifact. 


\paragraph{Dynamic correction error (DCE)}
In AR displays suffering from waveguide nonuniformity across different exit pupil positions, as described above, a dynamic correction algorithm can be implemented to invert the distortion, reducing spatial nonuniformity. This algorithm would be dependent on gaze location, thus relying on estimated positions from eye trackers, which may be inaccurate. Namely, inaccurate eye tracker readings will cause a wrong estimate of the pupil position and temporal color artifacts (by either not correcting perfectly for distortions or performing an imprecise correction for a distortion pattern that is not present).
To simulate this, the precision error was randomly generated per frame from a Gaussian distribution with a standard deviation of 0.27\textdegree, 0.39\textdegree, and 0.58\textdegree~for artifact levels 1, 2, and 3 respectively. Simulated precision errors are in alignment with existing VR headsets currently in the market \cite{schuetz2022eye}.




\paragraph{Color fringes}
An image's color channels may be optically misaligned due to various reasons such as mechanical shifts of optical components in head-mounted displays, imperfect fabrication processes, and thermal loads. As a first-order approximation, the artifact was simulated by shifting each color channel by variable amounts in 2D dimensions globally. 
For artifact level 1, the red channel frame was shifted by [0, 0.5] pixels in [x,y] direction while the blue channel frame was shifted by [-0.5, 0] pixels. 
For artifact level 2, the red channel frame was shifted by [1, 0.5] pixels while the blue channel frame was shifted by [0.5,-1] pixels. 
For artifact level 3, the red channel frame was shifted by [1.5, 1] pixels while the blue channel frame was shifted by [1, -1] pixels. 
The green channel frame was kept unshifted for all the artifact levels. 

\paragraph{Chroma subsampling}
Chroma subsampling is a common technique used in video compression and signal transmission to reduce spatial resolutions of chroma channels while maintaining the overall image quality. The technique takes advantage of a much-reduced sensitivity of the visual system to high frequencies modulated along chromatic (isoluminant) color directions. Spatial resolutions of chroma channels were reduced by 1/4,  1/8, and 1/12 for artifact level 1, 2, and 3 respectively. 

\begin{figure}
    \centering
    \includegraphics[width=0.9\linewidth]{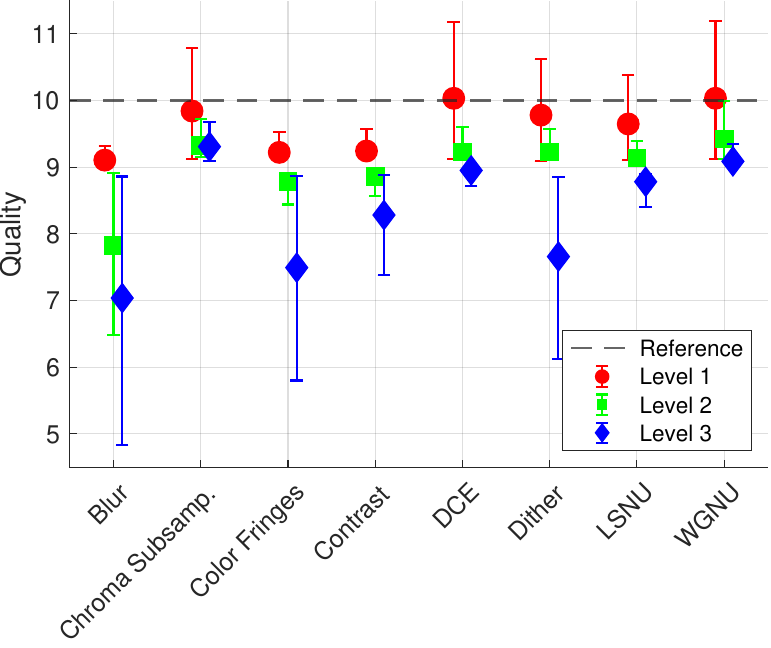}
    \caption{The results of our display artifact study aggregated over all users and scenes are shown. Quality scores are expressed in JODs, counting down from 10 by convention. Values above the nominal reference are the result of chance selection for artifacts that appear very subtle. Error bars represent 95\% confidence intervals, obtained via bootstrapping.}
    \label{fig:color-quality-study-aggregate}
\end{figure}

\paragraph{Results} The results of our study are shown in \figref{color-quality-study-aggregate} in aggregate form over all scenes, and in full in Figure~2
in the \href{https://www.cl.cam.ac.uk/research/rainbow/projects/colorvideovdp/}{supplementary}. \figref{color-quality-study-aggregate} shows a good range of quality levels, with some distorted videos almost indistinguishable from the reference. Such conditions are useful to test whether a metric can disregard invisible distortions. Having all artifacts graded on a single linear perceptual JOD scale allows for direct comparison, and reveals interesting interactions between individual videos and artifacts types. 
For instance, while blur was graded as very disturbing across all forms of content, waveguide nonuniformity was fairly visible for a web-browsing scenario ('wiki'), but almost invisible in a dark scene ('bonfire') (see Figure~14 in the \href{https://www.cl.cam.ac.uk/research/rainbow/projects/colorvideovdp/}{supplementary}). 


\section{Metric validation}
\label{sec:evaluation}

\begin{table*}[]
    \centering
    \caption{Datasets used for training and testing.}
    \label{tab:datasets}
    \begin{tabular}{c|c|c|c|c|l}
        \toprule    
         Dataset & Used for & Type & Scenes & Conditions & Distortions \\
        \midrule    
        \midrule    
         \ourdataset{} & Train \& test & SDR video & 14 & 336  & 8 display artifacts \\
         UPIQ \cite{Mikhailiuk2021} & Train \& test & SDR/HDR images & 84 & 4159 & 34 distortion types \\
         LIVE HDR \cite{Shang2022} & Test & HDR video & 21 & 210 & H.265, bicubic upscaling \\      
         LIVE VQA \cite{Seshadrinathan2010} & Test & HDR video & 10 & 150 & H.264, MPEG-2, transmission \\      
         KADID-10k \cite{Lin2019} & Test & SDR images & 81 & 10125 & 25 distortion types\\      
        \bottomrule         
    \end{tabular}
\end{table*}



A typical validation of a quality metric involves reporting correlation values for each individual dataset. While we still report such correlation values (in the supplementary \href{https://www.cl.cam.ac.uk/research/rainbow/projects/colorvideovdp/}{HTML reports}), here, we undertake a more challenging task of predicting absolute quality in JOD units, which could generalize across datasets scaled in such units. Therefore, unlike most work in this area, we do not train \ourmethod{} individually for each dataset, but instead, we train a single version of the metric on multiple datasets with the goal of generalizing to new (unseen) data. 

\paragraph{Datasets} We split the datasets into those used solely for testing and those used for both training and testing. The five selected datasets are listed in \tableref{datasets}. The two datasets used for both training and testing are \ourdataset{}, explained in detail in \secref{xr-david}, and UPIQ \cite{Mikhailiuk2021}. We selected the UPIQ dataset because it contains a large collection of both SDR and HDR images (over 4000) and is scaled in JOD units, similar to \ourdataset{}. To test our metric on unseen datasets (cross-dataset validation), we chose LIVE~HDR because it is a modern dataset representative of video streaming applications, KADID-10k because of its size (over 10k images), and LIVE~VQA because it is widely used to test video quality metrics. Because KADID-10k and a part of UPIQ were collected in uncontrolled crowdsourcing experiments, we had to approximate the display specification and viewing conditions for those. We avoided video datasets that were collected in uncontrolled conditions (e.g., KonViD-1k).


\paragraph{Training and testing sets} The two datasets used for both training and testing were split into 7 parts: 5 parts were used for training, and 2 parts were used for testing. Each scene is present in only one part so that no scene is shared across the sets. 7 parts were selected because 7 is the common denominator for the number of unique scenes in \ourdataset{} and LIVE~HDR (see \tableref{datasets}). 

\paragraph{Training} Training a video quality metric is problematic as a very large amount of data is used to predict a single quality value. For example, 60 frames of 4K video requires the processing of almost 500 million pixels to infer just a single quality score. This makes gradient computation (through backpropagation) infeasible because of the memory requirements. Previous work dealt with this problem in several ways: some metrics were designed to extract low dimensional features from a video and then train a regression mapping those features to quality scores \cite{vmaf1}. This approach, however, does not allow training or tuning the feature extraction stage. Another group of methods operated on patches of limited resolution (e.g. 64$\times$64), assuming that the quality is the same across the entire image \cite{Prashnani2018}. This assumption, however, can be easily proved wrong for localized distortions or for content in which the effect of contrast masking varies across an image. Other metrics used numerical gradient computation \cite{Mantiuk2021}, which, however, becomes too expensive when optimizing a large number of parameters. 

We used a mixture of feature-based and end-to-end training. The parameters that are introduced after pooling across all pixels in a frame (such as JOD regression parameters) can be easily optimized by pre-computing pooled values/features ($\left\Vert \hat{D}_{b,c,f}(\pixcoord) \right\Vert_{\beta_x,\pixcoord}$ from \eqref{pooling}) and then optimizing the stages of the metric that follow the pooling stage.
This approach not only significantly accelerates the training process due to the reduced memory footprint of pooled features compared to full videos, but it also lets us operate on much larger batches. We found that large batches, with smoother gradients, are required for stable training of the pooling and regression parameters.

For the end-to-end training, we computed the full analytical gradient of the remaining parameters using two techniques. Firstly, we utilized gradient checkpointing \cite{chen2016checkpointing} in PyTorch, by rerunning a forward pass for each checkpointed segment during backward propagation. This technique trades some speed for reduced memory requirements. We inserted a checkpoint after each block of frames, where the block size was determined based on the available GPU memory. Secondly, during training, we randomly sampled 0.5-second-long sequences from each video clip (full sequences were used for testing). This approach improved training time and introduced a form of data augmentation. The feature-based and end-to-end training were run in an interleaved manner, with 50 epochs of (fast) feature-based training followed by a single epoch of end-to-end training. This lets us jointly train all the parameters of the metric. The trained parameters of the metric can be found in \tableref{cvvdp_params}.

\begin{table}[]
    \caption{The trained parameters of the \ourmethod{}. The parameters of the cross-channel masking can be found in \figref{xcm-weights}.}
    \label{tab:cvvdp_params}
    \centering
    \begin{tabular}{p{25mm}<{\raggedleft}p{55mm}}
    \toprule    
Model component & Parameters \\
Contrast sensitivity & $S_\ind{corr}=0.9683$ \\
Masking & $p=2.264$, $q_c=[1.303, 2.889, 3.681, 3.589]$, $k_\ind{C}=366.6$, $\sigma_\ind{sp}=3$, \\
& $k_{\ind{B},c}=[0.003633, 1.663, 4.119, 25.26]$ \\
Pooling & $k_\ind{I}=0.5779$, $w_c=[1, 1, 1, 0.8081]$ \\
JOD regression & $\alpha_\ind{JOD}=0.04396$, $\beta_\ind{JOD}=0.9302$ \\
\bottomrule    
    \end{tabular}
\end{table}

\subsection{Comparison with other metrics}
We compare the performance of \ourmethod{} with several state-of-the-art metrics, listed in \tableref{metric-table}. 

As our datasets include both SDR and HDR content, we need to ensure that those are handled correctly by all the metrics. 
For metrics that do not work with colorimetric data and instead operate on display-encoded (SDR) pixel values, we ran the evaluations using the original SDR pixel values. In the case of HDR content, we employed the perceptually-uniform transform (PU21) \cite{Mantiuk2021pu} to encode the pixel values.
The metrics that operate on colorimetric data (FovVideoVDP, HDR-VDP-3) were supplied the absolute luminance values, computed by the display model (\secref{display-model}). The metrics that account for the viewing distance or the resolution in pixels-per-degree (e.g., HDR-VDP-3, HDR-FLIP) were supplied with the correct values.

Because UPIQ, \ourdataset{} and LIVEHDR are scaled in the same JOD units, we fit a single logistic function to map metric prediction to JODs. As neither KADID-10k nor LIVEVQA is scaled in JOD units, we fit a logistic function separately for each metric and dataset to map metric prediction to the subjective scores. 

\begin{figure}
    \centering
    \includegraphics[width=\columnwidth]{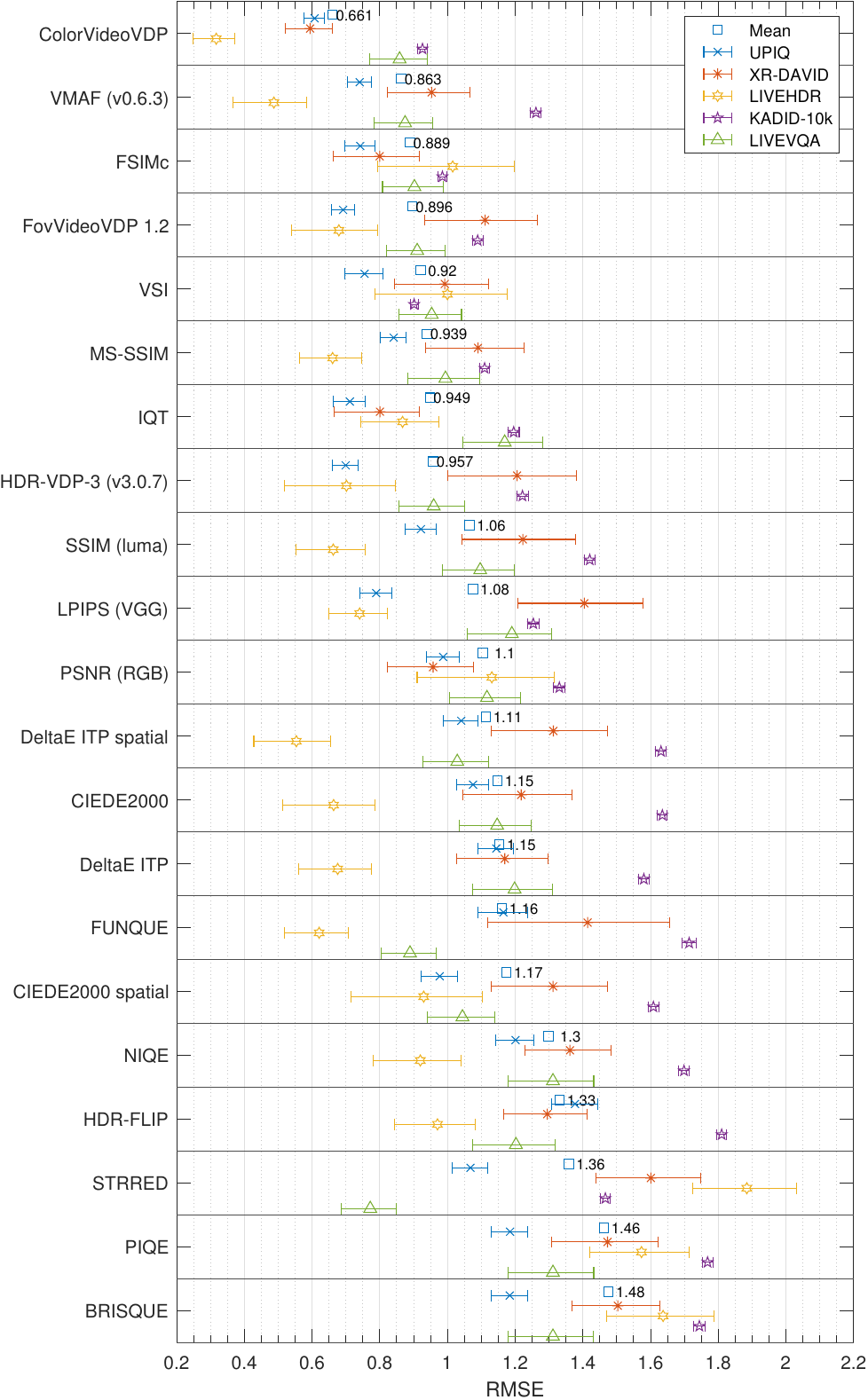}
    \caption{\ourmethod{} compared with existing metrics, listed in \tableref{metric-table}. The prediction errors are reported for testing datasets (LIVEVQA, LIVEHDR and KADID) and the test portion of UPIQ and \ourdataset{}. The error bars denote 95\% confidence intervals. Other performance metrics (SROCC and PLCC) and detailed reports can be found in the \href{https://www.cl.cam.ac.uk/research/rainbow/projects/colorvideovdp/}{supplementary HTML report}.}
    \label{fig:benchmark}
\end{figure}

The results, shown in \figref{benchmark}, indicate a substantial gain in performance of \ourmethod{} over the second-best metric --- VMAF. Image metrics that consider color --- FSIMc and VSI --- performed better than expected on \ourdataset{}, but rather poorly on LIVEHDR video dataset. The color difference formulas and their spatial extensions performed worse than image and quality metrics. The non-reference metrics --- NIQE, PIQE, and BRISQUE --- show a very small correlation with subjective judgments. Other performance indices (SROCC and PLCC) and detailed results can be found in the HTML reports included in the \href{https://www.cl.cam.ac.uk/research/rainbow/projects/colorvideovdp/}{supplementary}. Overall, while it is possible to find a metric that performs well on a selected dataset, such as STRRED on LIVEVQA (LIVEVQA was used to calibrate and test STRRED), \ourmethod{} offers good performance across all datasets with a wide variety of distortions and content. This could be explained by no other metric offering the same set of capabilities as \ourmethod{}; only our metric models spatiotemporal color vision and accounts for the display model.

\subsection{Ablations}
\label{sec:ablations}

\begin{figure}
    \centering
    \includegraphics[width=\columnwidth]{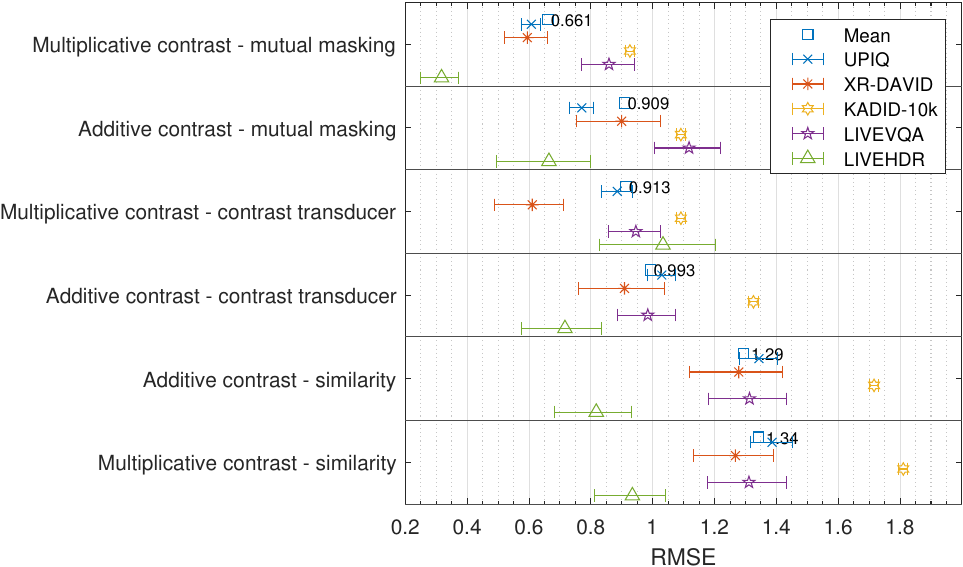}
    \caption{Ablation studies on the contrast encoding and masking models used in \ourmethod{}. The notation is the same as in \figref{benchmark}. The masking models used in the ablations are explained in the \href{https://www.cl.cam.ac.uk/research/rainbow/projects/colorvideovdp/}{supplementary}.}
    \label{fig:masking-ablations}
\end{figure}

We grouped ablation studies into those used to determine suitable contrast encoding and masking models, and into those used to test the importance of each component of the metric. The metric parameters were refitted for each ablation, as explained above. First, we tested a combination of two contrast encodings and three masking functions, all explained in detail in the \href{https://www.cl.cam.ac.uk/research/rainbow/projects/colorvideovdp/}{supplementary document}. The results of those ablations, shown in \figref{masking-ablations}, clearly indicate that the mutual masking model with multiplicative contrast encoding offers much better performance than the alternatives. The results indicate that the multiplicative contrast encoding is necessary for mutual masking and transducer models. As mentioned in \secref{xcm}, such encoding can unify results across different spatial frequencies \cite{Daly1993} and color directions \cite{Cass2009,Switkes_1988}. Although the contrast transducer is one of the best-established masking models \cite{Watson1997}, which accounts for facilitation and performs well on selected datasets \cite{Alam_Vilankar_Field_Chandler_2014}, it did not perform well as a part of the quality metric. The similarity formula exhibits masking properties and is used in many metrics, such as (MS-)SSIM \cite{Wang2003d}, but it did not result in acceptable performance when integrated into our metric.

\begin{figure}
    \centering
    \includegraphics[width=\columnwidth]{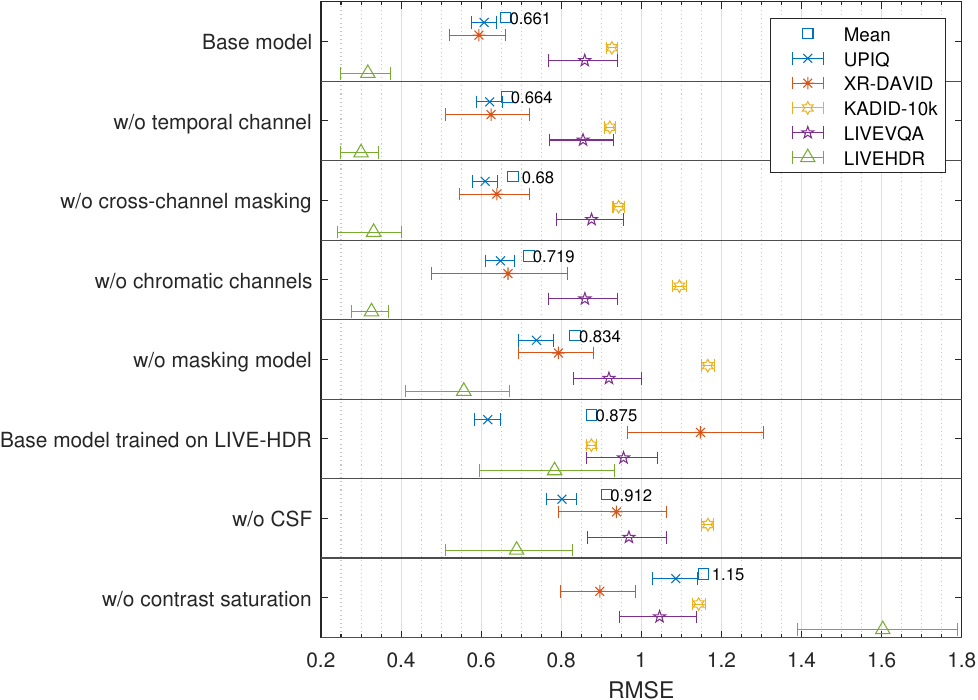}
    \caption{Ablation studies on \ourmethod{}. The top row shows the base model. Refer to the text for details.}
    \label{fig:ablations}
\end{figure}

Second, we tested the importance of each component of our metric, namely:
\begin{itemize}
    \item w/o temporal channel --- ignored the contribution of the achromatic transient channel; 
    \item w/o cross-channel masking --- masking was allowed within each channel, but not across the channels (matrix $k_{i,c}$ from \eqref{mask-min} was diagonal).
    \item w/o chromatic channels --- ignored the contribution of the two chromatic channels; 
    \item w/o masking model --- disabled the masking model but used the contrast encoding from \eqref{csf-normalization} and contrast saturation from \eqref{mask-clamped};
    \item w/o CSF --- the contrast sensitivity function did not vary with luminance or spatial frequency, but varied across the channels;    
    \item w/o contrast saturation --- disabled the contrast saturation formula from \eqref{mask-clamped}.
\end{itemize}
The results of these ablations, shown in \figref{ablations}, demonstrate that the contrast saturation formula is critical for the mutual masking model used in our metric. The metric also performs poorly if the contrast encoding is not modulated by the CSF (see ``w/o CSF'' in \figref{ablations}), or lacks the masking model. Although color is often regarded as a less critical aspect of quality assessment (especially in the context of video coding), here we show that it is important for the datasets that contain color distortions, such as \ourdataset{}, UPIQ, or KADID-10k. The cross-channel masking is an important but subtle effect, which results in gains mostly for the \ourdataset{} dataset. While the influence of the temporal channel may seem small in the results (see ``w/o temporal channel'' in \figref{ablations}), this is due to the gain only being observable for video datasets with temporal distortions, such as \ourdataset{}.

Finally, we tested the importance of the \ourdataset{} dataset when training \ourmethod{}. We retrained \ourmethod{}, but used UPIQ and LIVE-HDR as training sets (instead of UPIQ and \ourdataset{}). The quality scores of LIVE-HDR were scaled in JOD units, as explained in the \href{https://www.cl.cam.ac.uk/research/rainbow/projects/colorvideovdp/}{supplementary}. The results of this training, shown as ``Base model trained on LIVE-HDR'' in \figref{ablations}, demonstrate that the \ourdataset{} dataset is essential for training color video metrics. Video compression datasets, such as LIVEHDR, lack the variety of both temporal and color distortions, which is a necessary component for the calibration of video metrics. The correlation coefficients for all ablations and detailed reports can be found in the \href{https://www.cl.cam.ac.uk/research/rainbow/projects/colorvideovdp/}{supplementary HTML report}.

\subsection{Synthetic tests}
\label{sec:synthetic-tests}

\begin{figure}
    \centering
    \includegraphics[width=0.32\columnwidth]
    {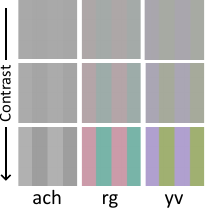}
    \includegraphics[width=0.32\columnwidth]{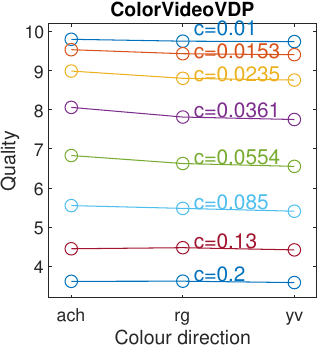}
    \includegraphics[width=0.32\columnwidth]{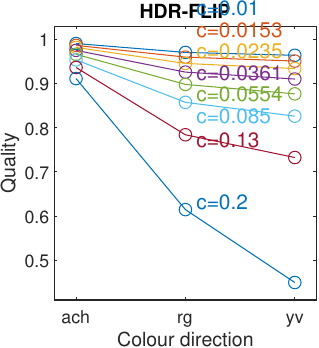}
    \caption{Testing of supra-threshold contrast matching across color directions. The lines in the plots represent metric responses for the contrast modulated along one of the cardinal directions of the DKL color space (examples shown on the left). The magnitude of the contrast was selected to match the data of \citeN{Switkes1999}. If a metric correctly predicts the magnitude of chromatic and achromatic contrast, the plots should form horizontal lines. The values for FLIP have been shown as 1-(FLIP mean difference) to make them comparable with other metrics. Note that the perceived magnitudes depend on the spatial frequency and, therefore, the viewing distance.}
    \label{fig:supra-cont-matching}
\end{figure}

Validation datasets and ablations may not capture the edge cases for which a metric may perform differently than expected. To examine these, we created 14 sets of synthetic test and reference pairs, containing contrast, masking, flicker patterns, and typical distortions. One such example, shown in \figref{supra-cont-matching}, demonstrates the metric's ability to estimate the magnitude of supra-threshold contrast correctly. The lines shown in the plots connect contrast across three color directions that match in perceived magnitude, according to the data of \citeN{Switkes1999}. The figure shows that while \ourmethod{} can correctly predict suprathreshold contrast, HDR-FLIP \cite{Andersson2021a} overpredicts the contrast in the red-green and yellow-violet directions. The extensive report for all 14 sets and multiple metrics can be found in the \href{https://www.cl.cam.ac.uk/research/rainbow/projects/colorvideovdp/}{supplementary HTML report}.

\section{Applications}
\label{sec:applications}

As a general-purpose image and video difference metric, our metric can be used for a range of standard applications, such as optimization of video streaming. In this section, we present three proof-of-concept use cases, which go beyond such standard applications of video metrics. 

\begin{figure*}[t]
    \centering
    \includegraphics[width=\textwidth]{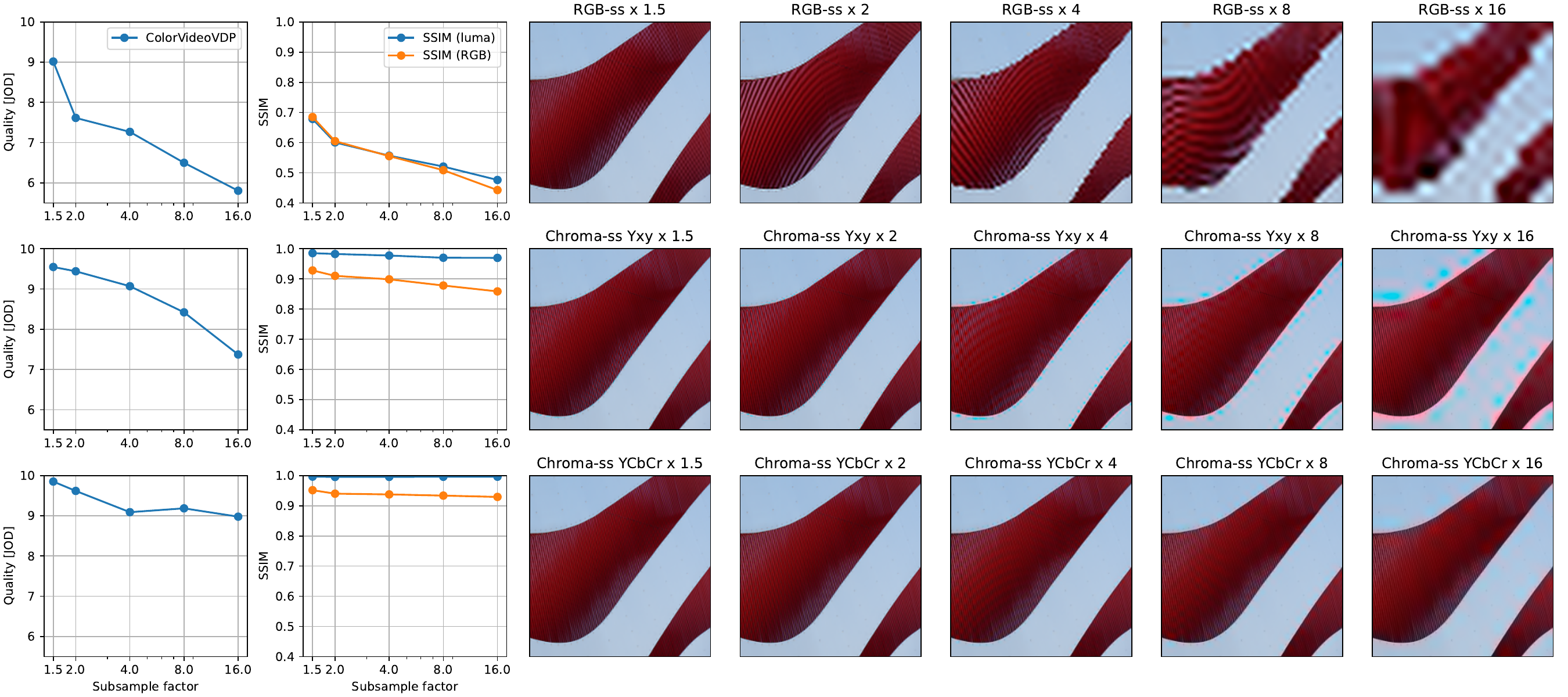}\\
    \caption{The comparison of quality loss due to subsampling of RGB channels (top row), chroma channels of Yxy colorspace (middle row), and chroma channels of YCbCr color space (bottom row). The plots show the predictions of \ourmethod{} (1st column) and SSIM (2nd column), the latter computed either on luma or RGB channels. SSIM fails to predict substantial quality loss when the chroma channels are severely subsampled. 
    }
    \label{fig:chroma_ss}
\end{figure*}

\subsection{Chroma subsampling}
\label{sec:applications_chroma_ss}
Chroma subsampling is a popular compression technique in which chroma channels are encoded with a lower resolution than luma. This works well in practice due to the lower sensitivity of our chromatic vision to high frequencies. The visibility of chroma subsampling artifacts cannot be easily predicted with pixel-wise color difference formulas, such as CIEDE2000, as they are unaware of image structure. Spatial metrics, such as SSIM or even FSIMc, do not operate on color spaces that could properly isolate chromatic and achromatic mechanisms. This is shown in an example in \figref{chroma_ss}, in which SSIM fails to predict a substantial loss of quality at high chromatic subsampling rates, even if the metric is computed on the RGB channels (the original SSIM operates only on luma). The SSIM (RGB) predictions for ${\times}16$ subsampling of $xy$ channels (last column, second row) indicates only a moderate loss of quality (0.87), much lower than even ${\times}1.5$ RGB subsampling (third column, top row, 0.68), which poorly correlates with the perceived level of distortions. \ourmethod{} provides easy-to-interpret predictions, which correspond well with the perceived image quality. It shows, for example, the strength of the YCbCr space, which can well balance subsampling distortions between the achromatic and chromatic channels (the chromatic plane of YCbCr is not isoluminant).


\subsection{Optimization}

\ourmethod{} is fully differentiable, and it can be potentially used as an image loss function when training deep neural networks. However, a good quality metric may not serve as a good loss function \cite{Mustafa_2022}, in particular when the features used by a metric are not surjective \cite{Ding2021}. Moreover, 
complex functions can make the landscape of the loss highly non-convex and impede convergence. For that reason, a popular "perceptual loss" \cite{Johnson_2016} must be combined with L1 or L2 to ensure convergence.

We found that when \ourmethod{} is used as an image loss function (without L1 or L2 term) in an image reconstruction task (see \cite[Sec.~3]{Ding2021}), it converges well when the reconstructed image is close to the reference. However, the convergence is worse when the test image is affected by random noise or is far from the reference. This problem is alleviated when \ourmethod{} is used when optimizing low-dimensional problems, such as the selection of encoding parameters in video streaming. In the next section, we show an example of \ourmethod{} used as a loss function in a high-dimensional problem.

\subsection{Adaptive chroma subsampling}

\begin{figure}
    \centering
    \includegraphics{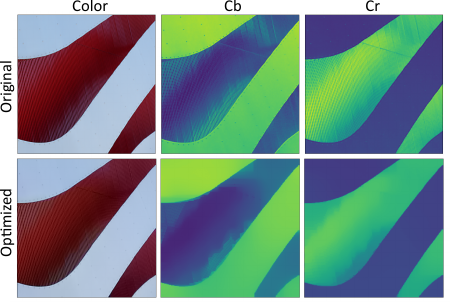}
    \caption{Adaptive chroma subsampling removes high frequencies from chromatic Cb and Cr planes (2nd and 3rd column) of the YCbCr color space without affecting the appearance of the color image (1st column). Such adaptive subsampling can effectively improve image compression without the loss of chromatic details.}
    \label{fig:adaptive-chroma-ss}
\end{figure}

Traditional chroma subsampling, as demonstrated in \secref{applications_chroma_ss}, globally reduces the resolution (sampling rate) of the chromatic planes regardless of image content. However, the visibility of chromatic distortion will strongly depend on image content, and therefore, we argue that chroma subsampling should be adaptive and vary across an image. Here, we show that we can use \ourmethod{} as a differential loss to adaptively remove high frequencies from the image chroma channels while accounting for all factors that contribute to the visibility of the introduced changes. To do this, we optimize for the color image $I$:
\begin{equation}
\argmin_{I}||I-I_\ind{org}||_\ind{cvvdp} + \lambda \left|| \nabla I_\ind{Cb} \right||_1 + \lambda|| \nabla I_\ind{Cr} ||_1
\end{equation}
where $I_\ind{org}$ is the original image, $\nabla I_\ind{Cb}$ and $\nabla I_\ind{Cr}$ are the gradients of the Cb and Cr planes of the YCbCr color space and $\lambda$ is a regularization constant. $||\cdot||_\ind{cvvdp}$ denotes \ourmethod{} used as a differentiable loss and $||\cdot||_1$ is an L1 norm. \figref{adaptive-chroma-ss} shows an example of such adaptive chroma subsampling. In this particular example, we could reduce the size of a PNG file by 10\% with no impact on image quality.

\begin{figure}[t]
    \centering
    \includegraphics[width=\linewidth]{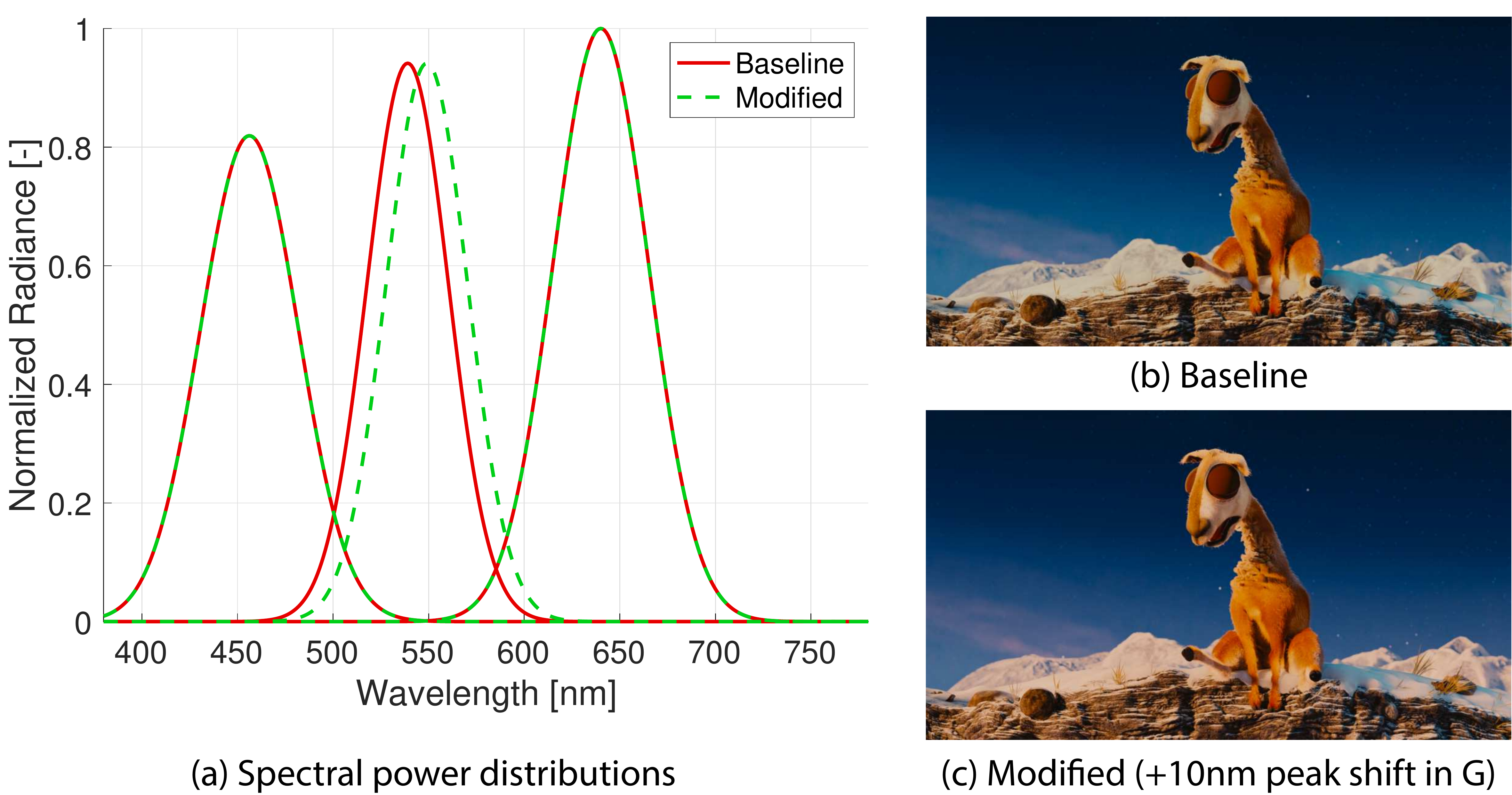}
    \caption{(a) Spectral power distributions for baseline (red solid line) and modified curves (green dashed line), (b) Proof image shown on display with baseline primaries, (c) Proof image shown on display with modified primaries. © caminandes.com}
    \label{fig:SPD_baseline_vs_mod}
\end{figure}

\subsection{Setting display color tolerance specifications}
\label{sec:applications_tolerance_specs}
In a display manufacturing setting, it is common for display primaries' spectra to differ from the desired target values between individual units and suppliers. Manufacturers typically set specifications for how much each primary can deviate from the ideal target. Traditionally, it has been characterized in wavelength shifts, changes in chromaticity and luminance, or color difference metrics such as $\Delta\text{E}^*_{ab}$ and CIEDE2000. \ourmethod{} can be used to set these specifications in an interpretable manner (in terms of JODs) with respect to sample content, as our metric is aware of image structure. 

As an example use case, assume a VR display is being characterized in a factory calibration setting. We simulate the reference by generating a synthetic set of R, G, and B primaries producing a P3 gamut using Gaussian curves, as depicted by the red solid line in \figref{SPD_baseline_vs_mod}(a). Next, we simulate possible spectral primary deviations by varying the spectral peak and full width at half maximum (FWHM) for each primary to examine tolerances (green dashed line in \figref{SPD_baseline_vs_mod}(a)). A test image rendered with both baseline and distorted primaries is shown in \figref{SPD_baseline_vs_mod} (b) and (c), respectively. Finally, JODs scores were calculated between images with baseline and modified primaries for each spectral peak shift and FWHM change for each of four test images (taken from \emph{Caminandes}, \emph{Icons}, \emph{Panel}, and \emph{VR}, see the \href{https://www.cl.cam.ac.uk/research/rainbow/projects/colorvideovdp/}{supplementary document}).



\begin{figure}[t]
    \centering
    \includegraphics[width=\linewidth]{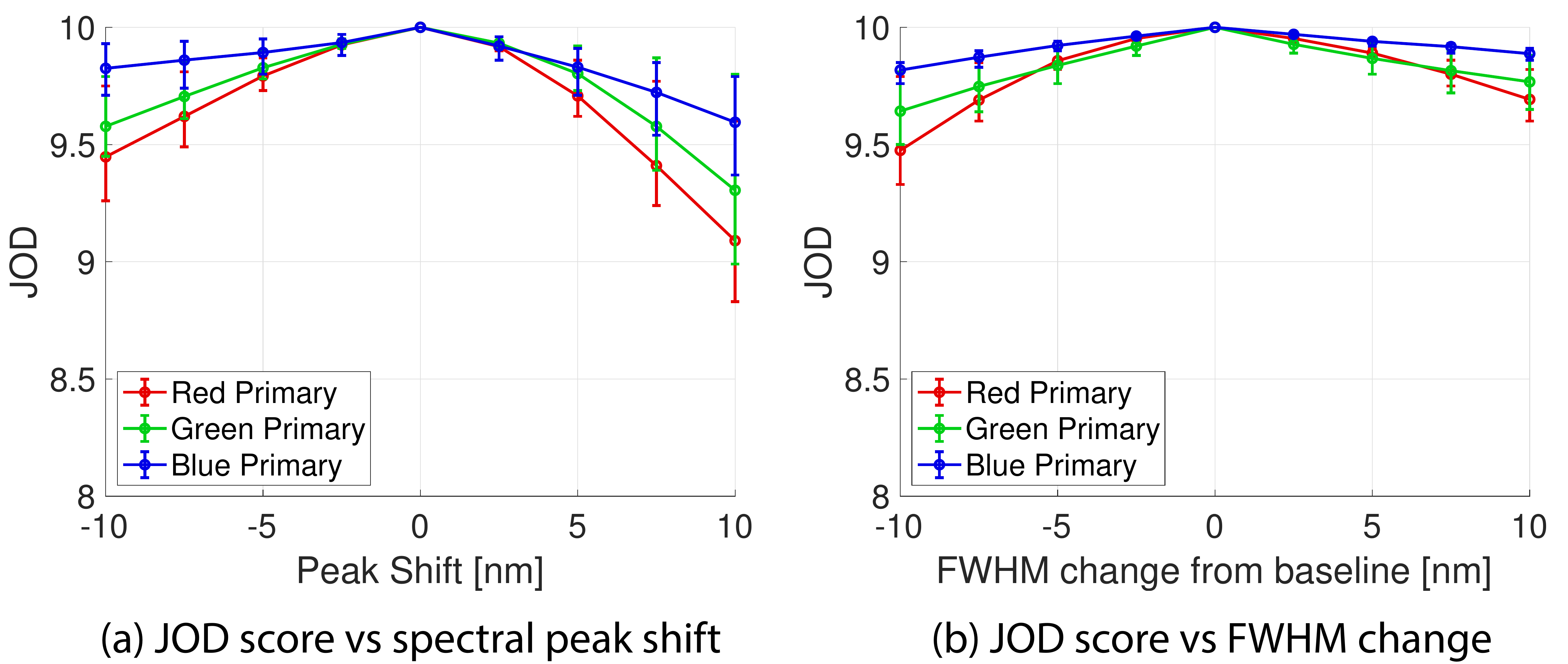}
    \caption{(a) JOD scores vs spectral peak shift for R, G, B primaries. (b) JOD scores vs. FWHM change from baseline for R, G, B primaries. Error bars show maximum and minimum JOD scores among 4 scenes for each case.}
    \label{fig:JOD_vs_PkShift_FWHM}
\end{figure}

The results are shown in \figref{JOD_vs_PkShift_FWHM}. Note that JOD scores change nontrivially depending on the direction of the peak shift, FWHM magnitude, and type of primary (R, G, or B). In our imaginary example, factory tolerance specifications could then be derived by setting an acceptable limit for $\Delta$JOD deviation from the baseline via a psychophysical or "golden eye" study. Further investigation can be conducted by looking into individual images for worst performers, and future tolerances can be easily tightened by reducing the $\Delta$JOD specification.



\subsection{Observer metamerism and variability}
\label{sec:applications_metamerism}
Observer metamerism has been an issue for wide-color-gamut displays. As wide color gamut displays typically have spectrally narrower primaries, it is increasingly likely that, when calibrated for a standard observer, they will appear color-inaccurate for individuals who deviate from this profile \cite{bodner201878,hung201961}. The severity of this observer metamerism depends on both the content being shown and the display characteristics, and can be evaluated by using \ourmethod{}. 

\begin{figure}
    \includegraphics[width=\linewidth]{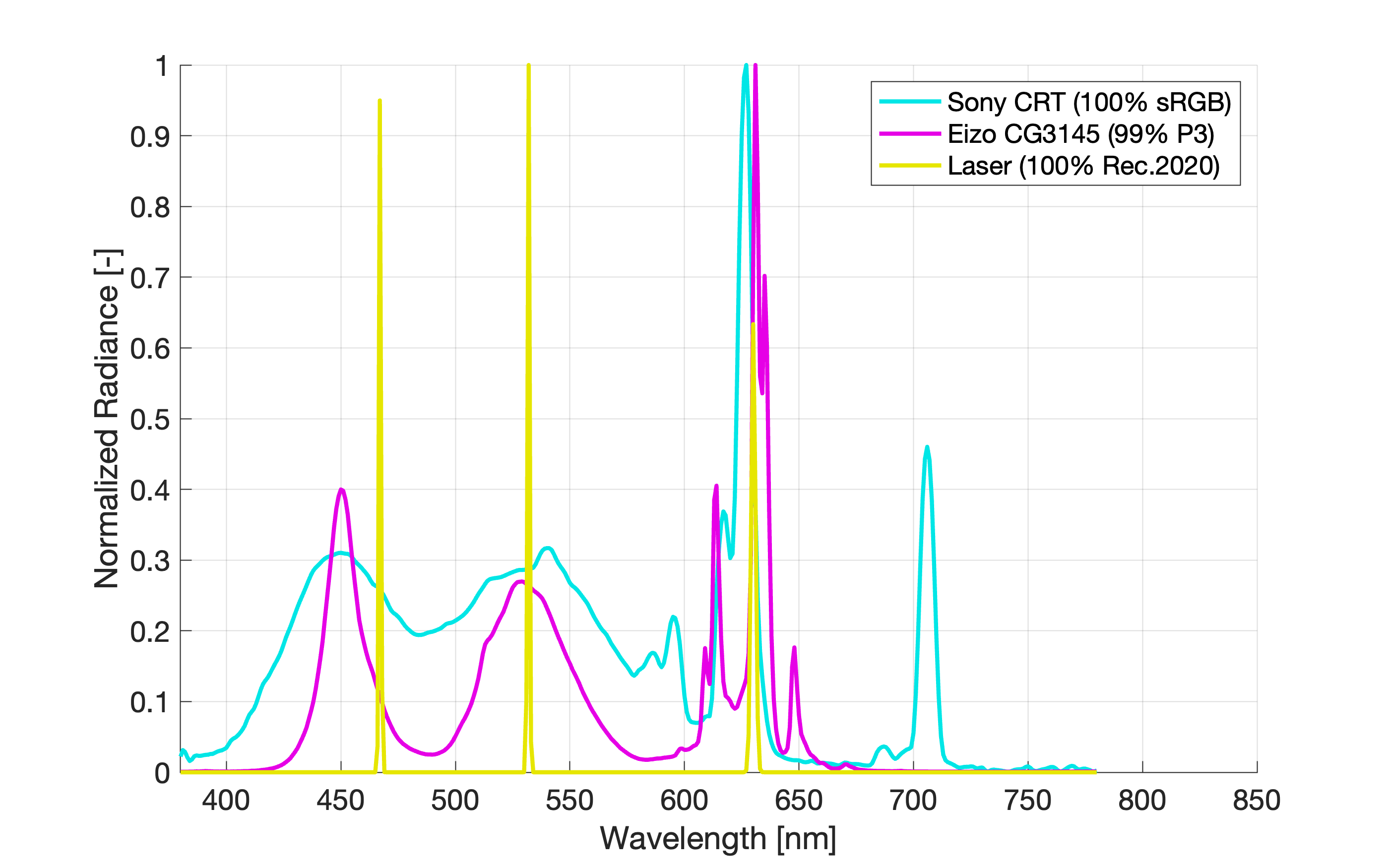}
    \\
    \includegraphics[width=\linewidth]{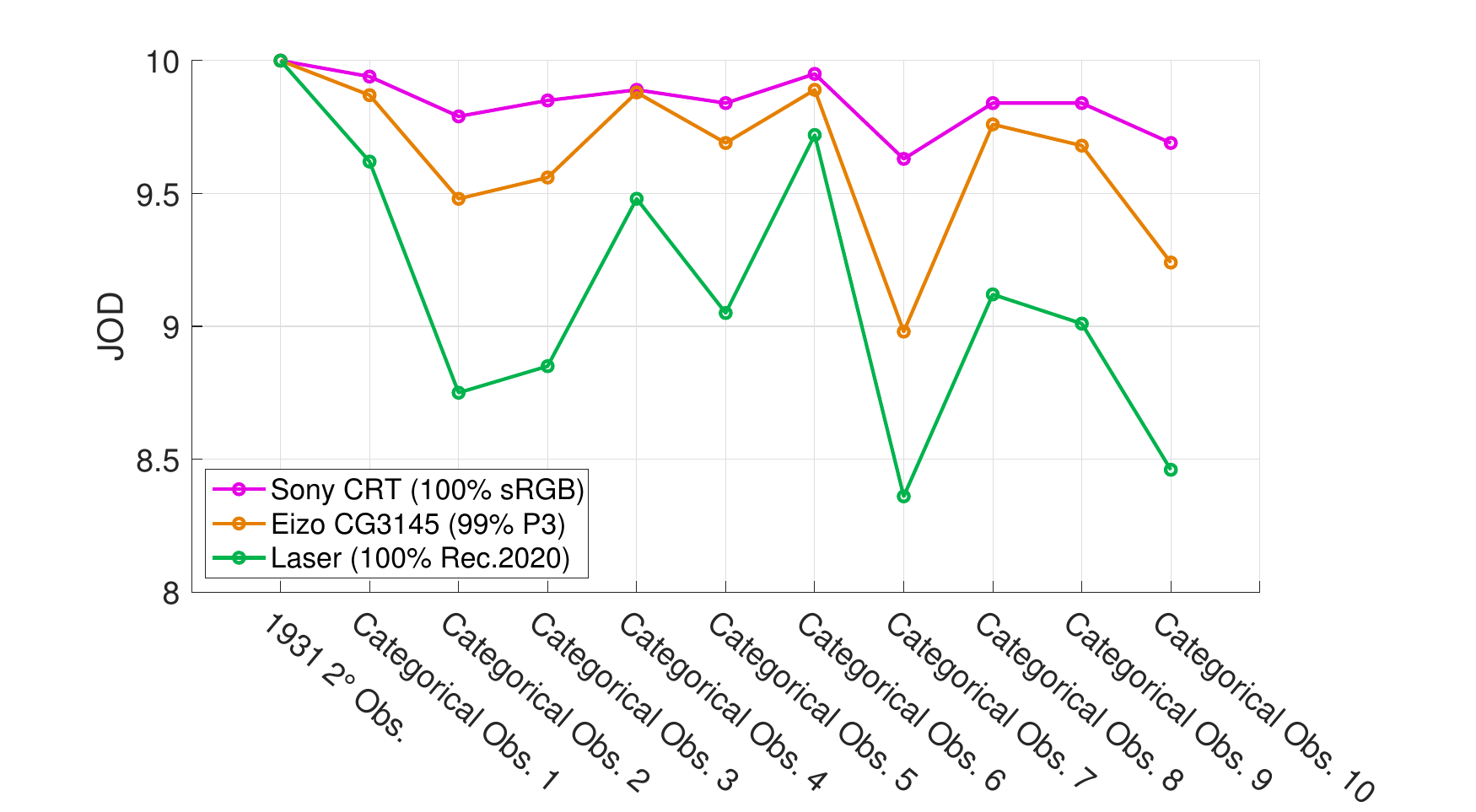}
    \vspace{-4mm}
    \caption{(Top) Spectral power distributions for three types of displays used to simulate observer metamerism (see \secref{applications_metamerism}). (Bottom) JOD scores for all observers relative to 1931 2$^{\circ}$ standard observer for the `Wiki' scene, for three display types.}
    \label{fig:applications_metamerism}
    \vspace{-4mm}
\end{figure}
\begin{figure*}[p]
    \centering
    \includegraphics[width=\textwidth]{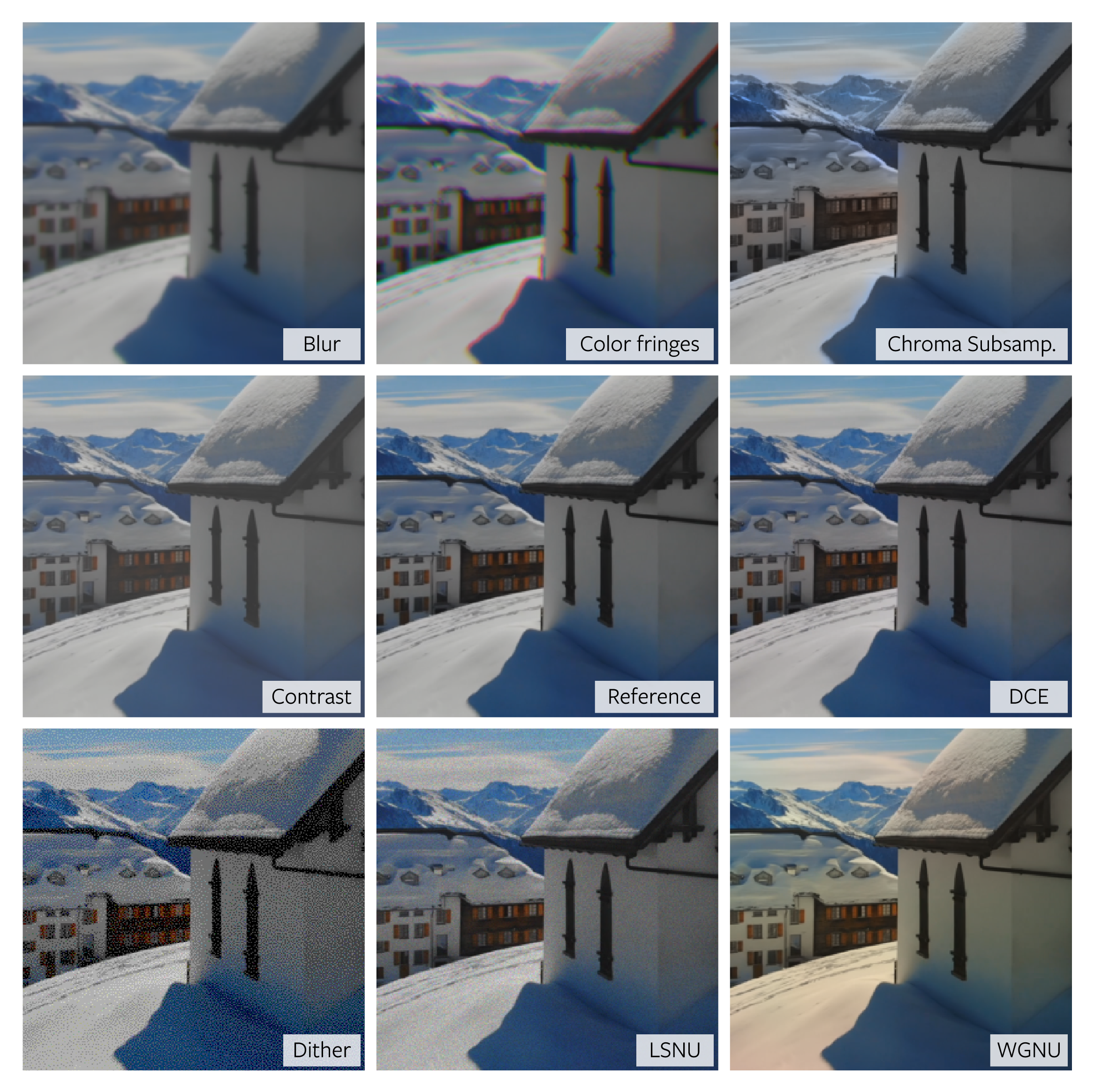}
    \caption{All 8 distortions as used in the \ourdataset{} study at level 3 (most distorted). The undistorted reference is shown in the center for comparison. An inset of the \emph{Snow} scene is shown to enhance artifact visibility. Note that DCE (Dynamic Correction Error) does not exhibit much visible distortion as the artifact is mostly present in the temporal domain. ``\href{https://www.pexels.com/video/a-drone-shot-of-a-church-in-belalp-3818213/}{A Drone Shot of a Church in Belalp}'' by SwissHumanity Stories under a Creative Commons license.}
    \label{fig:allartifacts}
\end{figure*}

To demonstrate this, we calculated JOD scores for the `Wiki' scene, simulating three different display primary spectral profiles and 11 different observer functions. The evaluated displays were Sony BVM32 CRT (100\% of sRGB), Eizo ColorEdge 3145 (99\% P3), and laser primaries (100\% ITU-R BT.2020 color gamuts). Laser spectra were synthetically generated. Spectral power distributions for all primaries are shown in \figref{applications_metamerism} (top). To simulate individual differences in color vision, we used Asano and Fairchild's 10 categorical observers \shortcite{asano2020categorical}, as well as the 1931 2$^{\circ}$ standard observer used for reference. These 10-observer functions were derived as representative means for a color-normal population based on an individual colorimetric observer model \cite{asano2016individual}. The results are shown in \figref{applications_metamerism} (bottom). As expected, the narrower spectra incur a larger metameric error for non-standard observers, which is reflected in the higher $\Delta$JOD values. \ourmethod{} can be used to estimate the risk of metameric error for populations on content, and to evaluate or optimize solutions for this problem. 


\section{Conclusions}
In this work we introduce \ourmethod{}, a general-purpose image and video metric that models several challenging aspects of vision simultaneously. Notably, our algorithm is calibrated in psychophysical JOD units, models color, high-dynamic-range, and spatiotemporal aspects of vision. Our metric is explainable, and each component used to build our pipeline is based on proven psychophysical models of the human visual system.

An important aspect of our metric is its extensive calibration. Our metric is simultaneously calibrated on 3 large datasets containing a variety of distortion types, which included our novel \ourdataset{} video quality dataset, containing a range of display hardware-oriented artifacts critically important to the development of future displays. We demonstrated that our metric generalizes well to unseen datasets. 

Finally, \ourmethod{} is efficiently implemented to run on a GPU. This aspect is increasingly important in the modern display landscape, as resolution, frame rate, field of view, and bit-depth continue to increase, adding complexity to analyzed content. It is also fully differentiable, making it possible to use it as a loss function in optimization problems.


\begin{acks}
We thank Dounia Hammou for her insightful comments. We thank Cameron Wood, Eloise Moore, Helen Ayele, Sydnie Gregory and Ka Yan Wat for help running the user study. Thanks go to Curtis Torrey and Xin Li for assistance with experimental hardware, and to Jacqueline Beyrouty for help with logistics.
\end{acks}

\bibliographystyle{ACM-Reference-Format}
\bibliography{main}


\begin{thebibliography}{70}


\ifx \showCODEN    \undefined \def \showCODEN     #1{\unskip}     \fi
\ifx \showDOI      \undefined \def \showDOI       #1{#1}\fi
\ifx \showISBNx    \undefined \def \showISBNx     #1{\unskip}     \fi
\ifx \showISBNxiii \undefined \def \showISBNxiii  #1{\unskip}     \fi
\ifx \showISSN     \undefined \def \showISSN      #1{\unskip}     \fi
\ifx \showLCCN     \undefined \def \showLCCN      #1{\unskip}     \fi
\ifx \shownote     \undefined \def \shownote      #1{#1}          \fi
\ifx \showarticletitle \undefined \def \showarticletitle #1{#1}   \fi
\ifx \showURL      \undefined \def \showURL       {\relax}        \fi
\providecommand\bibfield[2]{#2}
\providecommand\bibinfo[2]{#2}
\providecommand\natexlab[1]{#1}
\providecommand\showeprint[2][]{arXiv:#2}

\bibitem[Alam et~al\mbox{.}(2014)]%
        {Alam_Vilankar_Field_Chandler_2014}
\bibfield{author}{\bibinfo{person}{M.~M. Alam}, \bibinfo{person}{K.~P. Vilankar}, \bibinfo{person}{David~J Field}, {and} \bibinfo{person}{Damon~M Chandler}.} \bibinfo{year}{2014}\natexlab{}.
\newblock \showarticletitle{Local masking in natural images: A database and analysis}.
\newblock \bibinfo{journal}{\emph{Journal of Vision}} \bibinfo{volume}{14}, \bibinfo{number}{8} (\bibinfo{date}{July} \bibinfo{year}{2014}), \bibinfo{pages}{22–22}.
\newblock
\showISSN{1534-7362}
\urldef\tempurl%
\url{https://doi.org/10.1167/14.8.22}
\showDOI{\tempurl}
\newblock
\shownote{Citation Key: Alam2014}.


\bibitem[Anderson and Burr(1985)]%
        {Anderson1985}
\bibfield{author}{\bibinfo{person}{Stephen~J. Anderson} {and} \bibinfo{person}{David~C. Burr}.} \bibinfo{year}{1985}\natexlab{}.
\newblock \showarticletitle{{Spatial and temporal selectivity of the human motion detection system}}.
\newblock \bibinfo{journal}{\emph{Vision Research}} \bibinfo{volume}{25}, \bibinfo{number}{8} (\bibinfo{date}{jan} \bibinfo{year}{1985}), \bibinfo{pages}{1147--1154}.
\newblock
\showISSN{00426989}
\urldef\tempurl%
\url{https://doi.org/10.1016/0042-6989(85)90104-X}
\showDOI{\tempurl}


\bibitem[Andersson et~al\mbox{.}(2020)]%
        {Andersson2020}
\bibfield{author}{\bibinfo{person}{Pontus Andersson}, \bibinfo{person}{Jim Nilsson}, \bibinfo{person}{Tomas Akenine-M{\"{o}}ller}, \bibinfo{person}{Magnus Oskarsson}, \bibinfo{person}{Kalle {\AA}str{\"{o}}m}, {and} \bibinfo{person}{Mark~D. Fairchild}.} \bibinfo{year}{2020}\natexlab{}.
\newblock \showarticletitle{{FLIP: A Difference Evaluator for Alternating Images}}.
\newblock \bibinfo{journal}{\emph{Proc. of the ACM on Computer Graphics and Interactive Techniques}} \bibinfo{volume}{3}, \bibinfo{number}{2} (\bibinfo{date}{aug} \bibinfo{year}{2020}), \bibinfo{pages}{1--23}.
\newblock
\showISSN{2577-6193}
\urldef\tempurl%
\url{https://doi.org/10.1145/3406183}
\showDOI{\tempurl}


\bibitem[Andersson et~al\mbox{.}(2021)]%
        {Andersson2021a}
\bibfield{author}{\bibinfo{person}{Pontus Andersson}, \bibinfo{person}{Jim Nilsson}, \bibinfo{person}{Peter Shirley}, {and} \bibinfo{person}{Tomas Akenine{-}M{\"{o}}ller}.} \bibinfo{year}{2021}\natexlab{}.
\newblock \showarticletitle{{Visualizing Errors in Rendered High Dynamic Range Images}}. In \bibinfo{booktitle}{\emph{Eurographics Short Papers}}.
\newblock
\urldef\tempurl%
\url{https://doi.org/10.2312/egs.20211015}
\showDOI{\tempurl}


\bibitem[Asano and Fairchild(2020)]%
        {asano2020categorical}
\bibfield{author}{\bibinfo{person}{Yuta Asano} {and} \bibinfo{person}{Mark~D Fairchild}.} \bibinfo{year}{2020}\natexlab{}.
\newblock \showarticletitle{Categorical observers for metamerism}.
\newblock \bibinfo{journal}{\emph{Color Research \& Application}} \bibinfo{volume}{45}, \bibinfo{number}{4} (\bibinfo{year}{2020}), \bibinfo{pages}{576--585}.
\newblock


\bibitem[Asano et~al\mbox{.}(2016)]%
        {asano2016individual}
\bibfield{author}{\bibinfo{person}{Yuta Asano}, \bibinfo{person}{Mark~D Fairchild}, {and} \bibinfo{person}{Laurent Blond{\'e}}.} \bibinfo{year}{2016}\natexlab{}.
\newblock \showarticletitle{Individual colorimetric observer model}.
\newblock \bibinfo{journal}{\emph{PloS one}} \bibinfo{volume}{11}, \bibinfo{number}{2} (\bibinfo{year}{2016}), \bibinfo{pages}{e0145671}.
\newblock


\bibitem[Ashraf et~al\mbox{.}(2024)]%
        {Ashraf2024}
\bibfield{author}{\bibinfo{person}{Maliha Ashraf}, \bibinfo{person}{Rafał~K. Mantiuk}, \bibinfo{person}{Alexandre Chapiro}, {and} \bibinfo{person}{Sophie Wuerger}.} \bibinfo{year}{2024}\natexlab{}.
\newblock \showarticletitle{castleCSF — A Contrast Sensitivity Function of Color, Area, Spatio-Temporal frequency, Luminance and Eccentricity}.
\newblock \bibinfo{journal}{\emph{Journal of Vision}}  \bibinfo{volume}{24} (\bibinfo{year}{2024}), \bibinfo{pages}{5}.
\newblock
\urldef\tempurl%
\url{https://doi.org/10.1167/jov.24.4.5}
\showDOI{\tempurl}


\bibitem[Bodner et~al\mbox{.}(2018)]%
        {bodner201878}
\bibfield{author}{\bibinfo{person}{Ben Bodner}, \bibinfo{person}{Neil Robinson}, \bibinfo{person}{Robin Atkins}, {and} \bibinfo{person}{Scott Daly}.} \bibinfo{year}{2018}\natexlab{}.
\newblock \showarticletitle{78-1: Correcting Metameric Failure of Wide Color Gamut Displays}. In \bibinfo{booktitle}{\emph{SID Symposium Digest of Technical Papers}}, Vol.~\bibinfo{volume}{49}. Wiley Online Library, \bibinfo{pages}{1040--1043}.
\newblock


\bibitem[Burt and Adelson(1983)]%
        {Burt1983}
\bibfield{author}{\bibinfo{person}{P. Burt} {and} \bibinfo{person}{E. Adelson}.} \bibinfo{year}{1983}\natexlab{}.
\newblock \showarticletitle{{The Laplacian Pyramid as a Compact Image Code}}.
\newblock \bibinfo{journal}{\emph{IEEE Transactions on Communications}} \bibinfo{volume}{31}, \bibinfo{number}{4} (\bibinfo{date}{apr} \bibinfo{year}{1983}), \bibinfo{pages}{532--540}.
\newblock
\showISSN{0096-2244}
\urldef\tempurl%
\url{https://doi.org/10.1109/TCOM.1983.1095851}
\showDOI{\tempurl}


\bibitem[Cass et~al\mbox{.}(2009)]%
        {Cass2009}
\bibfield{author}{\bibinfo{person}{John Cass}, \bibinfo{person}{C.~W.~G. Clifford}, \bibinfo{person}{David Alais}, {and} \bibinfo{person}{Branka Spehar}.} \bibinfo{year}{2009}\natexlab{}.
\newblock \showarticletitle{{Temporal structure of chromatic channels revealed through masking}}.
\newblock \bibinfo{journal}{\emph{Journal of Vision}} \bibinfo{volume}{9}, \bibinfo{number}{5} (\bibinfo{date}{may} \bibinfo{year}{2009}), \bibinfo{pages}{17--17}.
\newblock
\showISSN{1534-7362}
\urldef\tempurl%
\url{https://doi.org/10.1167/9.5.17}
\showDOI{\tempurl}


\bibitem[Chen et~al\mbox{.}(2016)]%
        {chen2016checkpointing}
\bibfield{author}{\bibinfo{person}{Tianqi Chen}, \bibinfo{person}{Bing Xu}, \bibinfo{person}{Chiyuan Zhang}, {and} \bibinfo{person}{Carlos Guestrin}.} \bibinfo{year}{2016}\natexlab{}.
\newblock \showarticletitle{Training deep nets with sublinear memory cost}.
\newblock \bibinfo{journal}{\emph{arXiv preprint arXiv:1604.06174}} (\bibinfo{year}{2016}).
\newblock


\bibitem[Cheon et~al\mbox{.}(2021)]%
        {Cheon_Yoon_Kang_Lee_2021}
\bibfield{author}{\bibinfo{person}{Manri Cheon}, \bibinfo{person}{Sung-Jun Yoon}, \bibinfo{person}{Byungyeon Kang}, {and} \bibinfo{person}{Junwoo Lee}.} \bibinfo{year}{2021}\natexlab{}.
\newblock \showarticletitle{Perceptual Image Quality Assessment with Transformers}. In \bibinfo{booktitle}{\emph{2021 IEEE/CVF Conference on Computer Vision and Pattern Recognition Workshops (CVPRW)}}. \bibinfo{publisher}{IEEE}, \bibinfo{address}{Nashville, TN, USA}, \bibinfo{pages}{433–442}.
\newblock
\showISBNx{978-1-66544-899-4}
\urldef\tempurl%
\url{https://doi.org/10.1109/CVPRW53098.2021.00054}
\showDOI{\tempurl}


\bibitem[Choudhury et~al\mbox{.}(2021)]%
        {Choudhury2021}
\bibfield{author}{\bibinfo{person}{Anustup Choudhury}, \bibinfo{person}{Robert Wanat}, \bibinfo{person}{Jaclyn Pytlarz}, {and} \bibinfo{person}{Scott Daly}.} \bibinfo{year}{2021}\natexlab{}.
\newblock \showarticletitle{{Image quality evaluation for high dynamic range and wide color gamut applications using visual spatial processing of color differences}}.
\newblock \bibinfo{journal}{\emph{Color Research \& Application}} \bibinfo{volume}{46}, \bibinfo{number}{1} (\bibinfo{date}{feb} \bibinfo{year}{2021}), \bibinfo{pages}{46--64}.
\newblock
\showISSN{0361-2317}
\urldef\tempurl%
\url{https://doi.org/10.1002/col.22588}
\showDOI{\tempurl}


\bibitem[CIE(1993)]%
        {CIE_1993}
\bibfield{author}{\bibinfo{person}{CIE}.} \bibinfo{year}{1993}\natexlab{}.
\newblock \bibinfo{booktitle}{\emph{Parametric effects in colour-difference evaluation}}.
\newblock \bibinfo{type}{{T}echnical {R}eport}. \bibinfo{institution}{CIE 101-1993}.
\newblock


\bibitem[CIE(2018)]%
        {CIE15_2018}
\bibfield{author}{\bibinfo{person}{CIE}.} \bibinfo{year}{2018}\natexlab{}.
\newblock \showarticletitle{CIE 015: 2018 Colorimetry}.
\newblock  (\bibinfo{year}{2018}).
\newblock


\bibitem[Daly(1993)]%
        {Daly1993}
\bibfield{author}{\bibinfo{person}{S.J. Daly}.} \bibinfo{year}{1993}\natexlab{}.
\newblock \showarticletitle{{Visible differences predictor: an algorithm for the assessment of image fidelity}}.
\newblock In \bibinfo{booktitle}{\emph{Digital Images and Human Vision}}, \bibfield{editor}{\bibinfo{person}{Andrew~B. Watson}} (Ed.). Vol.~\bibinfo{volume}{1666}. \bibinfo{publisher}{MIT Press}, \bibinfo{pages}{179--206}.
\newblock
\showISBNx{978-0262231718}
\showISSN{0277786X}
\urldef\tempurl%
\url{https://doi.org/10.1117/12.135952}
\showDOI{\tempurl}


\bibitem[{De Valois} et~al\mbox{.}(1982)]%
        {de1982spatial}
\bibfield{author}{\bibinfo{person}{R.L. {De Valois}}, \bibinfo{person}{D.G. Albrecht}, {and} \bibinfo{person}{L.G. Thorell}.} \bibinfo{year}{1982}\natexlab{}.
\newblock \showarticletitle{{Spatial frequency selectivity of cells in macaque visual cortex}}.
\newblock \bibinfo{journal}{\emph{Vision Research}} \bibinfo{volume}{22}, \bibinfo{number}{5} (\bibinfo{year}{1982}), \bibinfo{pages}{545--559}.
\newblock
\showISSN{0042-6989}
\urldef\tempurl%
\url{https://doi.org/10.1016/0042-6989(82)90113-4}
\showDOI{\tempurl}


\bibitem[Denes et~al\mbox{.}(2020)]%
        {Denes2020}
\bibfield{author}{\bibinfo{person}{Gyorgy Denes}, \bibinfo{person}{Akshay Jindal}, \bibinfo{person}{Aliaksei Mikhailiuk}, {and} \bibinfo{person}{Rafa{\l}~K. Mantiuk}.} \bibinfo{year}{2020}\natexlab{}.
\newblock \showarticletitle{{A perceptual model of motion quality for rendering with adaptive refresh-rate and resolution}}.
\newblock \bibinfo{journal}{\emph{ACM Transactions on Graphics}} \bibinfo{volume}{39}, \bibinfo{number}{4} (\bibinfo{date}{jul} \bibinfo{year}{2020}), \bibinfo{pages}{133}.
\newblock
\showISSN{0730-0301}
\urldef\tempurl%
\url{https://doi.org/10.1145/3386569.3392411}
\showDOI{\tempurl}


\bibitem[Derrington et~al\mbox{.}(1984)]%
        {Derrington1984}
\bibfield{author}{\bibinfo{person}{A~M Derrington}, \bibinfo{person}{J Krauskopf}, {and} \bibinfo{person}{P Lennie}.} \bibinfo{year}{1984}\natexlab{}.
\newblock \showarticletitle{{Chromatic mechanisms in lateral geniculate nucleus of macaque.}}
\newblock \bibinfo{journal}{\emph{The Journal of Physiology}} \bibinfo{volume}{357}, \bibinfo{number}{1} (\bibinfo{date}{dec} \bibinfo{year}{1984}), \bibinfo{pages}{241--265}.
\newblock
\showISSN{00223751}
\urldef\tempurl%
\url{https://doi.org/10.1113/jphysiol.1984.sp015499}
\showDOI{\tempurl}


\bibitem[Didyk et~al\mbox{.}(2011)]%
        {Didyk2011}
\bibfield{author}{\bibinfo{person}{Piotr Didyk}, \bibinfo{person}{Tobias Ritschel}, \bibinfo{person}{Elmar Eisemann}, \bibinfo{person}{Karol Myszkowski}, {and} \bibinfo{person}{Hans-peter Seidel}.} \bibinfo{year}{2011}\natexlab{}.
\newblock \showarticletitle{{A perceptual model for disparity}}.
\newblock \bibinfo{journal}{\emph{ACM Transactions on Graphics}} \bibinfo{volume}{30}, \bibinfo{number}{4} (\bibinfo{date}{jul} \bibinfo{year}{2011}), \bibinfo{pages}{1}.
\newblock
\showISSN{07300301}
\urldef\tempurl%
\url{https://doi.org/10.1145/2010324.1964991}
\showDOI{\tempurl}


\bibitem[Ding et~al\mbox{.}(2021)]%
        {Ding2021}
\bibfield{author}{\bibinfo{person}{Keyan Ding}, \bibinfo{person}{Kede Ma}, \bibinfo{person}{Shiqi Wang}, {and} \bibinfo{person}{Eero~P. Simoncelli}.} \bibinfo{year}{2021}\natexlab{}.
\newblock \showarticletitle{{Comparison of Full-Reference Image Quality Models for Optimization of Image Processing Systems}}.
\newblock \bibinfo{journal}{\emph{International Journal of Computer Vision}} \bibinfo{volume}{129}, \bibinfo{number}{4} (\bibinfo{date}{apr} \bibinfo{year}{2021}), \bibinfo{pages}{1258--1281}.
\newblock
\showISSN{0920-5691}
\urldef\tempurl%
\url{https://doi.org/10.1007/s11263-020-01419-7}
\showDOI{\tempurl}
\showeprint[arxiv]{2005.01338}


\bibitem[Foley(1994)]%
        {Foley1994a}
\bibfield{author}{\bibinfo{person}{John~M. Foley}.} \bibinfo{year}{1994}\natexlab{}.
\newblock \showarticletitle{{Human luminance pattern-vision mechanisms: masking experiments require a new model}}.
\newblock \bibinfo{journal}{\emph{Journal of the Optical Society of America A}} \bibinfo{volume}{11}, \bibinfo{number}{6} (\bibinfo{date}{jun} \bibinfo{year}{1994}), \bibinfo{pages}{1710}.
\newblock
\showISSN{1084-7529}
\urldef\tempurl%
\url{https://doi.org/10.1364/JOSAA.11.001710}
\showDOI{\tempurl}


\bibitem[Hardy et~al\mbox{.}(1945)]%
        {hardy1945tests}
\bibfield{author}{\bibinfo{person}{LeGrand~H Hardy}, \bibinfo{person}{Gertrude Rand}, {and} \bibinfo{person}{M~Catherine Rittler}.} \bibinfo{year}{1945}\natexlab{}.
\newblock \showarticletitle{Tests for the detection and analysis of color-blindness. I. The Ishihara test: An evaluation}.
\newblock \bibinfo{journal}{\emph{JOSA}} \bibinfo{volume}{35}, \bibinfo{number}{4} (\bibinfo{year}{1945}), \bibinfo{pages}{268--275}.
\newblock


\bibitem[Hess and Snowden(1992)]%
        {Hess1992}
\bibfield{author}{\bibinfo{person}{R.F. Hess} {and} \bibinfo{person}{R.J. Snowden}.} \bibinfo{year}{1992}\natexlab{}.
\newblock \showarticletitle{{Temporal properties of human visual filters: number, shapes and spatial covariation}}.
\newblock \bibinfo{journal}{\emph{Vision Research}} \bibinfo{volume}{32}, \bibinfo{number}{1} (\bibinfo{date}{jan} \bibinfo{year}{1992}), \bibinfo{pages}{47--59}.
\newblock
\showISSN{00426989}
\urldef\tempurl%
\url{https://doi.org/10.1016/0042-6989(92)90112-V}
\showDOI{\tempurl}


\bibitem[Hung(2019)]%
        {hung201961}
\bibfield{author}{\bibinfo{person}{Po-Chieh Hung}.} \bibinfo{year}{2019}\natexlab{}.
\newblock \showarticletitle{61-3: Invited paper: CIE activities on wide colour gamut and high dynamic range imaging}. In \bibinfo{booktitle}{\emph{SID Symposium Digest of Technical Papers}}, Vol.~\bibinfo{volume}{50}. Wiley Online Library, \bibinfo{pages}{866--869}.
\newblock


\bibitem[{ITU-R BT. 2124}(2019)]%
        {ITU-RBT.2124}
\bibfield{author}{\bibinfo{person}{{ITU-R BT. 2124}}.} \bibinfo{year}{2019}\natexlab{}.
\newblock \bibinfo{booktitle}{\emph{{Objective metric for the assessment of the potential visibility of colour differences in television}}}.
\newblock \bibinfo{type}{{T}echnical {R}eport}.
\newblock


\bibitem[Johnson et~al\mbox{.}(2016)]%
        {Johnson_2016}
\bibfield{author}{\bibinfo{person}{Justin Johnson}, \bibinfo{person}{Alexandre Alahi}, {and} \bibinfo{person}{Li Fei-Fei}.} \bibinfo{year}{2016}\natexlab{}.
\newblock \bibinfo{booktitle}{\emph{Perceptual Losses for Real-Time Style Transfer and Super-Resolution}}. \bibinfo{series}{Lecture Notes in Computer Science}, Vol.~\bibinfo{volume}{9906}.
\newblock \bibinfo{publisher}{Springer International Publishing}, \bibinfo{address}{Cham}, \bibinfo{pages}{694–711}.
\newblock
\showISBNx{978-3-319-46474-9}
\urldef\tempurl%
\url{https://doi.org/10.1007/978-3-319-46475-6_43}
\showDOI{\tempurl}


\bibitem[Kim et~al\mbox{.}(2021)]%
        {Kim2021}
\bibfield{author}{\bibinfo{person}{Minjung Kim}, \bibinfo{person}{Maryam Azimi}, {and} \bibinfo{person}{Rafa{\l}~K. Mantiuk}.} \bibinfo{year}{2021}\natexlab{}.
\newblock \showarticletitle{{Color Threshold Functions: Application of Contrast Sensitivity Functions in Standard and High Dynamic Range Color Spaces}}.
\newblock \bibinfo{journal}{\emph{Electronic Imaging}} \bibinfo{volume}{33}, \bibinfo{number}{11} (\bibinfo{date}{jan} \bibinfo{year}{2021}), \bibinfo{pages}{153--1--153--7}.
\newblock
\showISSN{2470-1173}
\urldef\tempurl%
\url{https://doi.org/10.2352/ISSN.2470-1173.2021.11.HVEI-153}
\showDOI{\tempurl}


\bibitem[Laird et~al\mbox{.}(2006)]%
        {Laird2006}
\bibfield{author}{\bibinfo{person}{Justin Laird}, \bibinfo{person}{Mitchell Rosen}, \bibinfo{person}{Jeff Pelz}, \bibinfo{person}{Ethan Montag}, {and} \bibinfo{person}{Scott Daly}.} \bibinfo{year}{2006}\natexlab{}.
\newblock \showarticletitle{{Spatio-velocity CSF as a function of retinal velocity using unstabilized stimuli}}. In \bibinfo{booktitle}{\emph{SPIE 6057, Human Vision and Electronic Imaging XI}}.
\newblock


\bibitem[Laparra et~al\mbox{.}(2016)]%
        {Laparra_Simoncelli_2016}
\bibfield{author}{\bibinfo{person}{Valero Laparra}, \bibinfo{person}{Johannes Ballé}, \bibinfo{person}{Alexander Berardino}, {and} \bibinfo{person}{Eero~P Simoncelli}.} \bibinfo{year}{2016}\natexlab{}.
\newblock \showarticletitle{Perceptual image quality assessment using a normalized Laplacian pyramid}.
\newblock \bibinfo{journal}{\emph{Electronic Imaging}} \bibinfo{volume}{28}, \bibinfo{number}{16} (\bibinfo{year}{2016}), \bibinfo{pages}{1–6}.
\newblock
\showISSN{2470-1173}
\urldef\tempurl%
\url{https://doi.org/10.2352/ISSN.2470-1173.2016.16.HVEI-103}
\showDOI{\tempurl}


\bibitem[Li et~al\mbox{.}(2016a)]%
        {Li2016}
\bibfield{author}{\bibinfo{person}{Zhi Li}, \bibinfo{person}{Anne Aaron}, \bibinfo{person}{Ioannis Katsavounidis}, \bibinfo{person}{Anush Moorthy}, {and} \bibinfo{person}{Megha Manohara}.} \bibinfo{year}{2016}\natexlab{a}.
\newblock \bibinfo{booktitle}{\emph{{Toward A Practical Perceptual Video Quality Metric}}}.
\newblock \bibinfo{type}{{T}echnical {R}eport}. \bibinfo{institution}{The NETFLIX Tech Blog}.
\newblock
\urldef\tempurl%
\url{https://netflixtechblog.com/toward-a-practical-perceptual-video-quality-metric-653f208b9652}
\showURL{%
\tempurl}


\bibitem[Li et~al\mbox{.}(2016b)]%
        {vmaf1}
\bibfield{author}{\bibinfo{person}{Zhi Li}, \bibinfo{person}{Anne Aaron}, \bibinfo{person}{Ioannis Katsavounidis}, \bibinfo{person}{Anush Moorthy}, {and} \bibinfo{person}{Megha Manohara}.} \bibinfo{year}{2016}\natexlab{b}.
\newblock \bibinfo{title}{Toward A Practical Perceptual Video Quality Metric}.
\newblock
\newblock
\urldef\tempurl%
\url{https://netflixtechblog.com/toward-a-practical-perceptual-video-quality-metric-653f208b9652}
\showURL{%
\tempurl}


\bibitem[Lin et~al\mbox{.}(2019)]%
        {Lin2019}
\bibfield{author}{\bibinfo{person}{Hanhe Lin}, \bibinfo{person}{Vlad Hosu}, {and} \bibinfo{person}{Dietmar Saupe}.} \bibinfo{year}{2019}\natexlab{}.
\newblock \showarticletitle{{KADID-10k: A Large-scale Artificially Distorted IQA Database}}. In \bibinfo{booktitle}{\emph{2019 Eleventh International Conference on Quality of Multimedia Experience (QoMEX)}}, Vol.~\bibinfo{volume}{161}. \bibinfo{publisher}{IEEE}, \bibinfo{pages}{1--3}.
\newblock
\showISBNx{978-1-5386-8212-8}
\urldef\tempurl%
\url{https://doi.org/10.1109/QoMEX.2019.8743252}
\showDOI{\tempurl}


\bibitem[Mantiuk et~al\mbox{.}(2022)]%
        {mantiuk2022stelacsf}
\bibfield{author}{\bibinfo{person}{Rafa\l{}~K. Mantiuk}, \bibinfo{person}{Maliha Ashraf}, {and} \bibinfo{person}{Alexandre Chapiro}.} \bibinfo{year}{2022}\natexlab{}.
\newblock \showarticletitle{StelaCSF: A Unified Model of Contrast Sensitivity as the Function of Spatio-Temporal Frequency, Eccentricity, Luminance and Area}.
\newblock \bibinfo{journal}{\emph{ACM Trans. Graph.}} \bibinfo{volume}{41}, \bibinfo{number}{4}, Article \bibinfo{articleno}{145} (\bibinfo{date}{Jul} \bibinfo{year}{2022}), \bibinfo{numpages}{16}~pages.
\newblock
\showISSN{0730-0301}
\urldef\tempurl%
\url{https://doi.org/10.1145/3528223.3530115}
\showDOI{\tempurl}


\bibitem[Mantiuk and Azimi(2021)]%
        {Mantiuk2021pu}
\bibfield{author}{\bibinfo{person}{Rafal~K. Mantiuk} {and} \bibinfo{person}{Maryam Azimi}.} \bibinfo{year}{2021}\natexlab{}.
\newblock \showarticletitle{{PU21: A novel perceptually uniform encoding for adapting existing quality metrics for HDR}}. In \bibinfo{booktitle}{\emph{2021 Picture Coding Symposium (PCS)}}. \bibinfo{publisher}{IEEE}, \bibinfo{pages}{1--5}.
\newblock
\showISBNx{978-1-6654-2545-2}
\urldef\tempurl%
\url{https://doi.org/10.1109/PCS50896.2021.9477471}
\showDOI{\tempurl}


\bibitem[Mantiuk et~al\mbox{.}(2021)]%
        {Mantiuk2021}
\bibfield{author}{\bibinfo{person}{Rafa{\l}~K. Mantiuk}, \bibinfo{person}{Gyorgy Denes}, \bibinfo{person}{Alexandre Chapiro}, \bibinfo{person}{Anton Kaplanyan}, \bibinfo{person}{Gizem Rufo}, \bibinfo{person}{Romain Bachy}, \bibinfo{person}{Trisha Lian}, {and} \bibinfo{person}{Anjul Patney}.} \bibinfo{year}{2021}\natexlab{}.
\newblock \showarticletitle{{FovVideoVDP : A visible difference predictor for wide field-of-view video}}.
\newblock \bibinfo{journal}{\emph{ACM Transaction on Graphics}} \bibinfo{volume}{40}, \bibinfo{number}{4} (\bibinfo{year}{2021}), \bibinfo{pages}{49}.
\newblock
\urldef\tempurl%
\url{https://doi.org/10.1145/3450626.3459831}
\showDOI{\tempurl}


\bibitem[Mantiuk et~al\mbox{.}(2023)]%
        {Mantiuk2023}
\bibfield{author}{\bibinfo{person}{Rafal~K. Mantiuk}, \bibinfo{person}{Dounia Hammou}, {and} \bibinfo{person}{Param Hanji}.} \bibinfo{year}{2023}\natexlab{}.
\newblock \showarticletitle{{HDR-VDP-3: A multi-metric for predicting image differences, quality and contrast distortions in high dynamic range and regular content}}.
\newblock  (\bibinfo{date}{apr} \bibinfo{year}{2023}).
\newblock
\showeprint[arxiv]{2304.13625}
\urldef\tempurl%
\url{http://arxiv.org/abs/2304.13625}
\showURL{%
\tempurl}


\bibitem[Mantiuk et~al\mbox{.}(2011)]%
        {Mantiuk2011a}
\bibfield{author}{\bibinfo{person}{Rafa{\l}~K. Mantiuk}, \bibinfo{person}{Kil~Joong Kim}, \bibinfo{person}{Allan~G. Rempel}, {and} \bibinfo{person}{Wolfgang Heidrich}.} \bibinfo{year}{2011}\natexlab{}.
\newblock \showarticletitle{{HDR-VDP-2: A calibrated visual metric for visibility and quality predictions in all luminance conditions}}.
\newblock \bibinfo{journal}{\emph{ACM Transactions on Graphics}} \bibinfo{volume}{30}, \bibinfo{number}{4} (\bibinfo{date}{July} \bibinfo{year}{2011}), \bibinfo{pages}{1--14}.
\newblock
\showISBNx{978-1-4503-0943-1}
\showISSN{07300301}


\bibitem[McKeefry et~al\mbox{.}(2001)]%
        {McKeefry2001}
\bibfield{author}{\bibinfo{person}{D.J McKeefry}, \bibinfo{person}{I.J Murray}, {and} \bibinfo{person}{J.J Kulikowski}.} \bibinfo{year}{2001}\natexlab{}.
\newblock \showarticletitle{{Red–green and blue–yellow mechanisms are matched in sensitivity for temporal and spatial modulation}}.
\newblock \bibinfo{journal}{\emph{Vision Research}} \bibinfo{volume}{41}, \bibinfo{number}{2} (\bibinfo{date}{jan} \bibinfo{year}{2001}), \bibinfo{pages}{245--255}.
\newblock
\showISSN{00426989}
\urldef\tempurl%
\url{https://doi.org/10.1016/S0042-6989(00)00247-9}
\showDOI{\tempurl}


\bibitem[Mikhailiuk et~al\mbox{.}(2022)]%
        {Mikhailiuk2021}
\bibfield{author}{\bibinfo{person}{Aliaksei Mikhailiuk}, \bibinfo{person}{Maria Perez-Ortiz}, \bibinfo{person}{Dingcheng Yue}, \bibinfo{person}{Wilson Suen}, {and} \bibinfo{person}{Rafal Mantiuk}.} \bibinfo{year}{2022}\natexlab{}.
\newblock \showarticletitle{{Consolidated Dataset and Metrics for High-Dynamic-Range Image Quality}}.
\newblock \bibinfo{journal}{\emph{IEEE Transactions on Multimedia}}  \bibinfo{volume}{24} (\bibinfo{year}{2022}), \bibinfo{pages}{2125--2138}.
\newblock
\showISSN{1520-9210}
\urldef\tempurl%
\url{https://doi.org/10.1109/TMM.2021.3076298}
\showDOI{\tempurl}


\bibitem[Mikhailiuk et~al\mbox{.}(2021)]%
        {mikhailiuk2021active}
\bibfield{author}{\bibinfo{person}{Aliaksei Mikhailiuk}, \bibinfo{person}{Clifford Wilmot}, \bibinfo{person}{Maria Perez-Ortiz}, \bibinfo{person}{Dingcheng Yue}, {and} \bibinfo{person}{Rafa{\l}~K Mantiuk}.} \bibinfo{year}{2021}\natexlab{}.
\newblock \showarticletitle{Active sampling for pairwise comparisons via approximate message passing and information gain maximization}. In \bibinfo{booktitle}{\emph{2020 25th International Conference on Pattern Recognition (ICPR)}}. IEEE, \bibinfo{pages}{2559--2566}.
\newblock


\bibitem[Mustafa et~al\mbox{.}(2022)]%
        {Mustafa_2022}
\bibfield{author}{\bibinfo{person}{Aamir Mustafa}, \bibinfo{person}{Aliaksei Mikhailiuk}, \bibinfo{person}{Dan~Andrei Iliescu}, \bibinfo{person}{Varun Babbar}, {and} \bibinfo{person}{Rafal~K. Mantiuk}.} \bibinfo{year}{2022}\natexlab{}.
\newblock \showarticletitle{Training a Task-Specific Image Reconstruction Loss}. In \bibinfo{booktitle}{\emph{2022 IEEE/CVF Winter Conference on Applications of Computer Vision (WACV)}}. \bibinfo{publisher}{IEEE}, \bibinfo{address}{Waikoloa, HI, USA}, \bibinfo{pages}{21–30}.
\newblock
\showISBNx{978-1-66540-915-5}
\urldef\tempurl%
\url{https://doi.org/10.1109/WACV51458.2022.00010}
\showDOI{\tempurl}


\bibitem[Ooi and Dingliana(2022)]%
        {ooi2022color}
\bibfield{author}{\bibinfo{person}{Chun~Wei Ooi} {and} \bibinfo{person}{John Dingliana}.} \bibinfo{year}{2022}\natexlab{}.
\newblock \showarticletitle{Color LightField: Estimation Of View-point Dependant Color Dispersion In Waveguide Display}.
\newblock In \bibinfo{booktitle}{\emph{SIGGRAPH Asia Posters}}. \bibinfo{pages}{1--2}.
\newblock


\bibitem[Peli(1990)]%
        {Peli1990}
\bibfield{author}{\bibinfo{person}{Eli Peli}.} \bibinfo{year}{1990}\natexlab{}.
\newblock \showarticletitle{{Contrast in complex images}}.
\newblock \bibinfo{journal}{\emph{Journal of the Optical Society of America A}} \bibinfo{volume}{7}, \bibinfo{number}{10} (\bibinfo{date}{oct} \bibinfo{year}{1990}), \bibinfo{pages}{2032--40}.
\newblock
\showISSN{0740-3232}
\urldef\tempurl%
\url{https://doi.org/10.1364/JOSAA.7.002032}
\showDOI{\tempurl}


\bibitem[Peli(1995)]%
        {Peli1995}
\bibfield{author}{\bibinfo{person}{Eli Peli}.} \bibinfo{year}{1995}\natexlab{}.
\newblock \showarticletitle{{Suprathreshold contrast perception across differences in mean luminance: effects of stimulus size, dichoptic presentation, and length of adaptation}}.
\newblock \bibinfo{journal}{\emph{JOSA A}} \bibinfo{volume}{12}, \bibinfo{number}{5} (\bibinfo{year}{1995}), \bibinfo{pages}{817}.
\newblock
\showISSN{1084-7529}
\urldef\tempurl%
\url{https://doi.org/10.1364/JOSAA.12.000817}
\showDOI{\tempurl}


\bibitem[Perez-Ortiz et~al\mbox{.}(2019)]%
        {perez2019pairwise}
\bibfield{author}{\bibinfo{person}{Maria Perez-Ortiz}, \bibinfo{person}{Aliaksei Mikhailiuk}, \bibinfo{person}{Emin Zerman}, \bibinfo{person}{Vedad Hulusic}, \bibinfo{person}{Giuseppe Valenzise}, {and} \bibinfo{person}{Rafa{\l}~K Mantiuk}.} \bibinfo{year}{2019}\natexlab{}.
\newblock \showarticletitle{From pairwise comparisons and rating to a unified quality scale}.
\newblock \bibinfo{journal}{\emph{IEEE Transactions on Image Processing}}  \bibinfo{volume}{29} (\bibinfo{year}{2019}), \bibinfo{pages}{1139--1151}.
\newblock


\bibitem[Prashnani et~al\mbox{.}(2018)]%
        {Prashnani2018}
\bibfield{author}{\bibinfo{person}{Ekta Prashnani}, \bibinfo{person}{Hong Cai}, \bibinfo{person}{Yasamin Mostofi}, {and} \bibinfo{person}{Pradeep Sen}.} \bibinfo{year}{2018}\natexlab{}.
\newblock \showarticletitle{{PieAPP: Perceptual Image-Error Assessment Through Pairwise Preference}}. In \bibinfo{booktitle}{\emph{2018 IEEE/CVF Conference on Computer Vision and Pattern Recognition}}. \bibinfo{publisher}{IEEE}, \bibinfo{pages}{1808--1817}.
\newblock
\showISBNx{978-1-5386-6420-9}
\urldef\tempurl%
\url{https://doi.org/10.1109/CVPR.2018.00194}
\showDOI{\tempurl}


\bibitem[Quick et~al\mbox{.}(1978)]%
        {Quick1978}
\bibfield{author}{\bibinfo{person}{R.~Frank Quick}, \bibinfo{person}{W.~W. Mullins}, {and} \bibinfo{person}{T.~A. Reichert}.} \bibinfo{year}{1978}\natexlab{}.
\newblock \showarticletitle{{Spatial summation effects on two-component grating thresholds}}.
\newblock \bibinfo{journal}{\emph{Journal of the Optical Society of America}} \bibinfo{volume}{68}, \bibinfo{number}{1} (\bibinfo{date}{jan} \bibinfo{year}{1978}), \bibinfo{pages}{116}.
\newblock
\showISSN{0030-3941}
\urldef\tempurl%
\url{https://doi.org/10.1364/JOSA.68.000116}
\showDOI{\tempurl}


\bibitem[Schuetz and Fiehler(2022)]%
        {schuetz2022eye}
\bibfield{author}{\bibinfo{person}{Immo Schuetz} {and} \bibinfo{person}{Katja Fiehler}.} \bibinfo{year}{2022}\natexlab{}.
\newblock \showarticletitle{Eye tracking in virtual reality: Vive pro eye spatial accuracy, precision, and calibration reliability}.
\newblock \bibinfo{journal}{\emph{Journal of Eye Movement Research}} \bibinfo{volume}{15}, \bibinfo{number}{3} (\bibinfo{year}{2022}).
\newblock


\bibitem[Seshadrinathan et~al\mbox{.}(2010)]%
        {Seshadrinathan2010}
\bibfield{author}{\bibinfo{person}{Kalpana Seshadrinathan}, \bibinfo{person}{Rajiv Soundararajan}, \bibinfo{person}{Alan~Conrad Bovik}, {and} \bibinfo{person}{Lawrence~K. Cormack}.} \bibinfo{year}{2010}\natexlab{}.
\newblock \showarticletitle{{Study of Subjective and Objective Quality Assessment of Video}}.
\newblock \bibinfo{journal}{\emph{IEEE Transactions on Image Processing}} \bibinfo{volume}{19}, \bibinfo{number}{6} (\bibinfo{date}{jun} \bibinfo{year}{2010}), \bibinfo{pages}{1427--1441}.
\newblock
\showISSN{1057-7149}
\urldef\tempurl%
\url{https://doi.org/10.1109/TIP.2010.2042111}
\showDOI{\tempurl}


\bibitem[Shang et~al\mbox{.}(2022)]%
        {Shang2022}
\bibfield{author}{\bibinfo{person}{Zaixi Shang}, \bibinfo{person}{Joshua~P. Ebenezer}, \bibinfo{person}{Alan~C. Bovik}, \bibinfo{person}{Yongjun Wu}, \bibinfo{person}{Hai Wei}, {and} \bibinfo{person}{Sriram Sethuraman}.} \bibinfo{year}{2022}\natexlab{}.
\newblock \showarticletitle{{Subjective Assessment Of High Dynamic Range Videos Under Different Ambient Conditions}}. In \bibinfo{booktitle}{\emph{2022 IEEE International Conference on Image Processing (ICIP)}}. \bibinfo{publisher}{IEEE}, \bibinfo{pages}{786--790}.
\newblock
\showISBNx{978-1-6654-9620-9}
\urldef\tempurl%
\url{https://doi.org/10.1109/ICIP46576.2022.9897940}
\showDOI{\tempurl}


\bibitem[Sharma et~al\mbox{.}(2005)]%
        {Sharma2005}
\bibfield{author}{\bibinfo{person}{Gaurav Sharma}, \bibinfo{person}{Wencheng Wu}, {and} \bibinfo{person}{Edul~N. Dalal}.} \bibinfo{year}{2005}\natexlab{}.
\newblock \showarticletitle{{The CIEDE2000 color-difference formula: Implementation notes, supplementary test data, and mathematical observations}}.
\newblock \bibinfo{journal}{\emph{Color Research \& Application}} \bibinfo{volume}{30}, \bibinfo{number}{1} (\bibinfo{date}{feb} \bibinfo{year}{2005}), \bibinfo{pages}{21--30}.
\newblock
\showISSN{0361-2317}
\urldef\tempurl%
\url{https://doi.org/10.1002/col.20070}
\showDOI{\tempurl}


\bibitem[Soundararajan and Bovik(2012)]%
        {soundararajan2012video}
\bibfield{author}{\bibinfo{person}{Rajiv Soundararajan} {and} \bibinfo{person}{Alan~C Bovik}.} \bibinfo{year}{2012}\natexlab{}.
\newblock \showarticletitle{Video quality assessment by reduced reference spatio-temporal entropic differencing}.
\newblock \bibinfo{journal}{\emph{IEEE Transactions on Circuits and Systems for Video Technology}} \bibinfo{volume}{23}, \bibinfo{number}{4} (\bibinfo{year}{2012}), \bibinfo{pages}{684--694}.
\newblock


\bibitem[Stockman and Brainard(2010)]%
        {Stockman2010}
\bibfield{author}{\bibinfo{person}{Andrew Stockman} {and} \bibinfo{person}{David~H. Brainard}.} \bibinfo{year}{2010}\natexlab{}.
\newblock \showarticletitle{{Color vision mechanisms}}.
\newblock In \bibinfo{booktitle}{\emph{OSA handbook of optics}}. \bibinfo{pages}{11}.
\newblock


\bibitem[Stromeyer and Julesz(1972)]%
        {stromeyer1972spatial}
\bibfield{author}{\bibinfo{person}{C.~F. Stromeyer} {and} \bibinfo{person}{B. Julesz}.} \bibinfo{year}{1972}\natexlab{}.
\newblock \showarticletitle{{Spatial-Frequency Masking in Vision: Critical Bands and Spread of Masking}}.
\newblock \bibinfo{journal}{\emph{Journal of the Optical Society of America}} \bibinfo{volume}{62}, \bibinfo{number}{10} (\bibinfo{date}{oct} \bibinfo{year}{1972}), \bibinfo{pages}{1221}.
\newblock
\showISSN{0030-3941}
\urldef\tempurl%
\url{https://doi.org/10.1364/JOSA.62.001221}
\showDOI{\tempurl}


\bibitem[Switkes et~al\mbox{.}(1988)]%
        {Switkes_1988}
\bibfield{author}{\bibinfo{person}{Eugene Switkes}, \bibinfo{person}{Arthur Bradley}, {and} \bibinfo{person}{Karen~K. De~Valois}.} \bibinfo{year}{1988}\natexlab{}.
\newblock \showarticletitle{Contrast dependence and mechanisms of masking interactions among chromatic and luminance gratings}.
\newblock \bibinfo{journal}{\emph{Journal of the Optical Society of America A}} \bibinfo{volume}{5}, \bibinfo{number}{7} (\bibinfo{year}{1988}), \bibinfo{pages}{1149}.
\newblock
\showISSN{1084-7529}
\urldef\tempurl%
\url{https://doi.org/10.1364/josaa.5.001149}
\showDOI{\tempurl}


\bibitem[Switkes and Crognale(1999)]%
        {Switkes1999}
\bibfield{author}{\bibinfo{person}{Eugene Switkes} {and} \bibinfo{person}{Michael~A. Crognale}.} \bibinfo{year}{1999}\natexlab{}.
\newblock \showarticletitle{{Comparison of color and luminance contrast: apples versus oranges?}}
\newblock \bibinfo{journal}{\emph{Vision Research}} \bibinfo{volume}{39}, \bibinfo{number}{10} (\bibinfo{date}{may} \bibinfo{year}{1999}), \bibinfo{pages}{1823--1831}.
\newblock
\showISSN{00426989}
\urldef\tempurl%
\url{https://doi.org/10.1016/S0042-6989(98)00219-3}
\showDOI{\tempurl}


\bibitem[Venkataramanan et~al\mbox{.}(2022)]%
        {Venkataramanan_2022}
\bibfield{author}{\bibinfo{person}{Abhinau~K. Venkataramanan}, \bibinfo{person}{Cosmin Stejerean}, {and} \bibinfo{person}{Alan~C. Bovik}.} \bibinfo{year}{2022}\natexlab{}.
\newblock \showarticletitle{Funque: Fusion of Unified Quality Evaluators}. In \bibinfo{booktitle}{\emph{2022 IEEE International Conference on Image Processing (ICIP)}}. \bibinfo{publisher}{IEEE}, \bibinfo{address}{Bordeaux, France}, \bibinfo{pages}{2147–2151}.
\newblock
\showISBNx{978-1-66549-620-9}
\urldef\tempurl%
\url{https://doi.org/10.1109/ICIP46576.2022.9897312}
\showDOI{\tempurl}


\bibitem[Venkataramanan et~al\mbox{.}(2023)]%
        {Venkataramanan_2023}
\bibfield{author}{\bibinfo{person}{Abhinau~K. Venkataramanan}, \bibinfo{person}{Cosmin Stejerean}, \bibinfo{person}{Ioannis Katsavounidis}, {and} \bibinfo{person}{Alan~C. Bovik}.} \bibinfo{year}{2023}\natexlab{}.
\newblock \showarticletitle{One Transform To Compute Them All: Efficient Fusion-Based Full-Reference Video Quality Assessment}.
\newblock  \bibinfo{number}{arXiv:2304.03412} (\bibinfo{date}{Nov.} \bibinfo{year}{2023}).
\newblock
\urldef\tempurl%
\url{http://arxiv.org/abs/2304.03412}
\showURL{%
\tempurl}
\newblock
\shownote{arXiv:2304.03412 [eess]}.


\bibitem[Wang et~al\mbox{.}(2003)]%
        {Wang2003d}
\bibfield{author}{\bibinfo{person}{Z Wang}, \bibinfo{person}{E.P. Simoncelli}, {and} \bibinfo{person}{A.C. Bovik}.} \bibinfo{year}{2003}\natexlab{}.
\newblock \showarticletitle{{Multiscale structural similarity for image quality assessment}}. In \bibinfo{booktitle}{\emph{The Thrity-Seventh Asilomar Conference on Signals, Systems {\&} Computers, 2003}}. \bibinfo{publisher}{IEEE}, \bibinfo{pages}{1398--1402}.
\newblock
\showISBNx{0-7803-8104-1}
\urldef\tempurl%
\url{https://doi.org/10.1109/ACSSC.2003.1292216}
\showDOI{\tempurl}


\bibitem[Watson and Solomon(1997)]%
        {Watson1997}
\bibfield{author}{\bibinfo{person}{AB Watson} {and} \bibinfo{person}{JA Solomon}.} \bibinfo{year}{1997}\natexlab{}.
\newblock \showarticletitle{{Model of visual contrast gain control and pattern masking}}.
\newblock \bibinfo{journal}{\emph{Journal of the Optical Society of America A}} \bibinfo{volume}{14}, \bibinfo{number}{9} (\bibinfo{year}{1997}), \bibinfo{pages}{2379--2391}.
\newblock
\urldef\tempurl%
\url{http://www.opticsinfobase.org/abstract.cfm?URI=josaa-14-9-2379}
\showURL{%
\tempurl}


\bibitem[Watson(1993)]%
        {Watson1993}
\bibfield{author}{\bibinfo{person}{Andrew~B Watson}.} \bibinfo{year}{1993}\natexlab{}.
\newblock \showarticletitle{{DCTune: a technique for visual optimization of DCT quantization matrices for individual images}}. In \bibinfo{booktitle}{\emph{Society for Information Display Digest of Technical Papers XXIV}}. \bibinfo{pages}{946--949}.
\newblock


\bibitem[Watson et~al\mbox{.}(1983)]%
        {Watson1983}
\bibfield{author}{\bibinfo{person}{Andrew~B. Watson}, \bibinfo{person}{H.~B. Barlow}, {and} \bibinfo{person}{John~G. Robson}.} \bibinfo{year}{1983}\natexlab{}.
\newblock \showarticletitle{{What does the eye see best?}}
\newblock \bibinfo{journal}{\emph{Nature}} \bibinfo{volume}{302}, \bibinfo{number}{5907} (\bibinfo{date}{mar} \bibinfo{year}{1983}), \bibinfo{pages}{419--422}.
\newblock
\showISSN{0028-0836}
\urldef\tempurl%
\url{https://doi.org/10.1038/302419a0}
\showDOI{\tempurl}


\bibitem[Wuerger et~al\mbox{.}(2020)]%
        {Wuerger2020}
\bibfield{author}{\bibinfo{person}{Sophie Wuerger}, \bibinfo{person}{Maliha Ashraf}, \bibinfo{person}{Minjung Kim}, \bibinfo{person}{Jasna Martinovic}, \bibinfo{person}{Mar{\'{i}}a P{\'{e}}rez-Ortiz}, {and} \bibinfo{person}{Rafa{\l}~K. Mantiuk}.} \bibinfo{year}{2020}\natexlab{}.
\newblock \showarticletitle{{Spatio-chromatic contrast sensitivity under mesopic and photopic light levels}}.
\newblock \bibinfo{journal}{\emph{Journal of Vision}} \bibinfo{volume}{20}, \bibinfo{number}{4} (\bibinfo{date}{April} \bibinfo{year}{2020}), \bibinfo{pages}{23}.
\newblock
\showISSN{1534-7362}


\bibitem[Ye et~al\mbox{.}(2019)]%
        {Ye2019}
\bibfield{author}{\bibinfo{person}{Nanyang Ye}, \bibinfo{person}{Krzysztof Wolski}, {and} \bibinfo{person}{Rafal~K. Mantiuk}.} \bibinfo{year}{2019}\natexlab{}.
\newblock \showarticletitle{{Predicting Visible Image Differences Under Varying Display Brightness and Viewing Distance}}. In \bibinfo{booktitle}{\emph{2019 IEEE/CVF Conference on Computer Vision and Pattern Recognition (CVPR)}}. \bibinfo{publisher}{IEEE}, \bibinfo{pages}{5429--5437}.
\newblock
\showISBNx{978-1-7281-3293-8}
\urldef\tempurl%
\url{https://doi.org/10.1109/CVPR.2019.00558}
\showDOI{\tempurl}


\bibitem[Zerman et~al\mbox{.}(2018)]%
        {zerman2018relation}
\bibfield{author}{\bibinfo{person}{Emin Zerman}, \bibinfo{person}{Vedad Hulusic}, \bibinfo{person}{Giuseppe Valenzise}, \bibinfo{person}{Rafa{\l}~K Mantiuk}, {and} \bibinfo{person}{Fr{\'e}d{\'e}ric Dufaux}.} \bibinfo{year}{2018}\natexlab{}.
\newblock \showarticletitle{The relation between MOS and pairwise comparisons and the importance of cross-content comparisons}.
\newblock \bibinfo{journal}{\emph{Electronic Imaging}} \bibinfo{volume}{2018}, \bibinfo{number}{14} (\bibinfo{year}{2018}), \bibinfo{pages}{1--6}.
\newblock


\bibitem[Zhang et~al\mbox{.}(2014)]%
        {Zhang2014}
\bibfield{author}{\bibinfo{person}{Lin Zhang}, \bibinfo{person}{Ying Shen}, {and} \bibinfo{person}{Hongyu Li}.} \bibinfo{year}{2014}\natexlab{}.
\newblock \showarticletitle{{VSI: A Visual Saliency-Induced Index for Perceptual Image Quality Assessment}}.
\newblock \bibinfo{journal}{\emph{IEEE Transactions on Image Processing}} \bibinfo{volume}{23}, \bibinfo{number}{10} (\bibinfo{date}{oct} \bibinfo{year}{2014}), \bibinfo{pages}{4270--4281}.
\newblock
\showISSN{1057-7149}
\urldef\tempurl%
\url{https://doi.org/10.1109/TIP.2014.2346028}
\showDOI{\tempurl}


\bibitem[Zhang et~al\mbox{.}(2011)]%
        {LinZhang2011}
\bibfield{author}{\bibinfo{person}{Lin Zhang}, \bibinfo{person}{Lei Zhang}, \bibinfo{person}{Xuanqin Mou}, {and} \bibinfo{person}{David Zhang}.} \bibinfo{year}{2011}\natexlab{}.
\newblock \showarticletitle{{FSIM: A Feature Similarity Index for Image Quality Assessment}}.
\newblock \bibinfo{journal}{\emph{IEEE Transactions on Image Processing}} \bibinfo{volume}{20}, \bibinfo{number}{8} (\bibinfo{date}{aug} \bibinfo{year}{2011}), \bibinfo{pages}{2378--2386}.
\newblock
\showISSN{1057-7149}
\urldef\tempurl%
\url{https://doi.org/10.1109/TIP.2011.2109730}
\showDOI{\tempurl}


\bibitem[Zhang et~al\mbox{.}(2018)]%
        {zhang2018perceptual}
\bibfield{author}{\bibinfo{person}{Richard Zhang}, \bibinfo{person}{Phillip Isola}, \bibinfo{person}{Alexei~A Efros}, \bibinfo{person}{Eli Shechtman}, {and} \bibinfo{person}{Oliver Wang}.} \bibinfo{year}{2018}\natexlab{}.
\newblock \showarticletitle{The Unreasonable Effectiveness of Deep Features as a Perceptual Metric}. In \bibinfo{booktitle}{\emph{CVPR}}. \bibinfo{pages}{586--595}.
\newblock
\urldef\tempurl%
\url{https://doi.org/10.1109/CVPR.2018.00068}
\showDOI{\tempurl}


\bibitem[Zhang and Wandell(1997)]%
        {Zhang1997a}
\bibfield{author}{\bibinfo{person}{X. Zhang} {and} \bibinfo{person}{B.~A. Wandell}.} \bibinfo{year}{1997}\natexlab{}.
\newblock \showarticletitle{{A spatial extension of CIELAB for digital color-image reproduction}}.
\newblock \bibinfo{journal}{\emph{Journal of the Society for Information Display}} \bibinfo{volume}{5}, \bibinfo{number}{1} (\bibinfo{year}{1997}), \bibinfo{pages}{61}.
\newblock
\showISSN{10710922}
\urldef\tempurl%
\url{https://doi.org/10.1889/1.1985127}
\showDOI{\tempurl}


\end{thebibliography}



\begin{thebibliography}{13}


\ifx \showCODEN    \undefined \def \showCODEN     #1{\unskip}     \fi
\ifx \showDOI      \undefined \def \showDOI       #1{#1}\fi
\ifx \showISBNx    \undefined \def \showISBNx     #1{\unskip}     \fi
\ifx \showISBNxiii \undefined \def \showISBNxiii  #1{\unskip}     \fi
\ifx \showISSN     \undefined \def \showISSN      #1{\unskip}     \fi
\ifx \showLCCN     \undefined \def \showLCCN      #1{\unskip}     \fi
\ifx \shownote     \undefined \def \shownote      #1{#1}          \fi
\ifx \showarticletitle \undefined \def \showarticletitle #1{#1}   \fi
\ifx \showURL      \undefined \def \showURL       {\relax}        \fi
\providecommand\bibfield[2]{#2}
\providecommand\bibinfo[2]{#2}
\providecommand\natexlab[1]{#1}
\providecommand\showeprint[2][]{arXiv:#2}

\bibitem[Ashraf et~al\mbox{.}(2024)]%
        {Ashraf2024}
\bibfield{author}{\bibinfo{person}{Maliha Ashraf}, \bibinfo{person}{Rafał~K. Mantiuk}, \bibinfo{person}{Alexandre Chapiro}, {and} \bibinfo{person}{Sophie Wuerger}.} \bibinfo{year}{2024}\natexlab{}.
\newblock \showarticletitle{castleCSF — A Contrast Sensitivity Function of Color, Area, Spatio-Temporal frequency, Luminance and Eccentricity}.
\newblock \bibinfo{journal}{\emph{Journal of Vision}}  \bibinfo{volume}{24} (\bibinfo{year}{2024}), \bibinfo{pages}{5}.
\newblock
\urldef\tempurl%
\url{https://doi.org/10.1167/jov.24.4.5}
\showDOI{\tempurl}


\bibitem[Barten(1999)]%
        {Barten1999}
\bibfield{author}{\bibinfo{person}{Peter G.~J. Barten}.} \bibinfo{year}{1999}\natexlab{}.
\newblock \bibinfo{booktitle}{\emph{{Contrast sensitivity of the human eye and its effects on image quality}}}.
\newblock \bibinfo{publisher}{SPIE Press}. 208 pages.
\newblock
\showISBNx{0819434965}
\urldef\tempurl%
\url{http://books.google.com/books?hl=en&lr=&id=kPyyBAomC4cC&pgis=1}
\showURL{%
\tempurl}


\bibitem[Daly(1993)]%
        {Daly1993}
\bibfield{author}{\bibinfo{person}{S.J. Daly}.} \bibinfo{year}{1993}\natexlab{}.
\newblock \showarticletitle{{Visible differences predictor: an algorithm for the assessment of image fidelity}}.
\newblock In \bibinfo{booktitle}{\emph{Digital Images and Human Vision}}, \bibfield{editor}{\bibinfo{person}{Andrew~B. Watson}} (Ed.). Vol.~\bibinfo{volume}{1666}. \bibinfo{publisher}{MIT Press}, \bibinfo{pages}{179--206}.
\newblock
\showISBNx{978-0262231718}
\showISSN{0277786X}
\urldef\tempurl%
\url{https://doi.org/10.1117/12.135952}
\showDOI{\tempurl}


\bibitem[Foley(1994)]%
        {Foley1994a}
\bibfield{author}{\bibinfo{person}{John~M. Foley}.} \bibinfo{year}{1994}\natexlab{}.
\newblock \showarticletitle{{Human luminance pattern-vision mechanisms: masking experiments require a new model}}.
\newblock \bibinfo{journal}{\emph{Journal of the Optical Society of America A}} \bibinfo{volume}{11}, \bibinfo{number}{6} (\bibinfo{date}{jun} \bibinfo{year}{1994}), \bibinfo{pages}{1710}.
\newblock
\showISSN{1084-7529}
\urldef\tempurl%
\url{https://doi.org/10.1364/JOSAA.11.001710}
\showDOI{\tempurl}


\bibitem[Georgeson and Sullivan(1975)]%
        {Georgeson1975a}
\bibfield{author}{\bibinfo{person}{B~Y M~A Georgeson} {and} \bibinfo{person}{G~D Sullivan}.} \bibinfo{year}{1975}\natexlab{}.
\newblock \showarticletitle{{Contrast constancy: deblurring in human vision by spatial frequency channels}}.
\newblock \bibinfo{journal}{\emph{The Journal of Physiology}} \bibinfo{volume}{252}, \bibinfo{number}{3} (\bibinfo{year}{1975}), \bibinfo{pages}{627--656}.
\newblock


\bibitem[Georgeson(1991)]%
        {Geor1991}
\bibfield{author}{\bibinfo{person}{MA Georgeson}.} \bibinfo{year}{1991}\natexlab{}.
\newblock \showarticletitle{{Contrast overconstancy}}.
\newblock \bibinfo{journal}{\emph{Journal of the Optical Society of America A}} (\bibinfo{year}{1991}).
\newblock
\urldef\tempurl%
\url{http://www.opticsinfobase.org/josaa/ViewMedia.cfm?id=4026&seq=0}
\showURL{%
\tempurl}


\bibitem[Hess(1990)]%
        {Hess_1990}
\bibfield{author}{\bibinfo{person}{R.~F. Hess}.} \bibinfo{year}{1990}\natexlab{}.
\newblock \showarticletitle{Vision at low light levels: role of spatial, temporal and contrast filters*}.
\newblock \bibinfo{journal}{\emph{Ophthalmic and Physiological Optics}} \bibinfo{volume}{10}, \bibinfo{number}{4} (\bibinfo{date}{Oct.} \bibinfo{year}{1990}), \bibinfo{pages}{351–359}.
\newblock
\showISSN{0275-5408, 1475-1313}
\urldef\tempurl%
\url{https://doi.org/10.1111/j.1475-1313.1990.tb00881.x}
\showDOI{\tempurl}


\bibitem[Kim et~al\mbox{.}(2021)]%
        {Kim2021}
\bibfield{author}{\bibinfo{person}{Minjung Kim}, \bibinfo{person}{Maryam Azimi}, {and} \bibinfo{person}{Rafa{\l}~K. Mantiuk}.} \bibinfo{year}{2021}\natexlab{}.
\newblock \showarticletitle{{Color Threshold Functions: Application of Contrast Sensitivity Functions in Standard and High Dynamic Range Color Spaces}}.
\newblock \bibinfo{journal}{\emph{Electronic Imaging}} \bibinfo{volume}{33}, \bibinfo{number}{11} (\bibinfo{date}{jan} \bibinfo{year}{2021}), \bibinfo{pages}{153--1--153--7}.
\newblock
\showISSN{2470-1173}
\urldef\tempurl%
\url{https://doi.org/10.2352/ISSN.2470-1173.2021.11.HVEI-153}
\showDOI{\tempurl}


\bibitem[Kulikowski(1976)]%
        {Kulikowski1976a}
\bibfield{author}{\bibinfo{person}{J.J. Kulikowski}.} \bibinfo{year}{1976}\natexlab{}.
\newblock \showarticletitle{{Effective contrast constancy and linearity of contrast sensation}}.
\newblock \bibinfo{journal}{\emph{Vision Research}} \bibinfo{volume}{16}, \bibinfo{number}{12} (\bibinfo{date}{jan} \bibinfo{year}{1976}), \bibinfo{pages}{1419--1431}.
\newblock
\showISSN{00426989}
\urldef\tempurl%
\url{https://doi.org/10.1016/0042-6989(76)90161-9}
\showDOI{\tempurl}


\bibitem[Peli(1995)]%
        {Peli1995}
\bibfield{author}{\bibinfo{person}{Eli Peli}.} \bibinfo{year}{1995}\natexlab{}.
\newblock \showarticletitle{{Suprathreshold contrast perception across differences in mean luminance: effects of stimulus size, dichoptic presentation, and length of adaptation}}.
\newblock \bibinfo{journal}{\emph{JOSA A}} \bibinfo{volume}{12}, \bibinfo{number}{5} (\bibinfo{year}{1995}), \bibinfo{pages}{817}.
\newblock
\showISSN{1084-7529}
\urldef\tempurl%
\url{https://doi.org/10.1364/JOSAA.12.000817}
\showDOI{\tempurl}


\bibitem[Peli et~al\mbox{.}(1991)]%
        {Peli1991}
\bibfield{author}{\bibinfo{person}{Eli Peli}, \bibinfo{person}{Jian Yang}, \bibinfo{person}{Robert Goldstein}, {and} \bibinfo{person}{Adam Reeves}.} \bibinfo{year}{1991}\natexlab{}.
\newblock \showarticletitle{{Effect of luminance on suprathreshold contrast perception}}.
\newblock \bibinfo{journal}{\emph{Journal of the Optical Society of America A}} \bibinfo{volume}{8}, \bibinfo{number}{8} (\bibinfo{date}{aug} \bibinfo{year}{1991}), \bibinfo{pages}{1352}.
\newblock
\showISSN{1084-7529}
\urldef\tempurl%
\url{https://doi.org/10.1364/JOSAA.8.001352}
\showDOI{\tempurl}


\bibitem[Switkes and Crognale(1999)]%
        {Switkes1999}
\bibfield{author}{\bibinfo{person}{Eugene Switkes} {and} \bibinfo{person}{Michael~A. Crognale}.} \bibinfo{year}{1999}\natexlab{}.
\newblock \showarticletitle{{Comparison of color and luminance contrast: apples versus oranges?}}
\newblock \bibinfo{journal}{\emph{Vision Research}} \bibinfo{volume}{39}, \bibinfo{number}{10} (\bibinfo{date}{may} \bibinfo{year}{1999}), \bibinfo{pages}{1823--1831}.
\newblock
\showISSN{00426989}
\urldef\tempurl%
\url{https://doi.org/10.1016/S0042-6989(98)00219-3}
\showDOI{\tempurl}


\bibitem[Watson and Solomon(1997)]%
        {Watson1997}
\bibfield{author}{\bibinfo{person}{AB Watson} {and} \bibinfo{person}{JA Solomon}.} \bibinfo{year}{1997}\natexlab{}.
\newblock \showarticletitle{{Model of visual contrast gain control and pattern masking}}.
\newblock \bibinfo{journal}{\emph{Journal of the Optical Society of America A}} \bibinfo{volume}{14}, \bibinfo{number}{9} (\bibinfo{year}{1997}), \bibinfo{pages}{2379--2391}.
\newblock
\urldef\tempurl%
\url{http://www.opticsinfobase.org/abstract.cfm?URI=josaa-14-9-2379}
\showURL{%
\tempurl}


\end{thebibliography}


\end{document}


\title{Supplementary of ColorVideoVDP: A visual difference predictor for image, video and display distortions}

\author{}



\keywords{}

\maketitle


This supplementary document describes:
\begin{itemize}
    \item the procedure used to convert from the cone contrast units used in castleCSF to the DKL contrast used in \ourmethod{} (\secref{cc-to-dkl});
    \item the contrast encoding (\secref{contrast-enc}) and contrast masking functions (\secref{cont-masking-models}) that were considered for \ourmethod{} and are included in the ablations;
    \item several mathematical techniques that were necessary to make the model differentiable (\secref{training-considerations});
    \item timings of \ourmethod{} compared to VMAF (\secref{timings});
    \item the experiment used to obtain a JOD quality scaling for the LIVEHDR dataset (\secref{livehdr-alignment});
    \item details on the content used for XR-DAVID dataset (\secref{xr-david-content}).
\end{itemize}

\section{Contrast sensitivity in the DKL color space}
\label{sec:cc-to-dkl}

The castleCSF \cite{Ashraf2024} model predicts sensitivity in different contrast units than those used by \ourmethod{}. To use castleCSF in \ourmethod{}, we need to convert between the two contrast units. The contrast conversion procedure is similar to the one used in \cite{Kim2021}. 

The sensitivity used in castleCSF is defined as the inverse of cone contrast:
\begin{equation}
     S_\ind{cc} = \sqrt{3}\left(\left( \frac{\coneinc{L}}{\conebkg{L}}\right)^2 + \left( \frac{\coneinc{M}}{\conebkg{M}}\right)^2 + \left( \frac{\coneinc{S}}{\conebkg{S}}\right)^2\right)^{-0.5}\,,
     \label{eq:cone-contrast}
\end{equation}
where \coneinc{L}, \coneinc{M}, and \coneinc{S} are the amplitudes of the cone responses for the stimuli and \conebkg{L}, \conebkg{M}, and \conebkg{S} are the cone responses for the corresponding background. \ourmethod{} encodes contrast for the three cardinal dimensions of the DKL space ($c\in\{\text{Ach},\text{RG},\text{YV}\}$) as:
\begin{equation}
    C_{c} = \frac{\laplpyr_{c}}{L}\,,
     \label{eq:dkl-contrast}
\end{equation}
where $L=\conebkg{L}+\conebkg{M}$ is luminance, and $\laplpyr_{c}$ is the amplitude of the achromatic, red-green, or yellow-violet response in the DKL space (an increment in that space and also the coefficient of the Laplacian pyramid). Here, we omitted unnecessary indices from Eq.~(6) in the main paper. The sensitivity in the DKL-contrast units is the inverse of \eqref{dkl-contrast}:
\begin{equation}
     S_{\ind{DKL}{,c}} = C^{-1}_{c} = \frac{L}{\laplpyr_{c}}\,,
     \label{eq:dkl-sensitivity}
\end{equation}

To convert $S_\ind{cc}$ into $S_{\ind{DKL}{,c}}$, we first obtain from castleCSF the sensitivity $S_\ind{cc}$ along each cardinal color direction ($c$) of the DKL color space. Then, we find the contrast $C_{c}$ that results in the threshold cone contrast corresponding to the predicted sensitivity ($S^{-1}_\ind{cc}$ since the threshold contrast is the inverse of the sensitivity). As there is no closed-form solution, $C_{c}$ needs to be found by non-linear root finding. In the optimization loop, we allow the DKL contrast $C_{c}$ to change only along the given cardinal color direction. This calculation is repeated for each spatial frequency, luminance, and color direction. The sensitivities in the DKL contrast space are precomputed and stored in a look-up table for later use by \ourmethod{}, as explained in Sec.~3.4 of the main paper. 

\section{Contrast encoding}
\label{sec:contrast-enc}



\begin{figure}
    \centering
    \includegraphics[width=\columnwidth]{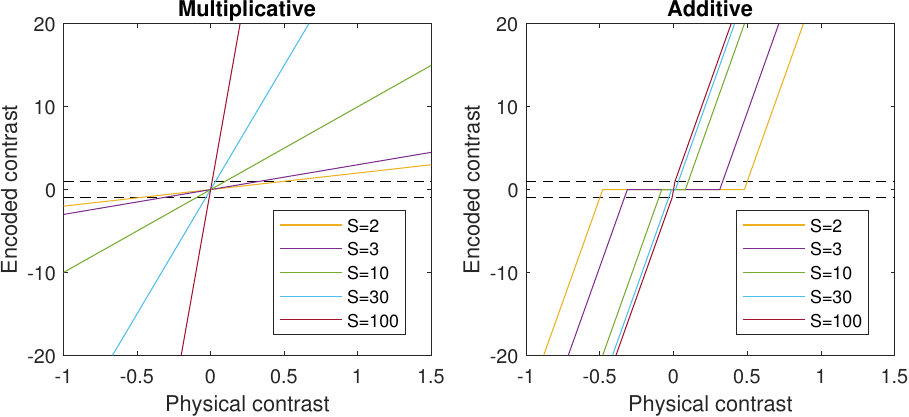}
    \caption{Mapping from input physical contrast to encoded contrast for multiplicative (left) and additive (right) encoding.}
    \label{fig:plot-cont-enc}
\end{figure}

\paragraph{Multiplicative contrast normalization} Once we have separated the band-limited contrast in each frequency band and visual channel (two achromatic and two chromatic), we want to encode it in a way that correlates well with the perceived magnitude of the contrast. To this end, many visual models employ the normalization by the contrast sensitivity function $S(\cdot)$
\begin{equation}
    C^\prime = C\,S(\rho_b,L_\ind{bkg},c)\,,
\end{equation}
where $C$ is the physical and $C^\prime$ is encoded contrast, $\rho_b$ is the peak frequency of the band $b$, $c$ is the channel index, and $L_\ind{bkg}$ is the background luminance. Since contrast can be expressed as $C=\nicefrac{\Delta{L}}{L_\ind{bkg}}$, we can write:
\begin{equation}
    C^\prime = C\,S(\rho_b,L_\ind{bkg},c) = \frac{\Delta{L}}{L_\ind{bkg}}\frac{L_\ind{bkg}}{\Delta{L_\ind{thr}}} = \frac{\Delta{L}}{\Delta{L_\ind{thr}}}\,,
    \label{eq:csf-norm}
\end{equation}
where $\Delta{L_\ind{thr}}$ is the luminance difference corresponding to the detection threshold. It follows that $C^\prime$ encodes multiples of the detection threshold. Notably, it is equal to 1 when the contrast $C$ is exactly at the threshold value. This property is necessary for any contrast encoding employed by our method because masking models, discussed in the next section, rely on it. The encoding is plotted in the left panel of \figref{plot-cont-enc}.


The above contrast normalization by sensitivity brings several benefits. Daly \shortcite{Daly1993} showed that such a normalization is necessary to unify masking predictions across spatial frequencies. Peli et al. \shortcite{Peli1991} demonstrated that this normalization accounts for contrast matching across luminance levels. 


\paragraph{Additive contrast normalization} The multiplicative normalization from \eqref{csf-norm} can be justified by the physical limits of the eye's optics at high frequencies, or by the lateral inhibition at low frequencies \cite{Barten1999}. However, it does not explain contrast constancy across spatial frequencies \cite{Georgeson1975a} --- that is, the observation that the perceived magnitude of large supra-threshold contrast appears the same regardless of spatial frequency. Contrast constancy can be better explained by Kulikowski's model of contrast matching \cite{Kulikowski1976a}, in which the detection threshold is subtracted from (rather than divided by) the absolute contrast
\begin{equation}
C^\prime_{K}=C - \frac{1}{S(\rho_b,L_\ind{bkg},c)}\,.
\end{equation}
Note that the inverse of the contrast sensitivity in this equation is the threshold visibility contrast. Kulikowski \shortcite{Kulikowski1976a} demonstrated that perceived contrast is matched across luminance levels when the corresponding $C^\prime_{K}$ is matching (the sensitivity $S$ in the equation varies across luminance levels). Kulikowski's model assumes that the detection thresholds are caused by additive (neural) noise, and therefore, the visual system "eliminates" the noise by subtracting it from the contrast signal. 

To support both contrast polarities, we modify the above formula as follows:
\begin{equation}
    C^\prime = \sgn{(C)}\max\left\{g\left(|C| - \frac{1}{S(\rho_b,L_\ind{bkg},c)}\right)+1,0 \right\}\,,
    \label{eq:add-cont}
\end{equation}
where $g$ is the gain that controls the strength of the contrast above the threshold. As in the previous case, the constant 1 is used to ensure that the encoded contrast has a value of 1 at the threshold. This is an important modification as it allows us to represent contrasts below the detection threshold. The sign function ensures that we preserve the polarity of the contrast. The encoding is plotted in the right panel of \figref{plot-cont-enc}.


To check how well each contrast encoding represents perceived contrast, in the following sections we generate contrast-matching predictions and compare them with those reported in the literature. 

\begin{figure}
    \centering
    \includegraphics[width=\columnwidth]{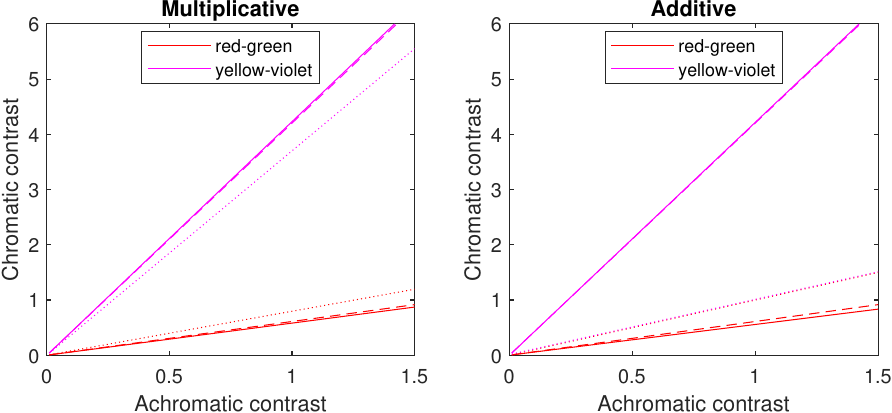}
    \caption{The lines connecting the perceived magnitude of achromatic contrast that matches in appearance the perceived magnitude of chromatic contrast, either red-green or yellow-violet. The dashed lines represent the model based on the measurements of \citeN{Switkes1999}. The dotted lines show the predictions of the original models from \eqref{csf-norm} and \eqref{add-cont} and the continuous lines the same models but after the adjustment. Note that in the right plot for additive contrast encoding the two dotted lines overlap each other.}
    \label{fig:matching-chromatic}
\end{figure}

\subsection{Matching chromatic and achromatic contrast} Since our metric needs to evaluate the impact of both achromatic and chromatic distortions, we need to ensure that the magnitude of both is correctly matched. \citeN{Switkes1999} measured color matches along multiple directions in the color space, including the cardinal directions of the DKL space that we use on \ourmethod. They found that suprathreshold contrast can be matched across achromatic and chromatic dimensions by simple multiplicative scaling. In \figref{matching-chromatic}, we plot their contrast-matching model as two dashed lines --- matching achromatic contrast to either reg-green or yellow-violet chromatic directions. The dotted lines, representing the predictions of our contrast encoding models from \eqref{csf-norm} and \eqref{add-cont}. The dotted lines in the left plot show that the multiplicative encoding with the normalization by the CSF is close to the measurements of Switkes and Crognale, but with some inaccuracy that grows with contrast. This is because the CSF is measured for very small (threshold) contrasts, and any inaccuracy at such fine scales is amplified for large contrast values. The additive model, shown as two overlapping dotted lines in the right plot, cannot predict contrast matches across color directions.

To improve the accuracy of these matches, we introduced corrections into the contrast encoding equation for the multiplicative contrast encoding:
\begin{equation}
    C^\prime = m_c\,C\,S(\rho_b,L_\ind{bkg},c)\,,
\end{equation}
where $m_c=\begin{bmatrix}1 & 1 & 1.45 & 0.95\end{bmatrix}$ for the color channels:
\begin{equation}
    c\in\{\ind{Ach}\sust, \ind{Ach}\trans, \ind{RG}, \ind{YV}\}
\end{equation} corresponding to the achromatic sustained, achromatic transient, chromatic red-green, and chromatic yellow-violet channels. The additive contrast encoding needs to include an additional gain factor: 
\begin{equation}
    C^\prime = \sgn{(C)}\max\left\{g\,m_c\left(|C| - \frac{1}{S(\rho_b,L_\ind{bkg},c)}\right)+1,0 \right\}\,,
    \label{eq:add-cont-cmatch}
\end{equation}
where $m_c=\begin{bmatrix}1 & 1 & 1.7 & 0.237\end{bmatrix}$. The adjusted contrast encoding equations are plotted as continuous lines \figref{matching-chromatic}.

\subsection{Matching contrast across frequencies}

\begin{figure}
    \centering
    \includegraphics[width=\columnwidth]{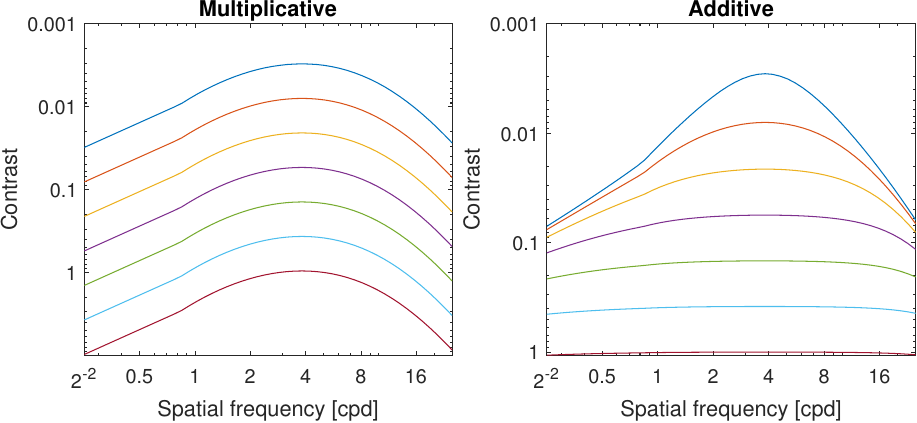}
    \caption{The lines connecting the matching magnitude of contrast across frequencies, as predicted by the multiplicative (left) and additive (right) contrast encoding.}
    \label{fig:matching-frequency}
\end{figure}

The seminal paper of \citeN{Georgeson1975a} demonstrated contrast constancy --- the observation that the perceived magnitude of contrast differs across spatial frequencies when the contrast is small (near the detection threshold) but when the contrast is far above the threshold, there is little difference in the perceived magnitude of contrast. This property is better captured by the additive contrast encoding model, shown in the right panel of \figref{matching-frequency}. This contrast constant effect is easy to observe in everyday life --- objects seen from close and far distances will fall into very different spatial frequency ranges. Yet, we do not observe changes in object appearance as we move closer or further away from them. The multiplicative contrast encoding technique does not directly model the property of contrast constancy. 

\subsection{Matching contrast across luminance}

\begin{figure}
    \centering
    \includegraphics[width=\columnwidth]{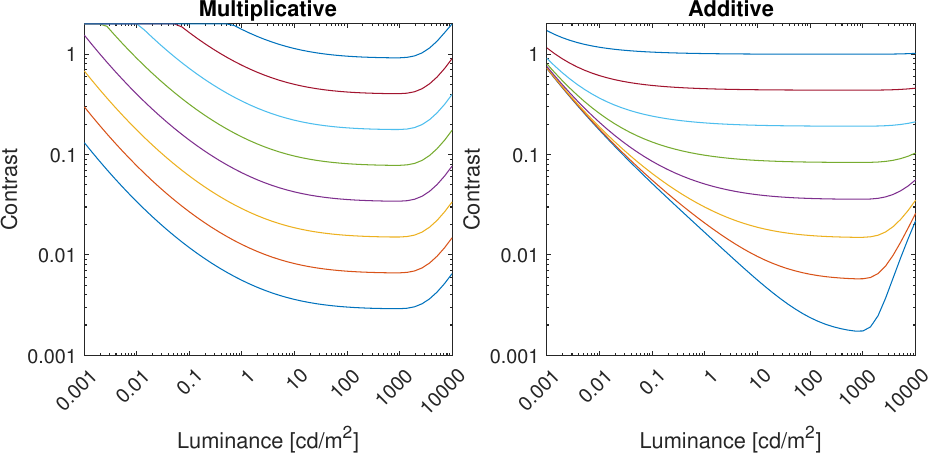}
    \caption{The lines connecting the matching magnitude of contrast across luminance, as predicted by the multiplicative (left) and additive (right) contrast encoding.}
    \label{fig:matching-luminance}
\end{figure}

Multiple authors investigated whether contrast constancy generalizes across luminance levels \cite{Kulikowski1976a,Georgeson1975a,Hess_1990,Peli1991,Peli1995} and they all observed quite a significant deviation from contrast constancy, especially when the luminance drops significantly below photopic levels. However, there is no agreement on how to model such a deviation from contrast constancy. Both \citeN{Kulikowski1976a} and \citeN{Geor1991} proposed additive contrast encoding to explain their contrast matching data. However, \citeN{Peli1995} demonstrated that a multiplicative model better explains the data when contrast is seen naturally rather than using a dichoptic presentation used in other studies (each eye sees a different luminance). The predictions for both models are shown in \figref{matching-luminance}.


\section{Contrast masking}
\label{sec:cont-masking-models}

The main purpose of the masking model is to transform physical differences in contrast between two images or frames into perceived differences. Here, consider three models of masking: a model based on contrast transducer functions, such as the one proposed by Watson and Solomon \cite{Watson1997}, a model based on mutual masking, such as the one proposed in Daly's original VDP \cite{Daly1993}, and the similarity formula, used in SSIM and other metrics.

\paragraph{Contrast transducer}
The difference between two band-limited images in Watson and Solomon's \cite{Watson1997} masking model is expressed as a difference between two contrast transducer functions:
\begin{equation}
D_{b,c,f}(\pixcoord)=\left| t(C_{b,c,f}^{\prime\ind{test}}(\pixcoord)) - t(C_{b,c,f}^{\prime\ind{ref}}(\pixcoord)) \right|
\end{equation}
where $b$ is the index of the spatial frequency band, $c$ is the channel index (two achromatic and two chromatic), and $f$ is the frame index. $C_{b,c,f}^{\prime\ind{test}}$ and $C_{b,c,f}^{\prime\ind{ref}}$ correspond to the encoded contrast in the test and reference images (refer to the main paper). The transducer function is formulated as:
\begin{equation}
    t(C^\prime_{b,c,f}) = g_\ind{T}\frac{\sgn(C^\prime_{b,c,f}(\pixcoord))\left|C^\prime_{b,c,f}(\pixcoord)\right|^p}{0.2+\sum_i k_{i,c} \left(\left|C^\prime_{b,i,f}\right|^{q_c} * g_{\sigma_\ind{sp}}\right)(\pixcoord)}\,,
\end{equation}
where $p$ and $q_c$ are the parameters of the model. The masking parameter $q_c$ is set separately for each channel (two achromatic and two chromatic channels). $g_\ind{T}$ is the gain of the transducer that lets us control the range of visual difference values. The constant of 0.2 was selected so that the facilitation (the dip in the contrast discrimination function) coincides with the measurements, as explained in \secref{contrast-masking-plots}. The expression in the denominator pools contrast in a local spatial neighborhood by convolving with a Gaussian kernel $g_{\sigma_\ind{sp}}$ with the standard deviation of $\sigma_\ind{sp}$. The sum in this expression pools contrast across channels according to the cross-masking coefficient $k_{c,i}$. 
$C^\prime$ is the encoded contrast, as explained in \secref{contrast-enc}.

\paragraph{Mutual masking}
Alternatively, the difference between two band-limited images can be expressed using the mutual masking model, as proposed by Daly \cite{Daly1993}:
\begin{equation}
    D_{b,c,f}(\pixcoord)=\frac{\left|C^{\prime\,\testF}_{b,c,f}(\pixcoord)-C^{\prime\,\refF}_{b,c,f}(\pixcoord)\right|^p}{1 + (C^\maskF_{b,c,f}(\pixcoord))^{q_c}}
    \label{eq:masking-model}
\end{equation}
First, the mutual masking of test and reference bands is calculated as \cite[p.192]{Daly1993}:
\begin{equation}
    C^\text{mm}_{b,c,f}(\pixcoord) = \min\left\{\left|C^{\prime\,\testF}_{b,c,f}(\pixcoord)\right|,\left|C^{\prime,\refF}_{b,c,f}(\pixcoord)\right|\right\}\,.
    \label{eq:mutual-masking}
\end{equation}
Then, similarly to the previous model, the mutual masking signal is pooled in a small local neighborhood by convolving with a Gaussian kernel $g_{\sigma_\ind{sp}}$, and combined across channels, accounting for cross-channel masking: 
\begin{equation}
    C^\maskF_{b,c,f}(\pixcoord) = \sum_i k_{i,c} (C^\ind{mm}_{b,i,f} * g_{\sigma_\ind{sp}})(\pixcoord)  \,.
    \label{eq:mask-min}
\end{equation}
One shortcoming of the mutual masking model is that it does not account for self-masking --- the case in which the reference contrast is 0 and the test contrast is attenuated when its value is high. In practice, $D_{b,c,f}$ can reach very high values (over 10\,000) when the sensitivity is high. This is unrealistic behavior as neurons cannot encode values of such high dynamic ranges. For that reason, we need to limit the range of contrast difference values with a smooth clamping function:
\begin{equation}
    \hat{D}_{B,c,f}(\pixcoord) = \frac{D_\ind{max}\,D_{B,c,f}(\pixcoord)}{D_\ind{max} + D_{B,c,f}(\pixcoord)}\,,
    \label{eq:mask-clamped}
\end{equation}
where $D_\ind{max}$ is the maximum value that the visual difference can attain. 

\paragraph{Similarity}
The alternative masking model, found in SSIM and many other metrics, can be formulated as:
\begin{equation}
    D_{b,c,f}(\pixcoord)=D_\ind{max} - D_\ind{max}\frac{2\,\left|C^{\prime\,\testF}_{b,c,f}(\pixcoord)\right|\,\left|C^{\prime\,\refF}_{b,c,f}(\pixcoord)\right|+\epsilon}{\left(\hat{C}^{\testF}_{b,c,f}\right)^2(\pixcoord) + \left(\hat{C}^{\refF}_{b,c,f}\right)^2(\pixcoord) + \epsilon}\,,
\end{equation}
where $\epsilon = D_\ind{max}-1$ is selected so that the resulting visual difference is 1 when the test contrast is at the detection threshold ($C^{\prime\,\testF}_{b,c,f}=1$) and the reference contrast is 0. $D_\ind{max}$ controls the maximum value of the visual difference.
The standard formula typically uses the same values in both nominator and denominator. Here, we modify the denominator so that it contains masking signal $\hat{C}^{\testF}_{b,c,f}$ and $\hat{C}^{\refF}_{b,c,f}$ associated with the test and reference images. For the test image
\begin{equation}
    \hat{C}^\testF_{b,c,f}(\pixcoord) = \sum_i k_{i,c} (C^{\prime\testF}_{b,i,f} * g_{\sigma_\ind{sp}})(\pixcoord)  \,.    
\end{equation}
and the masking signal for the reference frame is computed analogously.

\begin{figure}
    \centering
    \includegraphics[width=\columnwidth]{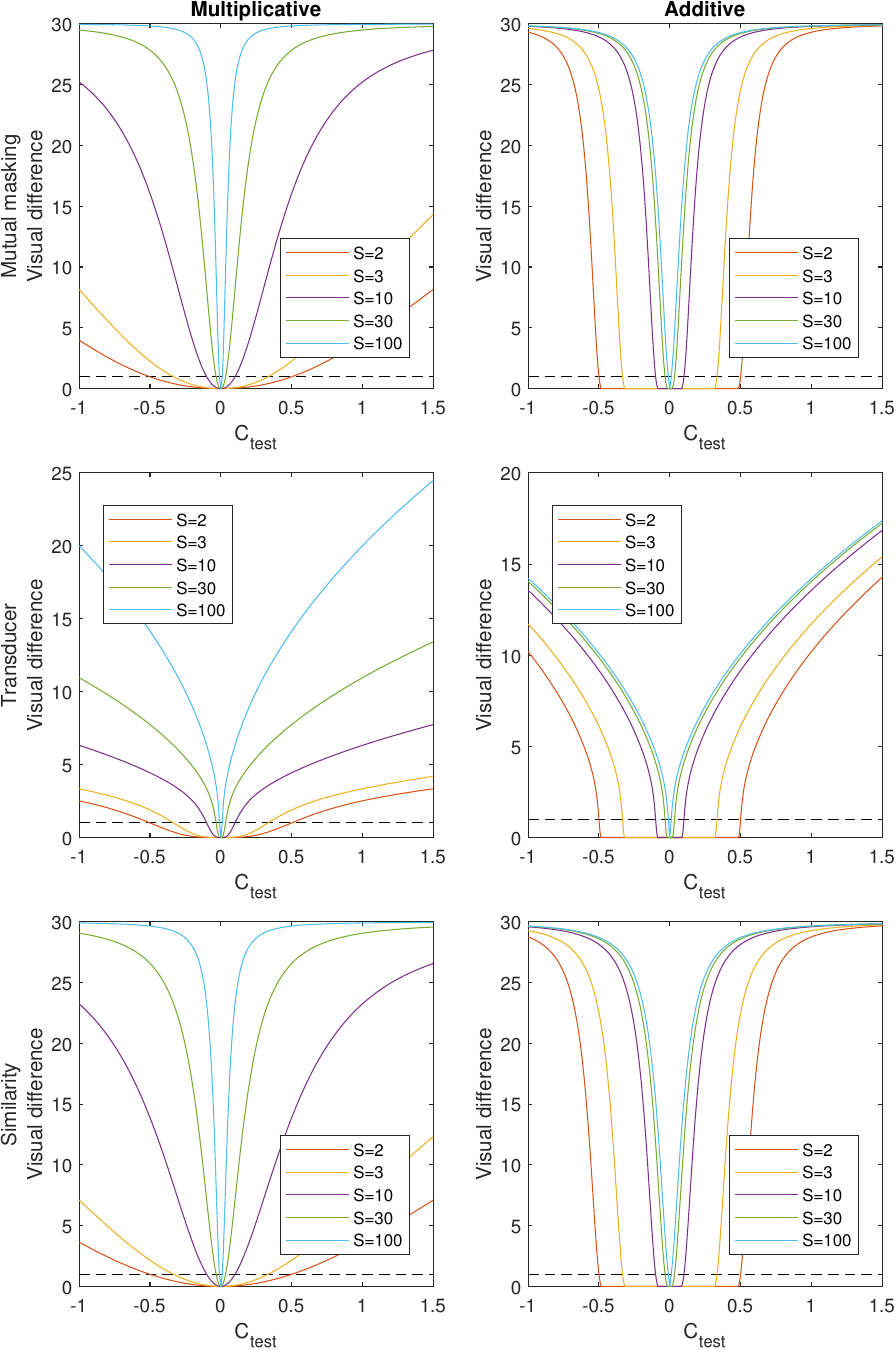}
    \caption{Model response for self-masking --- when the reference contrast is 0 (is a uniform field) but the test contrast (x-axis) and sensitivity (colors) vary. The predictions are shown for all combinations of contrast coding (columns) and masking models (rows). The dashed horizontal line at $D=1$ represents the difference at the detection threshold.}
    \label{fig:self-masking}
\end{figure}

\subsection{Self-masking}

To give more insights into the three masking models explained above and the two contrast encodings, we plot the model predictions for the case of self-masking in \figref{self-masking}. Self-masking is the case in which the contrast (in the test) is masked by itself and is not influenced by the contrast in the reference image (the reference contrast is). This scenario may appear e.g. when the reference image is a uniform field, and the test image contains a pattern that we want to detect. All the plots in \figref{self-masking} show that the visual difference is equal to 1 (the dashed horizontal line) when the test contrast is at the detection threshold (equal to $\nicefrac{1}{S}$). The additive contrast encoding (right column) will result in small contrast being ignored when the sensitivity is low, while the multiplicative encoding will compress contrast below the detection threshold. The mutual masking and similarity models are remarkably similar to each other in terms of self-masking. 

\begin{figure}
    \centering
    \includegraphics[width=\columnwidth]{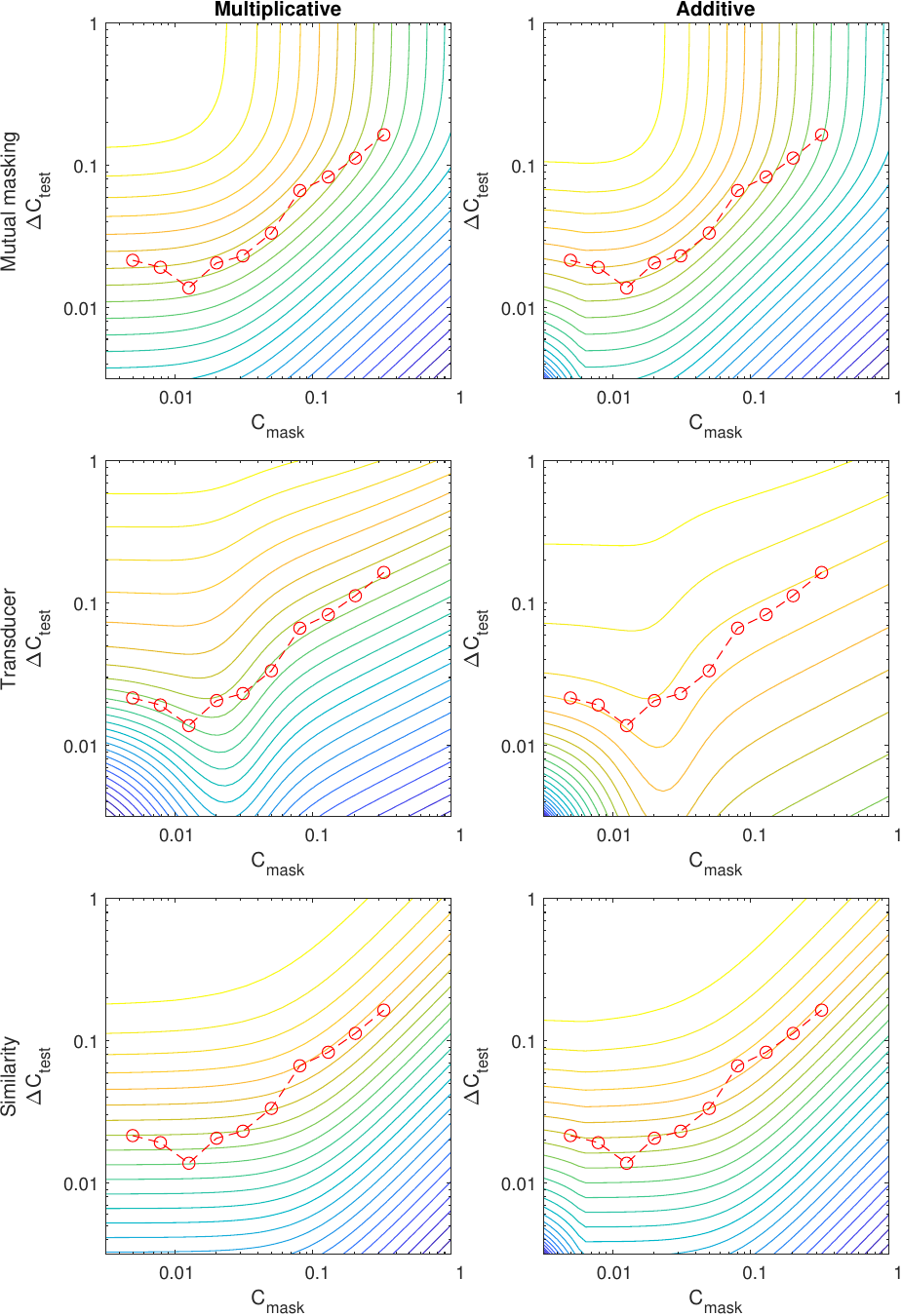}
    \caption{Model response for contrast masking. The reference image contains a certain physical contrast $C^\ind{mask}$ (x-axis), and the test image contains the physical contrast of $C^\ind{mask}+{\Delta}C^\ind{test}$ (y-axis). The contour lines denote the response of the model (blue for the smallest and yellow for the largest values). The red circles denote the contrast masking measurements from \cite[Figure 3b, data for KMF]{Foley1994a}. The axes roughly correspond to those found in the example in \figref{masking-example}.}
    \label{fig:contrast-masking}
\end{figure}

\begin{figure}
    \centering
    \includegraphics[width=\columnwidth]{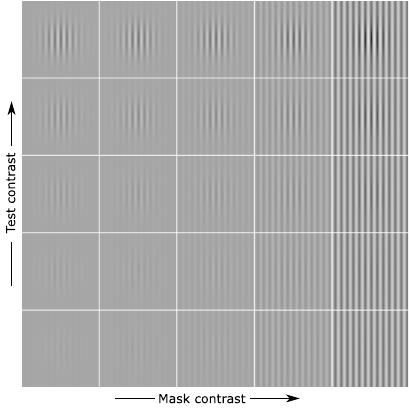}
    \caption{The example of typical stimuli used in contrast masking experiments. A test Gabor patch is superimposed (added) on the background of a sinusoidal grating of the same frequency (a masker). As the contrast of the masker is increased (towards the right), a higher test contrast of the Gabor patch is needed to detect it.}
    \label{fig:masking-example}
\end{figure}

\subsection{Contrast masking}
\label{sec:contrast-masking-plots}

The standard contrast masking experiment involves showing a test pattern superimposed on top of a masking pattern, such as the one shown in \figref{masking-example}. We simulate the same scenario for our six combinations of masking models and contrast encodings and show the results in \figref{contrast-masking}. In each plot, we include the measurements of contrast masking from \cite{Foley1994a} as red dots. We expect the contour lines to follow the curve formed by those measurements. In the plots, we can see that only the transducer models the facilitation that makes patterns easier to detect when the masker is near the detection threshold. This is shown as a small dip in the measurements by \citeN{Foley1994a}. Mutual masking and similarity models show similar behavior.




\section{Training considerations}
\label{sec:training-considerations}

While the contrast encoding and masking models proposed in the previous section are well-defined for an arbitrary contrast, these models are not differentiable out-of-the-box and, therefore, cannot be used in gradient-based optimization. To make them differentiable, we replace the sign function with a hyperbolic tangent function:
\begin{equation}
    \sgn(C) \approx \tanh(10000\,C)\,.
\end{equation}
The exponential function is not differentiable at 0, therefore, we need to approximate it as:
\begin{equation}
    |C|^p \approx (|C|+\epsilon)^p - \epsilon^p\,,
\end{equation}
where $\epsilon=0.00001$. 


\section{Quality metric timings}
\label{sec:timings}

\begin{figure}
    \centering
    \includegraphics[width=\columnwidth]{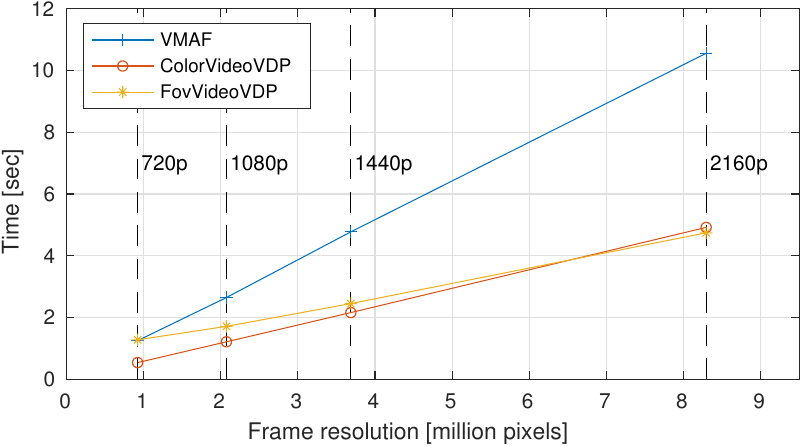}
    \caption{The processing times for 50-frame video of different resolutions (x-axis).
    \ourmethod{} and FovVideoVDP were run on an Nvidia Quadro RTX 8000 GPU, while VMAF was run on 4 cores of an Intel Xeon Gold 5218 CPU @ 2.30\,GHz.
    The dashed vertical lines indicate the standard video resolutions, from 720p to 2160p (4K). All the reported times were averaged across five runs.}
    \label{fig:timings-res}
\end{figure}

We measured the times required to process 50 video frames at resolutions ranging from 720p to 4K. The measurements were averaged across 5 runs and excluded the times required to load and decode the frames. The measured times, illustrated in \figref{timings-res}, show that \ourmethod{}
processes large videos faster than VMAF (note however, that VMAF was running on a CPU) and in about the same time as FovVideoVDP.
It must be noted, however, that unlike \ourmethod both VMAF and FovVideoVDP operate only on luma or luminance and ignore color channels. VMAF features are much less expensive to extract and FovVideoVDP processes only 2 visual channels (sustained and transient achromatic) compared to the 4 channels processed by \ourmethod{}.

\section{LIVE\,HDR dataset alignment experiment}
\label{sec:livehdr-alignment}

In order to enable training on multiple datasets, it is essential to bring their quality scores to a common scale. This ensures that quality scores from one dataset are directly comparable and equivalent to those from another dataset. While both XR-DAVID and UPIQ employ the same quality units (JODs), LIVE~HDR was originally collected using the mean-opinion-score (MOS) units. To account for this discrepancy, we ran an additional experiment that let us scale LIVE~HDR MOS values in the JOD units. 

\begin{figure}
    \centering
    \includegraphics[width=0.6\columnwidth]{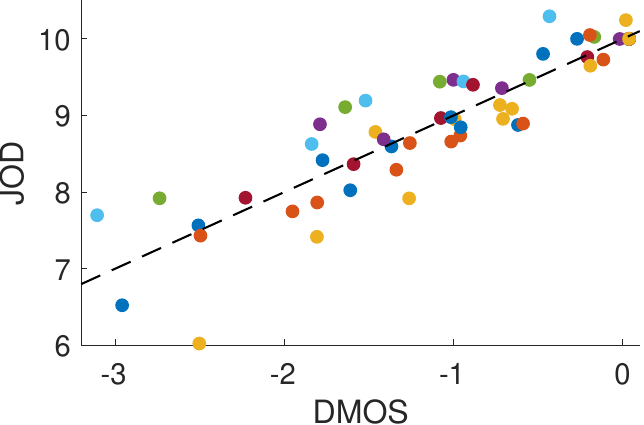}
    \caption{We conducted a subjective study, obtaining $\textnormal{JOD}$ scores for a subset of videos from the LIVE HDR dataset. A linear fit was obtained as follows: $\textnormal{JOD} = -0.054*\textnormal{DMOS} + 0.037$, which was deemed acceptable. $\textnormal{JOD}$ values are shifted from 0 to 10 by convention.}
    \label{fig:liveHDR}
\end{figure}

We first calculated differences of mean opinion scores (DMOS) by subtracting the score of the reference (highest quality video) from each condition. Next, in order to maximize the sampling of quality across the dataset, we carefully selected 10 videos that contained a wide distribution of MOS values. For each of these videos, we then selected 6 different levels of distortion. We used this content to conduct an experiment using the same experimental procedure as for XR-DAVID, as explained in the main paper. Display and environment settings were adapted to mimic those used in the original LIVE HDR study. 12 participants took part in the study, and completed 10 repetitions (batched) of \emph{ASAP} sampling, for a total of $10\times6\times10=600$ trials each. The resulting JOD values for each condition were used to find a linear mapping from the LIVE~HDR DMOS scores to our experimentally derived JOD scores. The resulting linear fit, illustrated in \figref{liveHDR}, demonstrates a satisfactory level of accuracy. Thus, we applied this mapping to the entire LIVE HDR dataset, effectively rescaling it to JOD units.

\section{XR-DAVID content}
\label{sec:xr-david-content}

\begin{figure*}[p]
    \centering
    \includegraphics[width=\textwidth]{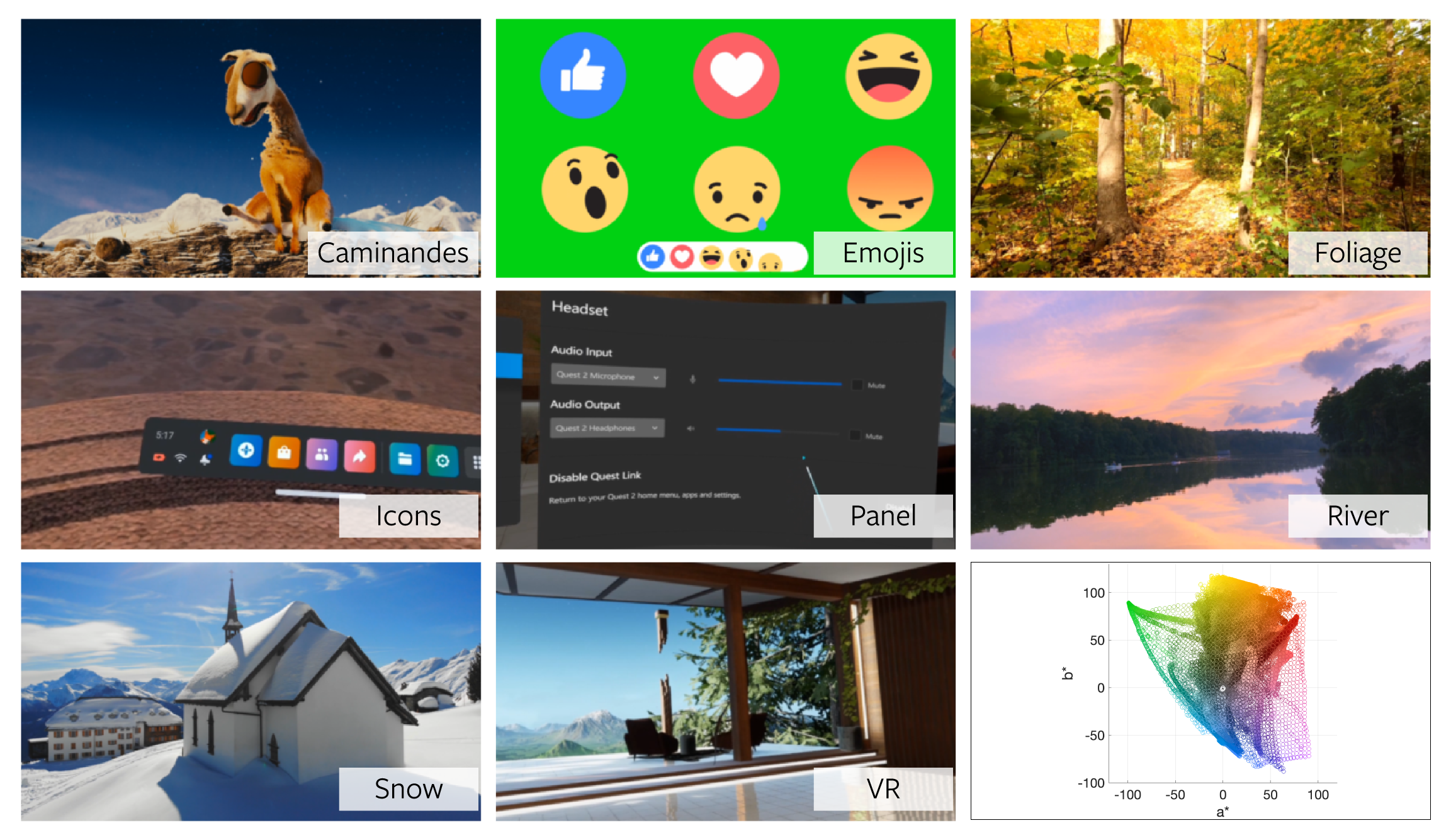}
    \caption{Thumbnails of 8 out of 14 scenes used in the XR-DAVID study. The bottom-right plot shows the color gamut coverage of all the pixels in the reference videos on a CIELAB a* b* plane. The thumbnails of the videos containing human subjects could not be included due to institutional policy.}
    \label{fig:thumbnails_allvid}
\end{figure*}

\begin{figure*}
    \vspace{-6mm}
    \subfloat[]{\label{level1}\includegraphics[width=0.83\textwidth]{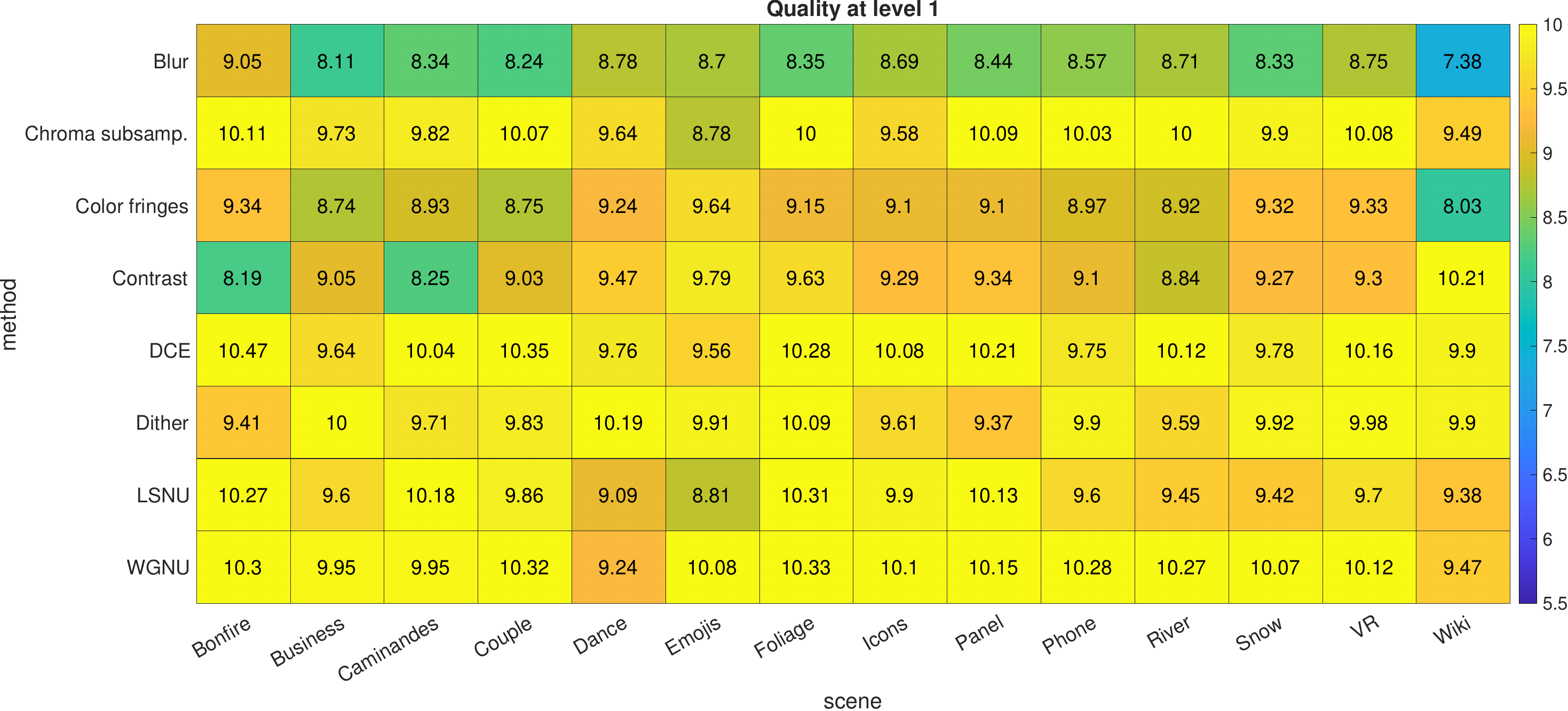}} \\
   \vspace{-6mm}
    \subfloat[]{\label{level2}\includegraphics[width=0.83\textwidth]{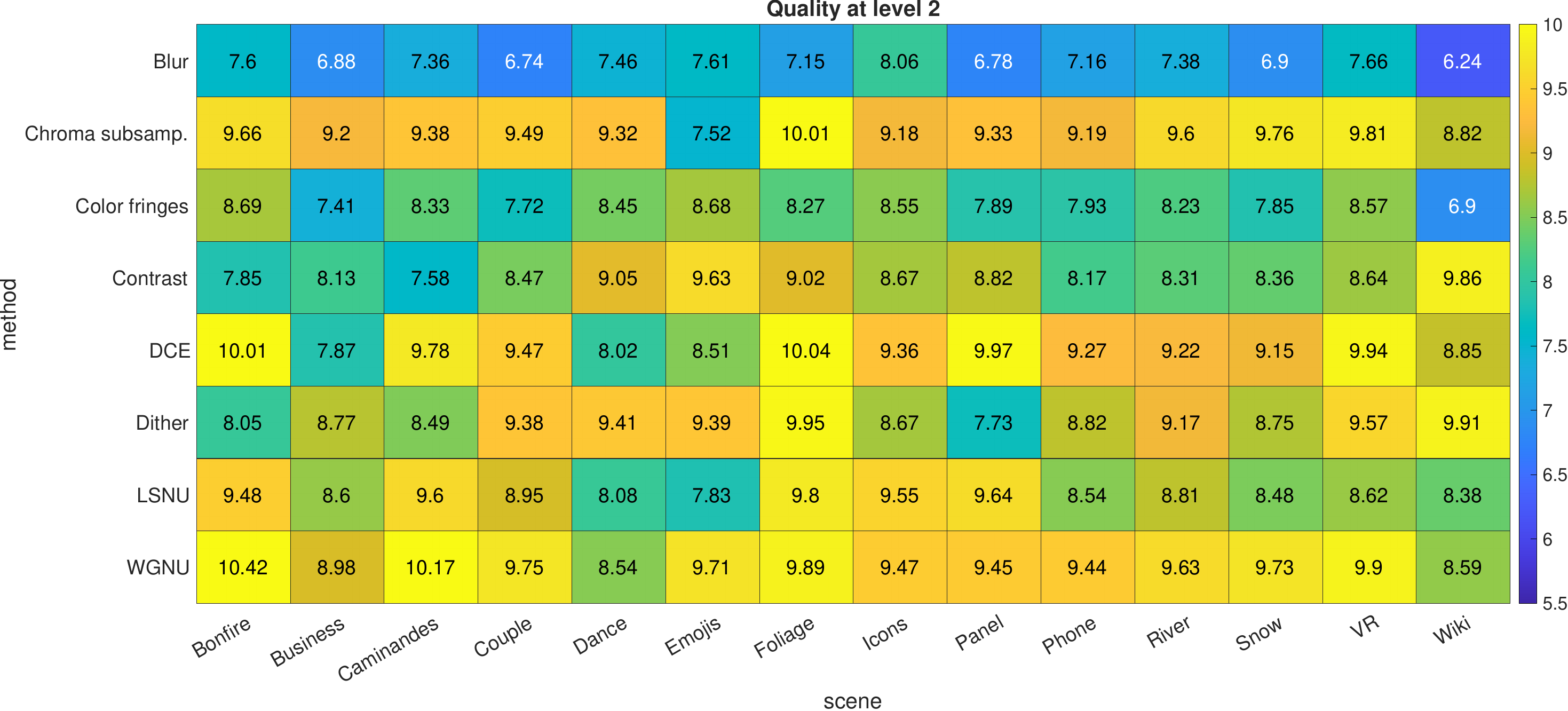}}\\
    \vspace{-6mm}
    \subfloat[]{\label{level3}\includegraphics[width=0.83\textwidth]{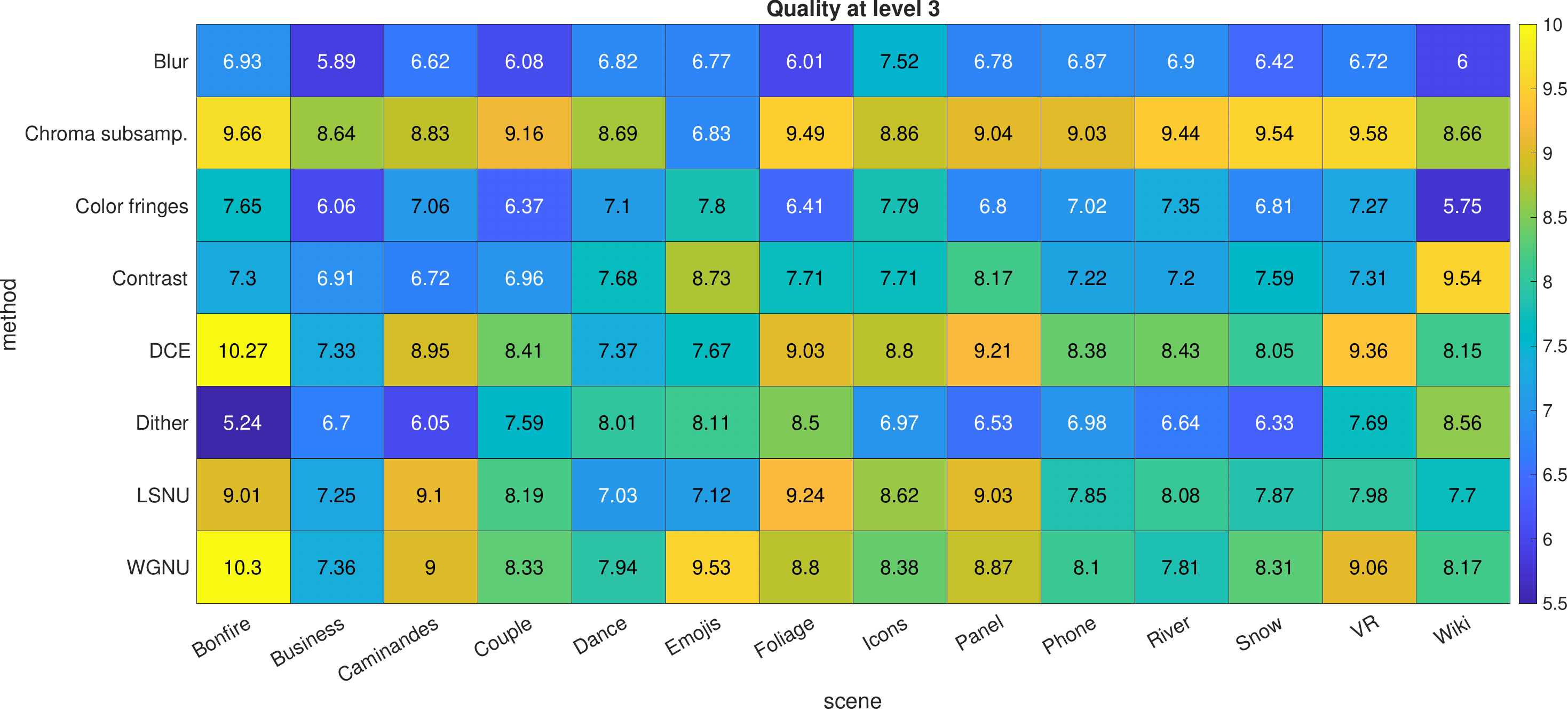}}
    \vspace{-6mm}
    \caption{The results of our XR-DAVID color video quality experiment. For each base video, 8 different artifacts were studied at 3 intensity levels (1:top, 2:middle, and 3:bottom). The responses are scaled on a single perceptual $\textnormal{JOD}$ scale, counting down from 10 by convention. Increasing magnitudes of perceived distortion can be observed at stronger distortion levels (top-to-bottom). In addition, large differences in  artifact visibility can be observed across content (columns).}
    \label{fig:color-quality-study-results}
\end{figure*}

Thumbnails of select scenes used in the XR-DAVID dataset are shown in \figref{thumbnails_allvid}. The source data for each video is shown in \tableref{xr-david-sources}. Note that for longer videos, sections of approximately $5$ seconds from the start of each video were used in the study, with the exception of Caminandes where a shot from approximately 1:54-2:00 of \emph{"Caminandes 3: Llamigos"} was used. Videos were downsampled to a resolution of $910\times540$ using bilinear interpolation, which produces an effective visual resolution of $\approx40$ppd when seen by users from a distance of $28.9$'. Prior to display, videos were upscaled using nearest-neighbor interpolation by a factor of $2$ to $1920\times1080$ to match the native resolution of the display. Detailed numerical results of our experiment are shown in \figref{color-quality-study-results} for each of the $3$ levels of intensity per artifact studied.\\

\begin{table}[]
    \centering
    \caption{XR-DAVID assets}
    \label{tab:xr-david-sources}
\begin{tabular}{ c||c|c }
 \toprule
 Video & Source & Duration\\
 \midrule \midrule
Bonfire    & Kindel Media \href{https://www.pexels.com/video/group-of-friends-having-pizza-and-beer-in-front-of-the-bonfire-7147814/}{[link]} &   5.6s\\
 
Business    & Kindel Media \href{https://www.pexels.com/video/a-woman-poses-by-leaning-on-a-ledge-inside-a-building-3044756/}{[link]} &   5.6s\\

Caminandes    & Blender Foundation \href{http://www.caminandes.com/}{[link]} &   4.8s\\

Couple    & RDNE Stock project \href{https://www.pexels.com/video/city-restaurant-couple-love-5642589/}{[link]} &   5.6s\\

Dance    & Anna Shvets \href{https://www.pexels.com/video/woman-with-headphones-dancing-4316089/}{[link]} &   5.6s\\

Emojis    & emirkhan bal \href{https://www.pexels.com/video/facebook-icon-like-green-background-7451172/}{[link]} &   5.6s\\
 
Foliage    & German Korb \href{https://www.pexels.com/video/the-forest-ground-covered-in-fall-leaves-5644241/}{[link]} &   5.6s\\
 
Icons    & Generated by the authors &   5.6s\\
 
Panel    & Generated by the authors &   5.6s\\

Cellphone    &  Lina Fresco \href{https://www.pexels.com/video/woman-using-a-red-smartphone-3819368/}{[link]} &   5.6s\\
 
River    &  Theresa. Nguyen \href{https://www.pexels.com/video/people-rowing-boats-at-a-beautiful-lake-5594635/}{[link]} &   3.7s\\
 
Snow    &  SwissHumanity Stories  \href{https://www.pexels.com/video/a-drone-shot-of-a-church-in-belalp-3818213/}{[link]} &   5.6s\\
 
River    &  Theresa. Nguyen \href{https://www.pexels.com/video/people-rowing-boats-at-a-beautiful-lake-5594635/}{[link]} &   3.7s\\
 
VR    & Generated by the authors &   5.6s\\

Wiki    & Generated by the authors &   5.2s\\
\bottomrule
\end{tabular}
\end{table}


\captionsetup[subfigure]{labelformat=empty}

\bibliographystyle{ACM-Reference-Format}
\bibliography{supplementary}